%% file: main.tex
\documentclass{article}
\usepackage{arxiv}
\usepackage[utf8]{inputenc}
\usepackage[T1]{fontenc}
\usepackage[numbers,sort&compress]{natbib}
\usepackage{amssymb}
\usepackage{amsmath}
\usepackage{graphicx}
\usepackage[dvipsnames]{xcolor}
\usepackage{xspace}
\usepackage{soul}
\usepackage{hyperref}
\usepackage{cleveref}
\usepackage{booktabs}
\usepackage{glossaries}
\usepackage[inline]{enumitem}
\usepackage{tikz}
\usepackage{tcolorbox}
\usepackage{pdflscape}
\usepackage{multirow,makecell}
\hypersetup{
  colorlinks=true, 
  urlcolor=blue, 
  linkcolor=black, 
  citecolor=black 
}
\soulregister\cite7
\soulregister\ref7
\soulregister\pageref7
\setstcolor{red}
\setul{}{.2ex}
\usepackage{rotating}

\glsdisablehyper
\newacronym{llm}{LLM}{Large Language Model}
\newacronym{co}{CO}{Combinatorial Optimization}
\newacronym{cop}{COP}{Combinatorial Optimization Problem}
\newacronym{lp}{LP}{Linear Programming}
\newacronym{pfsp}{PFSP}{Permutation Flowshop Scheduling Problem}
\newacronym{nl4opt}{NL4Opt}{Natural Language for Optimization}
\newacronym{nlp}{NLP}{Natural Language Processing}
\newacronym{stn}{STN}{Search Trajectory Network}
\newacronym{cp}{CP}{Constraint Programming}
\newacronym{milp}{MILP}{Mixed Integer Linear Programming}
\newacronym{nfl}{NFL}{No Free Lunch}
\newacronym{isa}{ISA}{Instance Space Analysis}
\newacronym{nn}{NN}{Neural Network}
\newacronym{gcp}{GCP}{Graph Coloring Problem}
\newacronym{kp}{KP}{Knapsack Problem}
\newacronym{bp}{BPP}{Bin Packing Problem}
\newacronym{jssp}{JSP}{Jobshop Scheduling Problem}
\newacronym{sat}{max-SAT}{Maximum Satisfiability Problem}
\newacronym{rtf}{RTF}{Reasoning–Task–Format}
\newacronym{mae}{MAE}{Mean Absolute Error}
\newacronym{mlp}{MLP}{Multi-Layer Perceptron}
\newacronym{relu}{ReLU}{Rectified Linear Unit}
\newacronym{tost}{TOST}{Two One-Sided Test}
\newacronym{ml}{ML}{Machine Learning}
\newacronym{eos}{EOS}{End-Of-Sequence}
\newacronym{cot}{CoT}{Chain-of-Thought}

\usepackage{xparse}

\NewDocumentCommand{\llama}{o o}{%
  Llama%
  \IfValueT{#1}{-#1}%
  \IfValueT{#2}{-#2}%
  \xspace
}
\NewDocumentCommand{\gpt}{o}{%
  gpt-oss%
  \IfValueT{#1}{-#1}%
  \xspace
}

\newcommand{\midfeature}{%
  \begin{tikzpicture}
    \draw[black] (0,0) circle (.5ex); 
    \fill[black] (0,0) -- ++(0:.5ex) arc[start angle=0, end angle=180, radius=.5ex] -- cycle; 
  \end{tikzpicture}%
}
\newcommand{\easyfeature}{%
\tikz\draw[black,fill=white] (0,0) circle (.5ex);%
}
\newcommand{\hardfeature}{%
\tikz\draw[black,fill=black] (0,0) circle (.5ex);%
}
\newcommand{\standard}{\textsf{standard}}
\newcommand{\code}{\textsf{code-like}}
\newcommand{\nlp}{\textsf{natural language}}
\newcommand{\transformers}{\textsf{transformers}}
\newcommand{\huggingFace}{\textsf{Hugging Face}}
\newcommand{\vllm}{\textsf{VLLM}}

\newcommand{\logisticRegression}{\textsf{LogisticRegression}}
\newcommand{\mostFrequent}{\textsf{MostFrequentClassifier}}
\newcommand{\lightGBM}{\textsf{LightGBM}}

\newcommand{\dvc}{\textsf{dvc}}
\newcommand{\git}{\textsf{git}}
\newcommand{\mlpClassifier}{\textsf{MLPClassifier}}
\newcommand{\python}{\textsf{Python}}

\begin{document}

\renewcommand{\headeright}{Accepted Manuscript}
\renewcommand{\undertitle}{Accepted for publication in Computers \& Operations Research}
\renewcommand{\shorttitle}{Behavior and Representation in Open-Weight LLMs for CO}

\title{Behavior and Representation in Open-Weight Large Language Models for Combinatorial Optimization: From Feature Extraction to Algorithm Selection}

\author{
  Francesca Da Ros \\
  University of Udine, Italy \\
  \texttt{francesca.daros@uniud.it}
  \And
  Luca Di Gaspero \\
  University of Udine, Italy \\
  \texttt{luca.digaspero@uniud.it}
  \And
  Kevin Roitero \\
  University of Udine, Italy \\
  \texttt{kevin.roitero@uniud.it}
}
\date{}

\maketitle

\begin{abstract}
Recent advances in Large Language Models (LLMs) open new perspectives for automation in optimization, yet little is known about whether their internal representations capture problem structure or algorithmic behavior. 
We investigate whether representations learned by frozen, open-weight LLMs for combinatorial optimization instances can support downstream decision tasks. 
The goal is not to replace exact feature extractors or to propose a new algorithm, but to assess whether such representations are reusable for feature recovery and algorithm selection.
Our methodology combines \emph{direct querying}, which tests explicit feature extraction, with \emph{probing} analyses of whether this information is implicitly encoded in the hidden layers. The probing framework is further extended to a per-instance \emph{algorithm selection} task. 
Experiments span four benchmark problems, three instance representations, and five open-weight models from $3$B to $120$B parameters across the \llama instruct and \gpt reasoning families, including a chain-of-thought study on \llama models.
Results show a limited ability to recover features explicitly, particularly those requiring structured computation, while part of this information remains implicitly encoded in the hidden states. 
A consistent gap separates implicit encoding from explicit retrieval. Model scale attenuates this gap, whereas the effect of chain-of-thought prompting depends strongly on model size, and in all cases the gap remains open. 
Reasoning-oriented models also tend to abstain more when reliable computation is not possible. 
Notably, the predictive power of LLM hidden-layer representations is comparable to traditional feature extraction across all five models, indicating they can act as effective surrogate descriptors for downstream optimization tasks.
\end{abstract}


\section{Introduction}
\label{sec:introduction}

\gls{co} has long been used to tackle decision-making problems across many domains, including logistics, scheduling, and resource allocation~\cite{DBLP:journals/cor/WuIHGSY24,DBLP:journals/cor/YuSBS23,DBLP:journals/cor/NascimentoQJ21}. Despite major advances in exact and heuristic methods, as well as the growing integration of \gls{ml} techniques~\cite{DBLP:journals/csur/Talbi21a}, developing and analyzing algorithms remains a largely manual, expertise-driven process. This process is time-consuming and difficult to automate, particularly when deep domain knowledge plays a critical role in performance.

In parallel, recent developments in \glspl{llm} have demonstrated their capacity to process text, generate executable code, and generalize across domains~\cite{DBLP:journals/corr/abs-2501-04040}. These capabilities have prompted researchers to explore how \glspl{llm} can support optimization-related tasks, from model formulation to heuristic generation and algorithmic analysis~\cite{fan2024artificialintelligenceoperationsresearch}. While many of these studies have produced encouraging results, they primarily assess the \emph{functional performance} of the models (i.e., what they can do) without investigating their \emph{representational understanding}, that is, whether their internal representations capture information about problem structure and algorithmic behavior.

Understanding how \glspl{llm} represent optimization problems is therefore particularly relevant.
If \glspl{llm} encode meaningful information about problem structure or algorithmic performance, they could provide problem-independent representations for downstream tasks, rather than acting as conventional feature extractors engineered for specific problem formulations.
Such representations could support algorithm selection, performance prediction, and hybrid optimization frameworks that combine learning and search.
Conversely, the absence of such structure would highlight their current limitations in providing general-purpose representations for combinatorial optimization.

In this study, we investigate the extent to which \glspl{llm} capture information about \glspl{co}, distinguishing between information encoded in their internal representations and information that can be explicitly recovered through direct querying. We focus on instance-level features and per-instance algorithm selection as complementary tests of problem characterization and algorithmic performance prediction.
We address three research questions:
\begin{enumerate}[label = \textbf{RQ\arabic*}]
\item\label{rq:direct-querying} To what extent can \glspl{llm} identify salient structural or numerical characteristics of \gls{co} instances from their representations?
\item\label{rq:probing-features} Can \glspl{llm} encode information about \gls{co} instances, and is this information accessible through probing?
\item\label{rq:probing-alg-selection} Can representations extracted from the hidden layers of \glspl{llm} serve as effective instance features for per-instance algorithm selection, and how does their predictive power compare to that of traditional handcrafted features?
\end{enumerate}
\ref{rq:direct-querying} captures the behavioral dimension of our analysis, examining explicit feature awareness and generation capabilities.
\ref{rq:probing-features} investigates the internal, latent encoding of instance-level information.
\ref{rq:probing-alg-selection} assesses the practical utility of \glspl{llm}, testing whether their learned representations can inform algorithmic decision. To answer these questions, we combine \emph{direct querying} and \emph{probing} analyses.

Our analysis focuses on five open-weight \glspl{llm} spanning the \llama instruct and \gpt reasoning families in the $3$B--$120$B parameter range, and all experiments are conducted on a stratified $5\%$ subset of the original benchmark collection, designed to preserve coverage across problems, feature complexity levels, and downstream labels while keeping the experimental campaign computationally feasible. Closed, commercial frontier models are excluded for reproducibility and cost reasons, since their behavior cannot be frozen across revisions of the study, their weights cannot be released alongside our artifact, and their per-token inference costs would be incompatible with the scale of our experiments.

In the first part of our study, we evaluate whether models exhibit awareness of instance-level features typically indirectly associated with algorithmic performance, and whether such features can be explicitly and implicitly retrieved from instance representations.
We further analyze the role of explicit reasoning by comparing standard direct querying with Chain-of-Thought prompting, examining how its impact varies across model scales.
In the second part, we use probing to examine whether the internal representations can act as surrogates for handcrafted features traditionally obtained through explicit feature engineering. We conduct experiments across four representative \glspl{cop}, using publicly available instance sets and algorithm portfolios from prior \gls{isa} studies.
Our evaluation spans multiple feature complexities, instance representations, model scales and prompting strategies, and probing configurations, providing a comprehensive perspective on the relationship between explicit and implicit representations in \glspl{llm}.

The central hypothesis of this work is that explicit numerical recovery and latent representational usefulness are distinct properties. An \gls{llm} may fail to compute or return an exact feature value when prompted, while its hidden states may still contain a feature-related signal that is useful for downstream prediction. This distinction is central to our analysis and motivates the joint use of direct querying, probing, and algorithm-selection experiments.

The main contributions of this work are as follows:
\begin{itemize}
\item We introduce a unified experimental framework for evaluating what \glspl{llm} capture about \glspl{cop} through direct querying and probing.
\item We validate the framework across five open-weight models spanning the $3$B--$120$B range, and analyze the role of explicit reasoning through a Chain-of-Thought study on \llama models, showing that its effect depends strongly on model scale.
\item We provide the first large-scale analysis of how \glspl{llm} encode structural, numerical, and algorithmic information across multiple problem domains and input representations.
\item We show that explicit retrieval and latent encoding do not fully coincide: representations may support downstream prediction even when direct querying fails to reliably recover feature values.
\item We release the dataset, code, and prompts to foster reproducibility and further exploration at the intersection of \glspl{llm} and optimization.\footnote{Code and data: \url{https://doi.org/10.5281/zenodo.21891840}.}
\end{itemize}

This study extends our earlier conference publication~\cite{DBLP:conf/evoapps/RosGR25}, which focused solely on direct querying of \glspl{llm} for feature extraction. Here, we expand the analysis along several dimensions: we introduce probing analyses to explore internal representations, generalize the experimental setup to multiple problems and input representations, include a downstream algorithm-selection task, and provide a broader comparative analysis; additionally, we extend the empirical validation from one to five open-weight models and evaluate, on a stratified $5\%$ subset of the benchmark collection, multiple prompting strategies beyond direct querying, including \gls{cot} prompting and native reasoning.

Accordingly, the study is positioned as a representation-analysis framework rather than as a solver-design contribution: it investigates whether frozen LLM representations can serve as problem-agnostic instance descriptors and compares their predictive value with handcrafted and ISA-based features in a standard algorithm-selection pipeline.

The remainder of this paper is organized as follows.
\Cref{sec:background} reviews the background and related work.
\Cref{sec:methodology} describes the proposed methodology.
\Cref{sec:experimental-analysis} presents the experimental analysis whereas \Cref{sec:discussion} interprets the findings under a unified framework.
Finally, \Cref{sec:conclusions} concludes the paper and outlines directions for future research.

\section{Background and Related Work}
\label{sec:background}

This section reports background and related work on the topics of \gls{llm} (\Cref{sec:llms}) and their interpretability (\Cref{sec:background:llm-interpretability}). Following, it summarizes relevant notions regarding \gls{llm} usage in \gls{co} (\Cref{sec:llm-for-co}) and algorithm selection (\Cref{sec:as-for-co}).

\subsection{Large Language Models}
\label{sec:llms}

\glspl{llm} have fundamentally transformed the landscape of \gls{nlp} and, increasingly, fields beyond it, including reasoning and programming~\cite{DBLP:journals/air/Kumar24,DBLP:journals/tist/ChangWWWYZCYWWYZCYYX24}.

\glspl{llm} are \gls{nn} architectures designed to model the probability distributions of natural language sequences. The fundamental processing unit of an \gls{llm} is the token, namely a sub-word or symbol representing a fragment of text, code, or other structured input. Given a sequence of \emph{tokens} $t_{1}, t_{2}, \dots, t_{n}$, an \gls{llm} estimates the conditional probability $P(t_{n} | t_{1}, \dots, t_{n-1})$, thereby learning to predict the next token conditioned on its preceding context. Through large-scale training on heterogeneous textual and symbolic corpora, such models acquire internal representations that capture syntactic, semantic, and relational regularities, which in turn enable them to perform a broad spectrum of reasoning, generation, and comprehension tasks~\cite{DBLP:conf/www/MeguellatiPSD25,DBLP:conf/www/CivelliBD25}.

At their core, most \glspl{llm} implement the Transformer architecture~\cite{DBLP:conf/nips/VaswaniSPUJGKP17}, which relies on self-attention mechanisms to compute contextual dependencies between all tokens in parallel~\cite{DBLP:journals/corr/BahdanauCB14}.
This design allows \glspl{llm} to handle long-range relationships efficiently and to generate outputs that are sensitive to the entire input sequence.

Users interact with the model through a prompt, which is a textual sequence that defines the input context and often specifies the desired task.
The model then produces one or more output tokens that, when decoded, yield natural language or symbolic text.
The process can be guided or restricted through prompt engineering or constrained decoding, the latter enforcing structural, syntactic, or semantic constraints on the generated output, e.g., requiring valid JSON, enforcing type consistency, or limiting values to predefined domains.

\subsection{Large Language Models Interpretability}
\label{sec:background:llm-interpretability}

Despite their remarkable capabilities, the internal mechanisms by which \glspl{llm} encode and organize knowledge remain only partially understood. Two complementary approaches are commonly employed to investigate these mechanisms: \emph{direct querying} (also referred to as \emph{direct questioning} or \emph{prompting}), which analyzes model behavior through carefully designed prompts, and \emph{probing}, which examines the information embedded within the model hidden representations.

Direct querying is a methodology used to assess the behavior of \glspl{llm} by presenting them with a prompt and analyzing their generated responses. It evaluates the model's ability to explicitly retrieve information when prompted, without inspecting internal activations or mechanisms. A well-known example of direct querying is the \emph{needle in a haystack} task,\footnote{\url{https://github.com/gkamradt/LLMTest_NeedleInAHaystack/tree/main}, accessed 14 Oct 2025.} in which a factual statement (i.e., a short text fragment) is embedded within a long context, and the model is prompted to retrieve it \emph{verbatim}. This methodology has been extended to other applications, such as document summarization and citation generation~\cite{laban2024summary}, to evaluate whether relevant information can be surfaced by the model when explicitly prompted to recall it.

Probing offers a systematic and empirically grounded framework for investigating the internal representations of \glspl{llm}. It provides a structured methodology to examine what kinds of information are captured within the models' internal mechanisms, where such information is stored, and how it is organized and retrieved across layers~\cite{DBLP:conf/coling/Ju0DYRL24,marks2024geometrytruthemergentlinear,li2025exploringmultilingualprobinglarge}. Unlike the output-based evaluation described above, which focuses on model behavior elicited through direct querying, probing operates by extracting the hidden layer activations produced as the model processes an input, and analyzing these activations using lightweight predictive models known as \emph{probes}. This approach enables the assessment of whether specific properties, including syntactic, semantic, numerical, and domain-specific features, are encoded in the model representations and can be recovered by a downstream model.

A typical probing setup comprises three main stages:
\begin{enumerate*}[label=\emph{(\roman*)}]
\item a controlled set of inputs is fed into a frozen \gls{llm} to collect activations from one or more internal layers of interest, typically the final layer;
\item an external model (i.e., the probe) is trained on these representations to predict a target property; and
\item the probe's performance is evaluated on unseen data to assess its generalization on the same property.
\end{enumerate*}
Common probe architectures include linear classifiers and simple \glspl{mlp}, each suited to investigating different aspects of how information is encoded within \glspl{llm}~\cite{belinkov-2022-probing,wallace-etal-2019-nlp,kim2019probing}. Simpler probes, such as linear regressors, are typically used to examine whether a property is linearly accessible from the model representations, whereas more complex, non-linear probes can explore relationships that may be distributed or non-linearly structured within the latent space~\cite{cunningham2023sparse}.

Although originally developed for linguistic analysis, probing has been applied to a broad range of interpretability tasks. Several studies have examined whether hidden layers encode syntactic categories or dependency relations, showing that intermediate layers tend to capture structural information, whereas deeper layers increasingly specialize in semantic content~\cite{vulic-etal-2020-probing,belinkov-2022-probing}. The same framework has also been extended to investigate higher-level capabilities such as discourse awareness~\cite{koto-etal-2021-discourse}, factual and relational knowledge~\cite{chanin-etal-2024-identifying}, and hierarchical reasoning~\cite{alleman-etal-2021-syntactic}. For instance, a probe may be trained to distinguish correct from incorrect factual triples (e.g., “Paris is the capital of France” vs.\ “France is the capital of Paris”) or to recover the latent coordinates of entities mentioned in text, thereby revealing whether spatial or relational concepts are implicitly represented within the model. Such applications show that probing can provide insights into the organization of latent representations in \glspl{llm}, extending beyond surface-level linguistic features.

A key advantage of probing lies in its architecture-agnostic nature: it treats the model as a frozen feature extractor and requires neither fine-tuning nor parameter updates. Consequently, it ensures that the subsequent analysis based on the model weights accurately reflects its learned internal state~\cite{beloucif-biemann-2021-probing-pre}. 

\subsection{Large Language Models for Combinatorial Optimization}
\label{sec:llm-for-co}

Several recent studies have investigated the use of \glspl{llm} in \gls{co}~\cite{wang2025largelanguagemodelsoperations,daros2025largelanguagemodelscombinatorial,fan2024artificialintelligenceoperationsresearch}. Early interest in combining \glspl{llm} with \gls{co} emerged from the \gls{nl4opt} competition~\cite{DBLP:conf/nips/RamamonjisonYLLCGHMBZZ21}. It explored whether \gls{nlp} techniques could assist in mathematical modeling through entity recognition and automatic problem formulation, initially focusing on \gls{lp} tasks.
Building upon this, subsequent research has diversified along two main directions:
\begin{enumerate*}[label = \emph{(\roman*)}]
\item \emph{Model definition and formulation}, where \glspl{llm} generate or translate optimization models from textual descriptions~\cite{DBLP:conf/ijcai/PanFWZHL025}; and
\item \emph{Heuristic and algorithm construction}, where \glspl{llm} design, adapt, or parameterize search procedures~\cite{DBLP:journals/corr/abs-2410-14716}.
\end{enumerate*}

With respect to model formulation, recent works have extended beyond \gls{lp} to encompass other paradigms such as \gls{cp}~\cite{DBLP:journals/corr/abs-2506-06052,DBLP:journals/corr/abs-2308-01589} and \gls{milp}~\cite{DBLP:conf/icml/AhmadiTeshniziG24}. In the second direction, applications include the automated discovery of novel heuristics, as exemplified by FunSearch~\cite{DBLP:journals/nature/RomeraParedesBNBKDREWFKF24}, although \citet{DBLP:conf/evoapps/SimRH25} later showed that several simple, classical heuristics outperform those generated by the model across extensive benchmark experiments. Alternative approaches focus on generating and evolving metaheuristics components, as in LLaMEA~\cite{DBLP:journals/tec/SteinB25}, or on enhancing existing algorithms~\cite{DBLP:journals/corr/abs-2503-10968}.

A smaller body of research has explored supportive and analytical roles of \glspl{llm} within \gls{co}, extending beyond problem solving.
For instance, \citet{DBLP:conf/gecco/Sartori0O24a} employed \glspl{llm} to assist users in interpreting and navigating \glspl{stn}, which are graph-based structures that capture the temporal evolution of (meta-)heuristic search processes~\cite{DBLP:journals/simpa/SartoriBO23}.

Despite the growing body of research, most existing studies prioritize functional performance over representational or explanatory understanding. Notable exceptions include the analysis of \gls{llm}-generated code through graph-based representations~\cite{DBLP:conf/gecco/SteinKKB25} and behavioral investigations of generated heuristics using \glspl{stn}~\cite{DBLP:journals/corr/abs-2507-03605}. However, how \glspl{llm} internally encode combinatorial structure, and whether such representations can be effectively accessed or exploited, remains largely unexplored. The limited number of systematic examinations of their hidden representations leaves a substantial gap between observable competence and latent understanding. This work aims to bridge this gap by investigating not only how combinatorial structures are represented within \glspl{llm}, but also how such information can be accessed through direct querying and probing, and whether it can be effectively leveraged in downstream tasks such as algorithm selection.

\subsection{Algorithm Selection in Combinatorial Optimization}
\label{sec:as-for-co}

Within \gls{co}, algorithmic performance varies substantially across instances: an algorithm that performs well on one subset of instances may be ineffective on another, depending on factors such as instance size, structure, or constraint tightness. This variability arises from the inherent complexity of most \glspl{cop}, which are NP-hard and exhibit widely differing computational characteristics across instances. The phenomenon is formalized by the \gls{nfl} theorem~\cite{DBLP:journals/tec/DolpertM97}, which states that no single algorithm can achieve superior performance across all possible problems or instance distributions.

The challenge of selecting or configuring algorithms according to instance characteristics was systematically conceptualized by \citet{DBLP:journals/ac/Rice76}, who defined the Algorithm Selection Problem as the task of learning a mapping between instance features and algorithm performance. In \citeauthor{DBLP:journals/ac/Rice76}'s framework, each problem instance is represented by a vector of measurable characteristics, and each algorithm is evaluated according to one or more performance metrics; the objective is to predict which algorithm is expected to perform best for a given instance, effectively framing the problem as a supervised learning task. This formulation laid the foundation for per-instance algorithm selection and inspired subsequent developments in algorithm portfolios: collections of complementary algorithms designed to exploit performance diversity across instances~\cite{DBLP:journals/jair/XuHHL08,DBLP:conf/sat/ShavitH24}.

Successful algorithm selection increasingly relies on quantitative instance characterization, that is, the identification of features that capture the structural and behavioral properties of problem instances. To support this process, frameworks such as \gls{isa}~\cite{DBLP:journals/csur/SmithMilesM23} and tools such as MATILDA\footnote{\url{https://matilda.unimelb.edu.au/matilda/}, accessed 13~Oct~2025.} have been developed, offering integrated environments that go beyond algorithm selection to include feature extraction, visualization, and mapping of algorithm performance across low-dimensional instance spaces~\cite{DBLP:journals/eor/LiuSWC24,DBLP:journals/scheduling/CosterMSSS22}.

Despite these advances, feature extraction still relies heavily on handcrafted, problem-specific descriptors that demand substantial domain expertise. To address this limitation, recent research has explored automated feature learning, where models learn informative representations directly from data. For instance, \glspl{nn} have been employed to infer latent instance representations~\cite{DBLP:journals/heuristics/AlissaSH23,DBLP:conf/cp/PellegrinoA0KM25}, extending also to black-box optimization settings~\cite{DBLP:conf/gecco/CenikjPE24}.
Our work differs both in scope and realization: to the best of our knowledge, this is the first study to systematically evaluate \glspl{llm} as feature extractors for combinatorial optimization, combining direct querying, probing, and downstream validation across multiple problems, representations, and model scales. It is also the first to investigate whether internal \gls{llm} layers can serve as predictors of algorithmic performance, thereby linking latent model representations to algorithm-selection outcomes.

\section{Methodology}
\label{sec:methodology}

In this section, we present the methodology adopted to investigate how \glspl{llm} represent and process \glspl{cop}, with the broader goal of assessing whether these models encode information that can support algorithm selection. Specifically, our analysis is structured around two complementary perspectives:
\begin{enumerate*}[label=\emph{(\roman*)}]
\item the \emph{explicit recoverability} of instance-level information, i.e., whether the models can retrieve target features when prompted; and
\item their \emph{implicit representations}, i.e., the extent to which internal layers encode information about problem instances and their properties.
\end{enumerate*}
To address these perspectives, we employ a twofold experimental strategy:
\begin{enumerate*}[label=\emph{(\roman*)}]
\item \emph{Direct querying}, wherein models are explicitly prompted to infer instance-level features; and
\item \emph{Probing}, wherein the models internal activations are analyzed to assess whether features or algorithmic performance signals are implicitly represented.
\end{enumerate*}

We remark that a key design choice in this study is the exclusive use of off-the-shelf \glspl{llm} without any form of task-specific fine-tuning. Thus, the objective of this work is to assess whether these models, as released, already encode meaningful representations of combinatorial optimization problems, and whether such representations can support downstream tasks such as feature extraction and algorithm selection. Introducing fine-tuning would shift the focus from analyzing intrinsic model behavior to evaluating model adaptation, opening a broad and orthogonal set of research questions, including the choice of supervision methods, fine-tuning strategies (e.g., supervised fine-tuning or reinforcement learning approaches), and data construction. Moreover, fine-tuning requires the selection of task-specific datasets, which may introduce additional biases and confound the analysis of what the model captures prior to adaptation. For these reasons, fine-tuning is intentionally outside the scope of this work and is left for future investigation.

We begin by outlining the \gls{co} setting, including the problems considered, the instance representations, the feature sets, and the algorithm portfolios (\Cref{sec:methodology:co-concepts}). We then describe the direct querying procedure (\Cref{sec:methodology:direct-querying}), followed by two probing analyses: the first focusing on the representation of problem features (\Cref{sec:methodology:probing-features}), and the second on information relevant to algorithm selection (\Cref{sec:methodology:probing-alg-sel}). The overall structure of our methodology is summarized in \Cref{tab:llm-behavior:method-overview}, which provides an overview of the specific objectives, approaches, and evaluation procedures associated with each experimental component.

\begin{table}[t]
    \centering
    \caption[Overview of methodological components]{Overview of the methodological components of this study. Each component addresses a specific research goal and employs a dedicated evaluation strategy.}
    \label{tab:llm-behavior:method-overview}
    \scriptsize
    \begin{tabular}{p{3cm}p{4.3cm}p{3.1cm}l}
    \toprule
    Goal & Methodology & Evaluation & Section \\
    \midrule
    Assess whether \glspl{llm} can explicitly identify problem features from instance representations & 
    \textbf{Direct Querying}: models are prompted to extract features directly from the instance & 
    Comparison between model-predicted and ground-truth features &
    \Cref{sec:methodology:direct-querying} \\
    \addlinespace
    Determine whether internal representations of \glspl{llm} encode problem features &
    \textbf{Feature Probing}: 
    regressors are trained on last-layer activations to predict instance features &
    Comparison between predicted and ground-truth features &
    \Cref{sec:methodology:probing-features} \\
    \addlinespace
    Determine whether internal representations of \glspl{llm} encode information predictive of algorithm performance &
    \textbf{Algorithm Selection Probing}: classifiers are trained on last-layer activations to predict the per-instance best algorithm &
    Classification accuracy, comparison with baseline methods &
    \Cref{sec:methodology:probing-alg-sel} \\
    \bottomrule
    \end{tabular}
\end{table}

\subsection{Combinatorial Optimization Concepts}
\label{sec:methodology:co-concepts}

We recall here the main concepts and definitions underlying \gls{co} to establish a common terminology and clarify how these notions are instantiated in this study.

Optimization problems aim to identify the best solution(s) from a search space, i.e., the set of feasible solutions, by optimizing one or more objective functions subject to a set of constraints. Each solution corresponds to an assignment of values to a collection of decision variables. In \gls{co}, these decision variables take discrete values, resulting in a finite, yet typically exponentially large, search space. Such a formulation encompasses both well-known benchmark problems, such as the seminal \gls{sat}, and a variety of real-world applications~\cite{DBLP:journals/heuristics/AhmetiM25,DBLP:journals/ai/KletzanderM24,DBLP:journals/cor/CeschiaGMM24}.

Each \gls{cop} defines a general problem structure that is instantiated through specific \emph{instances}, which provide the concrete input data. For example, the \gls{gcp} is defined on an undirected graph $G=(V,E)$, where $V$ denotes the set of nodes and $E$ the set of edges connecting pairs of nodes. The task is to assign a color to each node such that adjacent nodes receive different colors, yielding a valid coloring that minimizes the total number of colors used. An instance of the \gls{gcp} therefore specifies the graph topology, i.e., the nodes and edges that define $G$. Instances are typically stored in standardized formats and parsed into data structures suitable for algorithmic processing.

For example, \gls{gcp} instances are available in the DIMACS format,\footnote{\url{http://archive.dimacs.rutgers.edu/}, accessed 14 Oct 2025.} an abridged description of which is reported below:

\begin{tcolorbox}[colback=white,colframe=gray!75!black, fontupper=\scriptsize, enlarge bottom by=-1.5mm, enlarge top by=-2mm, boxsep=0mm]
In this format, nodes are numbered from 1 up to $n$.  There are $m$
edges in the graph.

Files are assumed to be well-formed and internally consistent: node identifier values are valid, nodes are defined uniquely, exactly $m$ edges are defined, and so forth. 

Comment lines give human-readable information about the file and are ignored by programs. Comment lines can appear anywhere in the file. Each comment line begins with a lower-case character c. There is one problem line per input file. 

The problem line must appear before any node or arc descriptor lines. The problem line has the following format: p FORMAT NODES EDGES. The lower-case character p signifies that this is the problem line. The FORMAT indicates the graph format. The NODES field contains an integer value specifying the number of nodes in the graph. The EDGES field contains an integer value specifying the number of edges in the graph.

There is one edge descriptor line for each edge in the graph, each with the following format: e $u$ $v$. The lower-case character e signifies that this is an edge descriptor line. For an edge $(u,v)$ the fields $u$ and $v$ specify its endpoints. Each edge $(u,v)$ appears exactly once in the input file and is not repeated as $(v,u)$. 
\end{tcolorbox}

An instance in this representation for the \gls{gcp} is illustrated below:

\begin{tcolorbox}[colback=white,colframe=gray!75!black, fontupper=\scriptsize, enlarge bottom by=-1.5mm, enlarge top by=-2mm, boxsep=0mm]
p edge 100 1902

e 1 2

e 1 6

e 1 10

\dots
\end{tcolorbox}

\noindent
In the remainder of this work, we refer to this standardized format as \standard{}. In addition to it, we also consider two alternative instance representations: natural language descriptions (\nlp{}) and code-like formulations derived from the MiniZinc modeling language (\code{}).
The same instance introduced above can be expressed in \nlp{} form as follows:

\begin{tcolorbox}[colback=white,colframe=gray!75!black, fontupper=\scriptsize, enlarge bottom by=-1.5mm, enlarge top by=-2mm, boxsep=0mm]
The instance is named {name}. The graph has 100 nodes and 1902 edges. There is an edge between node 1 and node 2. There is an edge between node 1 and node 6. There is an edge between node 1 and node 10. \dots
\end{tcolorbox}

The corresponding \code{} format is:\footnote{\url{https://www.hakank.org/minizinc/coloring_ip.mzn}, accessed 15 Oct 2025.}

\begin{tcolorbox}[colback=white,colframe=gray!75!black, fontupper=\scriptsize, enlarge bottom by=-1.5mm, enlarge top by=-2mm, boxsep=0mm]
int: n = \dots;

set of int: V = \dots;

int: num\_edges = \dots;

array[1..num\_edges, 1..2] of V: E = \dots;
\end{tcolorbox}

Problem instances can be further characterized through \emph{features}, i.e., measurable attributes that capture structural or statistical properties of an instance and are often problem-specific. Features enable comparisons across instances and support tasks such as \gls{isa} and algorithm selection.
In this study, features are considered in two complementary ways:
\begin{enumerate}
    \item \textbf{Feature extraction and interpretation}: we investigate whether, and to what extent, \glspl{llm} can identify or reconstruct relevant instance features directly from the instance representations. Features are grouped by their extraction complexity:\footnote{Extraction complexity is defined relative to the \gls{llm}, reflecting how difficult a feature is for the model to infer.}
    \begin{enumerate*}[label=\emph{(\roman*)}]
        \item \emph{directly extractable features}, which appear explicitly in the instance (\easyfeature{});
        \item \emph{low-effort features}, which require minimal computation, such as counts or extrema (\midfeature{}); and
        \item \emph{high-effort features}, which involve aggregation or derived calculations, such as averages (\hardfeature{}).
    \end{enumerate*}
    \Cref{tab:gcp-features} reports examples of features used for the \gls{gcp}.
    \item \textbf{Baseline for algorithm selection}: we use both the above features and additional descriptors defined in prior \gls{isa} studies \cite{DBLP:journals/eor/LiuSWC24,SmithMilesEtAl2014GCP,SmithMilesEtAl2009JSP,DBLP:journals/cor/Smith-MilesCM21} as inputs to traditional per-instance algorithm selection models. These baselines serve as reference points against which we compare models that instead rely on \gls{llm} hidden-layer representations.
\end{enumerate}

\begin{table}[t]
    \centering
    \caption{Graph Coloring Problem features.}
    \label{tab:gcp-features}
    \scriptsize
    \begin{tabular}{lp{8.5cm}c}
     \toprule
     Name & Description & Compl. \\
    \midrule
    \texttt{feat\_nodes} & The total number of nodes in the graph. & \easyfeature\\
    \texttt{feat\_edges} & The total number of edges in the graph. & \easyfeature\\
    \texttt{feat\_degree\_1} & The degree of node 1 in the graph, representing the number of connections per node. & \hardfeature\\
    \texttt{feat\_density} & The density of the graph, calculated as the ratio of 2 times the number of edges to the number of nodes times the number of nodes minus 1. & \hardfeature\\
    \texttt{feat\_ratio\_1} & The ratio of the number of nodes to the number of edges in the graph. & \hardfeature \\
    \texttt{feat\_ratio\_2} & The ratio of the number of edges to the number of nodes in the graph. & \hardfeature \\
    \texttt{feat\_degree\_mean} & The average degree of the nodes in the graph, representing the mean number of connections per node. & \hardfeature \\
    \texttt{feat\_degree\_max} & The maximum degree of the nodes in the graph, indicating the highest number of connections any node has. & \hardfeature \\
    \texttt{feat\_degree\_min} & The minimum degree of the nodes in the graph, indicating the lowest number of connections any node has. & \hardfeature \\
    \bottomrule
    \end{tabular}
\end{table}

\glspl{cop} are typically solved by \emph{algorithms} that explore the search space to identify high-quality solutions. These include \emph{exact methods}, which guarantee optimality but are often computationally prohibitive, and \emph{(meta-)heuristic methods}, which trade optimality for efficiency and scalability. In this study, we adopt an \emph{algorithm portfolio} perspective, considering sets of algorithms with complementary performance across instances. Algorithmic performance is evaluated by comparing the objective values obtained on each instance.

For illustration, in the case of the \gls{gcp}, we consider two graph-coloring heuristics, DSATUR (degree saturation) and MAXIS (maximal independent set), which have been demonstrated to exhibit complementary strengths and weaknesses through \gls{isa} \cite{DBLP:journals/cor/Smith-MilesB15}. The performance metric is the raw objective value, namely the number of colors in a valid coloring, computed on the same set of benchmark instances used in that study~\cite{DBLP:journals/cor/Smith-MilesB15}.

\begin{figure}
    \centering
    \includegraphics[width=1\linewidth]{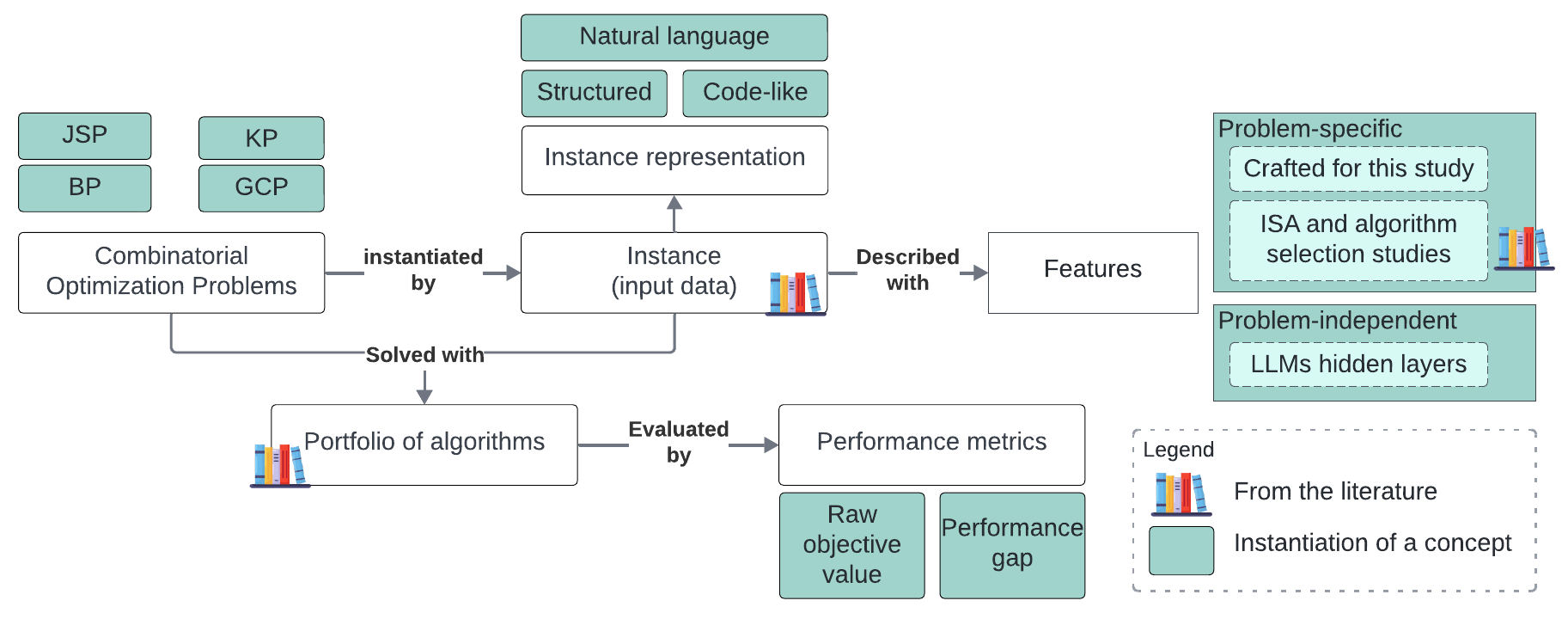}
    \caption[Optimization concepts]{Conceptual structure of the case studies and their instantiation in this work.}
    \label{fig:opt-concepts}
\end{figure}

\Cref{fig:opt-concepts} provides a schematic overview of this methodological framework, illustrating how the main \gls{co} concepts are organized and instantiated across the case studies considered in this work. Specifically, for each problem, we retrieve benchmark instances from prior \gls{isa} studies to ensure diversity in structural characteristics. Each instance is expressed through different representations, either structured (e.g., \standard{} or \code{}) or natural language (\nlp{}). Moreover, each instance is characterized by two complementary types of features. The first are \emph{handcrafted features}, conceived, designed, and extracted through manually written code, as in traditional algorithm selection analyses. The second are \emph{automatically derived features}, obtained directly from \gls{llm} hidden-layer activations without the need for explicit feature engineering. Quantitative characterization of the instances relies on these feature sets, while algorithm portfolios comprise methods identified in the literature as exhibiting complementary performance, evaluated through standard metrics such as raw objective value and performance gap.

\Cref{tab:problem-summary} summarizes the main characteristics of the case studies considered in this work. We selected four \glspl{cop} that cover different combinatorial structures and search spaces (assignments, permutations, and graphs): namely, the \gls{bp}, \acrfull{gcp}, \gls{jssp}, and \gls{kp}. These are well-established benchmark problems and are therefore not described here in detail. The corresponding instances, together with their \gls{isa} features and algorithm performance data used for algorithm selection, are available on the MATILDA\footnote{\url{https://matilda.unimelb.edu.au/matilda/}, accessed 13~Oct~2025.} website.
For each \gls{cop}, we report the number of benchmark instances, the number of algorithms included in the portfolio, and the number of features considered (--- means no feature of that type). The column \textsc{ISA} refers to the study from which the instances, algorithm portfolios, and the features labeled as \gls{isa} were originally derived. The columns \textsc{Standard desc.} and \textsc{Code-like} indicate the source of the standardized and code-like representations, respectively. 

\begin{table}[ht]
    \centering
    \caption{Summary of the considered problems.}
    \label{tab:problem-summary}
    \resizebox{1\textwidth}{!}{%
    \begin{tabular}{lrrp{4.6cm}rrrrrlcc}
    \toprule
    Problem & Instances & \multirow{2}{*}{\makecell[c]{Algor-\\ithms}} & \multicolumn{6}{c}{Features for the direct querying and probing extraction} & ISA & \multirow{2}{*}{\makecell[c]{Standard\\ desc.}} & \multirow{2}{*}{\makecell[c]{Code\\-like}}\\
    \cmidrule(lr){4-9}
     & (selected) && List & \multicolumn{2}{c}{Type}  & \multicolumn{3}{c}{Compl.} \\
     \cmidrule(lr){5-6} \cmidrule(lr){7-9}
     &&&& Int & Float & \easyfeature{} & \midfeature{} & \hardfeature{} \\
    \midrule
    BP & 8,815\,(440) & 2 & Bin capacity, items, maximum/minimum/average item weight & 4 & 1 & 2 & 2 & 1 & \cite{bpp-thesis} &  &\\
    \addlinespace
    GCP & 8,278\,(413)& 2 & Nodes, edges, graph density, ratio of the number of nodes to the number of edges in the graph and vice versa, degree of the first node, maximum / minimum / average degree & 5 & 4 & 2 & --- & 7 & \cite{DBLP:journals/cor/Smith-MilesB15} & \cite{dimacs} &  \\
    \addlinespace
    JSP & 4,467\,(223) & 12 & Jobs, machines, operations, maximum / minimum / average duration & 5 & 1 & 2 & 3 & 1 & \cite{DBLP:journals/cor/StrasslM22} &  & \cite{minizinc-archive}\\
    \addlinespace
    KP & 5,300\,(265) & 3 & Capacity, maximum / minimum / average weight/profit, weight / profit / efficiency of the first item, average efficiency & 7 & 4 & 1 & 7 & 3 & \cite{DBLP:journals/cor/Smith-MilesCM21} & & \cite{minizinc-archive}\\
    \bottomrule
    \end{tabular}
    }%
\end{table}

\subsection{Direct Querying of Large Language Models}
\label{sec:methodology:direct-querying}

Direct querying is employed to assess the extent to which LLMs can identify or describe relevant characteristics of \glspl{cop} based on given instance representations. This approach focuses on externally observable behavior. 
In this study, we consider two prompt variants within the direct-querying setting and three overall querying configurations. The prompt variants are \emph{standard direct querying}, also referred to as zero-shot querying, in which the model is asked to return only the target numeric value, and \emph{\glsdesc{cot} direct querying}, in which the prompt explicitly requests a short reasoning trace before the final structured output. In addition, for \gpt models we consider a \emph{native reasoning} configuration implemented through the Harmony interface: in this case, the model receives the same user-facing prompt as in the standard variant, but reasoning is produced through the model's native analysis channel under a fixed developer instruction, a controlled reasoning-effort setting, and the same structured-output constraints.

Prompts are constructed according to the \gls{rtf} framework to ensure consistent task framing and output structure across all models. Each prompt comprises three components:
\begin{itemize}
\item \textbf{Reasoning preamble:} defines the assumed role/domain expertise;
\item \textbf{Task description:} specifies the reasoning objective or operation to be performed;
\item \textbf{Format specification:} constrains the response to a predefined structure, ensuring syntactic validity and facilitating automated evaluation.
\end{itemize}

Both prompt variants share the same problem statement, feature specification, and instance input, and differ only in the output instructions. The standard variant asks the model to return the target value directly in JSON format, whereas the Chain-of-Thought variant asks for a short reasoning trace followed by the same target value. Native reasoning in \gpt models uses the same user-facing prompt as the standard variant, but differs in the interaction protocol and in the availability of an internal reasoning channel. This design allows us to isolate the effect of explicit intermediate reasoning while keeping the task specification unchanged.

The standard direct-querying prompt template used in this study is presented below. It shows how the reasoning preamble, task description, and format specification are combined.

\begin{tcolorbox}[colback=white,colframe=gray!75!black, fontupper=\scriptsize, enlarge bottom by=-1.5mm, enlarge top by=-2mm, boxsep=0mm]
\begin{verbatim}

You are given an instance of a combinatorial optimization problem: {problem_name}.
  This is not a coding task. Do not return any code.
  Extract the numeric value of the following feature from the instance:
  - Feature name: {feature_name}
  - Feature description: {feature_description}
  - Expected type: {feature_type}
  The instance is provided here: """
  {instance}
  """
  Instructions:
  - Return a JSON object only, with a single field "value", i.e., `{{ "value": ... }}`.
  - The "value" field should contain the numeric value of the required feature 
    and should be of the expected type (i.e., {feature_type}).
  - If the value is unknown/undeterminable, return `{{ "value": null }}`.
  - No explanations, no extra fields.
  Answer:
\end{verbatim}
\end{tcolorbox}

The Chain-of-Thought prompt variant retains the same task description and instance input, but modifies the instructions so that the model first provides a short explanation of how the value is derived from the raw problem data and then returns the target value in JSON format. The corresponding template is reported below:

\begin{tcolorbox}[colback=white,colframe=gray!75!black, fontupper=\scriptsize, enlarge bottom by=-1.5mm, enlarge top by=-2mm, boxsep=0mm]
\begin{verbatim}
You are given an instance of a combinatorial optimization problem: {problem_name}.
  This is not a coding task. Do not return any code.
  Extract the numeric value of the following feature from the instance:
  - Feature name: {feature_name}
  - Feature description: {feature_description}
  - Expected type: {feature_type}
  The instance is provided here: """
  {instance}
  """
  Note: the instance contains only raw problem data — feature names do not appear
  as labels inside it. You must derive the value by reading and interpreting the
  data structure directly.
  Instructions:
  - Before returning the JSON, reason step by step (2-4 sentences) about how you
    identified the value from the raw data structure, not from any label or name.
  - Return a JSON object with exactly two fields:
    - "reasoning": your concise step-by-step explanation.
    - "value": the numeric value of the required feature, of type {feature_type}.
  - If the value is unknown or undeterminable, set "value" to null.
  - No extra fields beyond "reasoning" and "value".
  Answer:
\end{verbatim}
\end{tcolorbox}

For \gpt models, direct querying is implemented through the Harmony message format, using three messages: a system message specifying the reasoning effort, a fixed developer instruction enforcing precise JSON-only extraction, and a user message containing the same task specification used in the standard direct-querying prompt. This setup enables native reasoning while preserving the same user-facing task formulation adopted for the other models.

Additionally, the \gls{llm} is equipped with a structured decoding layer,\footnote{\url{https://blog.vllm.ai/2025/01/14/struct-decode-intro.html}, accessed 17 Oct 2025.} which guides the generation process and constrains the output to follow a predefined JSON schema. This mechanism ensures syntactic validity and type consistency with the feature being predicted. Two schemas are used, one for each prompting variant. In the standard direct-querying setting, the output contains a single field \texttt{value}. In the Chain-of-Thought setting, the output contains exactly two fields, \texttt{reasoning} and \texttt{value}. In both cases, evaluation is performed exclusively on the \texttt{value} field, ensuring direct comparability between the two prompting variants.

\begin{tcolorbox}[
  colback=white,
  colframe=gray!75!black,
  fontupper=\scriptsize\ttfamily,
  enlarge bottom by=-1.5mm,
  enlarge top by=-2mm,
  boxsep=1mm,
  left=1mm,
  right=1mm
]
\begin{verbatim}
{
  "type": "object",
  "additionalProperties": false,
  "required": ["value"],
  "properties": {
    "value": { "anyOf": [ { "type": "[feature-type]" }, { "type": "null" } ] }
  }
}
\end{verbatim}
\end{tcolorbox}

For the Chain-of-Thought variant, the schema is added with the additional required field \texttt{\small "reasoning":\{"type":"string"\}}.

The correctness of model-generated features is evaluated by comparing them against ground-truth values computed from the corresponding instances, using both exact-match and error-based criteria. As for the Chain-of-Thought variant, only the numeric \texttt{value} field is considered in the quantitative evaluation; the \texttt{reasoning} field is retained as a behavioral artifact of the prompting strategy, but is not scored directly.

\begin{figure}
    \centering
    \includegraphics[width=1\linewidth]{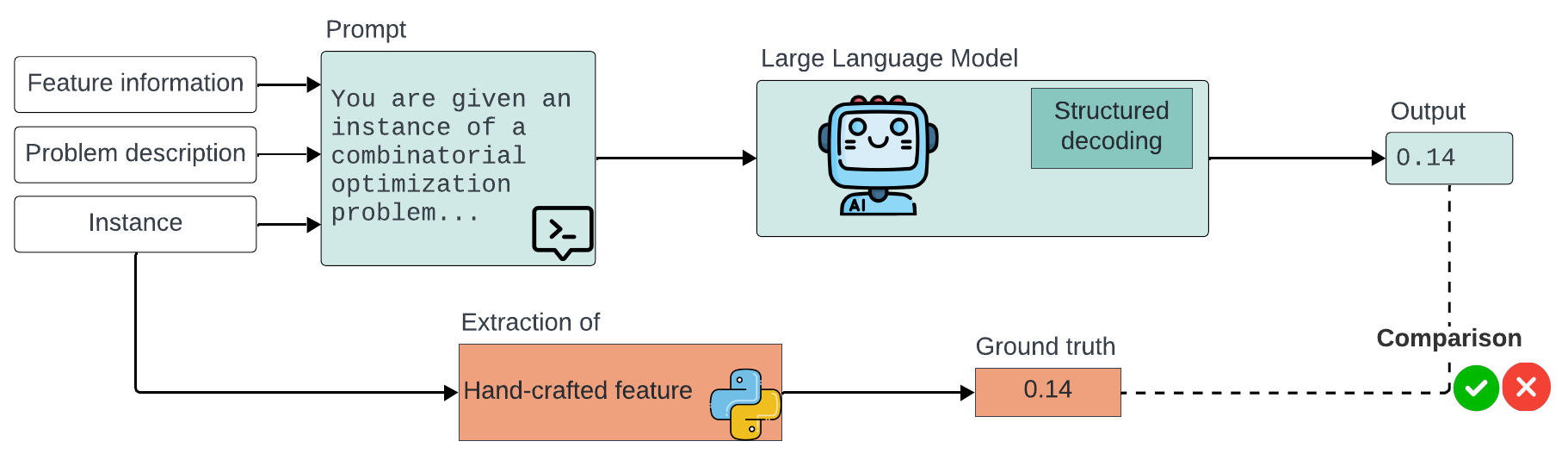}
    \caption{Overview of the direct querying pipeline. }
    \label{fig:direct-querying}
\end{figure}

The overall setup of the direct-querying experiment is depicted in \Cref{fig:direct-querying}. For each problem, the \gls{llm} receives a prompt reporting an instance in one of three formats together with a textual instruction specifying the target feature. Depending on the prompt variant and execution mode, the model either returns the target value directly (standard direct querying), produces a short reasoning trace followed by the same target value (Chain-of-Thought direct querying), or generates the final answer through a native reasoning channel while preserving the same user-facing task specification (native reasoning with \gpt models). The direct-querying procedure is repeated for all features defined for each problem. Predicted feature values are then compared with ground-truth values obtained by running handcrafted feature extractors on the same instance.

Generation budgets are adjusted to the prompt variant and to the underlying interaction format: 64 tokens for standard direct querying with instruct models, 512 for Chain-of-Thought prompting, and 1,024 for native reasoning. For \gpt models, a minimum budget of 512 tokens is enforced even in the standard setting, since the Harmony interface opens an internal analysis channel before the final JSON output.

\subsection{Probing of Large Language Models}
\label{sec:methodology:probing-features}

Probing is employed to investigate whether \glspl{llm} encode information about the structural and numerical characteristics of \gls{co} instances. In contrast to direct querying (\Cref{sec:methodology:direct-querying}), which analyzes externally observable behavior, probing examines the latent representations learned by the model, providing a complementary perspective on how instance properties are internally encoded.

For each combination of problem and instance representation, every instance is processed by the model without triggering text generation. Hidden-layer activations are extracted from the final layer, and a single vector representation for each instance is obtained by applying three common pooling strategies: \emph{mean}, \emph{max}, and \emph{last} (see, e.g., \citet{tang2024poolingattentioneffectivedesigns}). The \emph{mean} strategy computes the average of the last hidden-layer activations across all tokens; the \emph{max} strategy takes the maximum activation along each representation dimension across tokens; and the \emph{last} strategy uses the activation vector corresponding to the \gls{eos} token, i.e., the model final prediction signal indicating the termination of generation. These pooled vectors provide compact, contextualized embeddings of each instance, which are then used as input to downstream probing models to assess whether the model internally encodes instance-level structural and numerical characteristics.


Each pooled vector is then paired with the ground-truth feature values corresponding to the same instance, forming a dataset for supervised probing. For each feature and representation, we train a separate regression model (probe) to assess whether the feature can be accurately decoded from the hidden representations, thereby quantifying the extent to which such information is linearly accessible within the model's latent space. 

To assess both the accessibility and the complexity of the information encoded in the model representations, we employ three types of probes, covering a range of functional families from fully linear to boosted non-linear.

Since the hidden-state dimensionality exceeds the number of training instances on the $5\%$ subset, the linear and gradient-boosting probes operate on a shared preprocessing stack: standardization, removal of zero-variance components, and a principal-component projection to $\min(128, n_{\mathrm{train}}\!-\!1)$ components without whitening. The same preprocessing is adopted for the algorithm-selection probing of \Cref{sec:methodology:probing-alg-sel}.

\begin{itemize}
    \item \textbf{Linear probe:} a Ridge regressor ($\alpha = 10$) applied to the PCA-preprocessed  representation.
    High performance with this probe indicates that the feature is linearly decodable via a penalized linear combination of the principal components of the hidden state.
    
    \item \textbf{Simple probe:} a one-hidden-layer \gls{mlp} with
    $128$ hidden units, \gls{relu} activation, dropout
    ($p = 0.2$), and weight decay ($10^{-4}$), trained with Adam
    and early stopping (patience $5$, restoring the best weights)
    on a $20\%$ internal validation split. It operates directly
    on the raw pooled hidden state, since its hidden layer provides
    a learned compression of the input.
    
    \item \textbf{Complex probe:} a \lightGBM{} regressor
    ($200$ estimators, learning rate $0.05$, $\ell_2$ penalty $1.0$,
    minimum $20$ samples per leaf, feature subsampling $90\%$)
    applied to the same PCA-preprocessed representation. It is used to assess whether additional model capacity beyond the linear family yields substantial accuracy gains, which would suggest that the information is encoded in a non-linearly separable manner within the representation.
\end{itemize}
\noindent
The evaluation protocol mirrors that adopted for direct querying (\Cref{sec:methodology:direct-querying}), comparing predicted and ground-truth feature values through exact-match and error-based metrics on the same validation instances. The comparison of primary interest is between probing and direct querying, not between probe families; per-probe and per-pooling results are reported in \ref{app:feature-probing-complete-results}.

Notice that we interpret probing performance as evidence that a target property is decodable from the hidden representation under the chosen probe family. This should not be read as proof that the \gls{llm} explicitly computes the corresponding feature or internally implements the deterministic extraction procedure. In particular, probes may exploit feature-related correlations present in the instance distribution. For this reason, probing is used here as a complementary diagnostic of latent signal, not as a standalone claim of mechanistic understanding.


\subsection{Algorithm Selection using Large Language Models}
\label{sec:methodology:probing-alg-sel}

We further extend the probing analysis to examine whether the information encoded in the pooled layers of \glspl{llm} can support downstream tasks, specifically per-instance algorithm selection. In this setting, probing is formulated as a classification problem, where the objective is to identify, for each instance, the algorithm expected to achieve the best performance based on its latent representation.

The setup for this experiment mirrors that of the feature-probing analysis, employing the same probe types, with the key distinction that the target variable is now categorical—that is, the label of the algorithm achieving the best performance—rather than continuous. Each \gls{llm}-derived representation is paired with the ground-truth label of the best-performing algorithm for the corresponding instance, forming a dataset for supervised classification. A classifier probe is then trained to determine whether the identity of the winning algorithm can be decoded from the model hidden representations.

Performance is evaluated in terms of \emph{set-aware accuracy}, which accounts for instances where multiple algorithms achieve identical best performance (ties). During training, one of the tied algorithms is randomly selected as the target label, whereas during evaluation, a prediction is considered correct if the classifier selects any of the algorithms attaining the best performance for that instance. Formally, given $n$ instances and for each instance $i$ a set of acceptable labels $S_i$, set-aware accuracy is defined as:
\begin{equation}
\mathrm{Acc}_{\text{set}}
= \mathbb{E}_{(x_i, S_i)} \big[\, \mathbf{1}\{ \hat{y}(x_i) \in S_i \}\, \big]
\end{equation}
where $\hat{y}(x_{i})$ denotes the predicted (single) label for instance $i$, described by the features $x_i$, and $\mathbf{1}\{{\cdot}\}$ is the indicator function. $\mathbb{E}$ is the expected value. A prediction is thus counted as correct if it matches any of the true \emph{winning} algorithms for that instance.

To preserve balanced distributions across instances with different sets of winning algorithms, stratified sampling is performed over the powerset of possible labels (excluding the empty set). Each subset of algorithms defines a stratum, ensuring similar proportions of label sets in the training and evaluation splits.

To contextualize the results, we compare \gls{llm}-derived embeddings with three established baselines:
\begin{itemize}
    \item \textbf{Most frequent winner:} a naive baseline that always predicts the algorithm most frequently achieving the best performance in the training set.
    \item \textbf{ISA-based classifiers:} classifiers trained on instance features derived from previous \gls{isa} studies.
    \item \textbf{Handcrafted feature classifiers:} classifiers trained on the handcrafted features used in the preceding experiments.
\end{itemize}
All representation families, namely \gls{llm}-derived embeddings, handcrafted features, and ISA-based features, are evaluated using the same downstream classifier families and the same data partitions. This protocol isolates the effect of the input representation from differences due to classifier capacity or train/test variability.

Classifier performance is evaluated under a stratified $k$-fold cross-validation protocol.
The downstream classifiers considered in this setting are \mostFrequent{}, \logisticRegression{}, \mlpClassifier{}, and \lightGBM{}, covering a naive baseline, linear decision models, and more flexible non-linear predictors.
All classifiers are trained and tested on identical folds, ensuring paired comparisons across representations and models.
The same partitions are then reused in the subsequent mixed linear analysis to assess whether \gls{llm}-derived representations achieve predictive power comparable to traditional \gls{isa}-based and handcrafted features.



\begin{figure}
    \centering
    \includegraphics[width=1\linewidth]{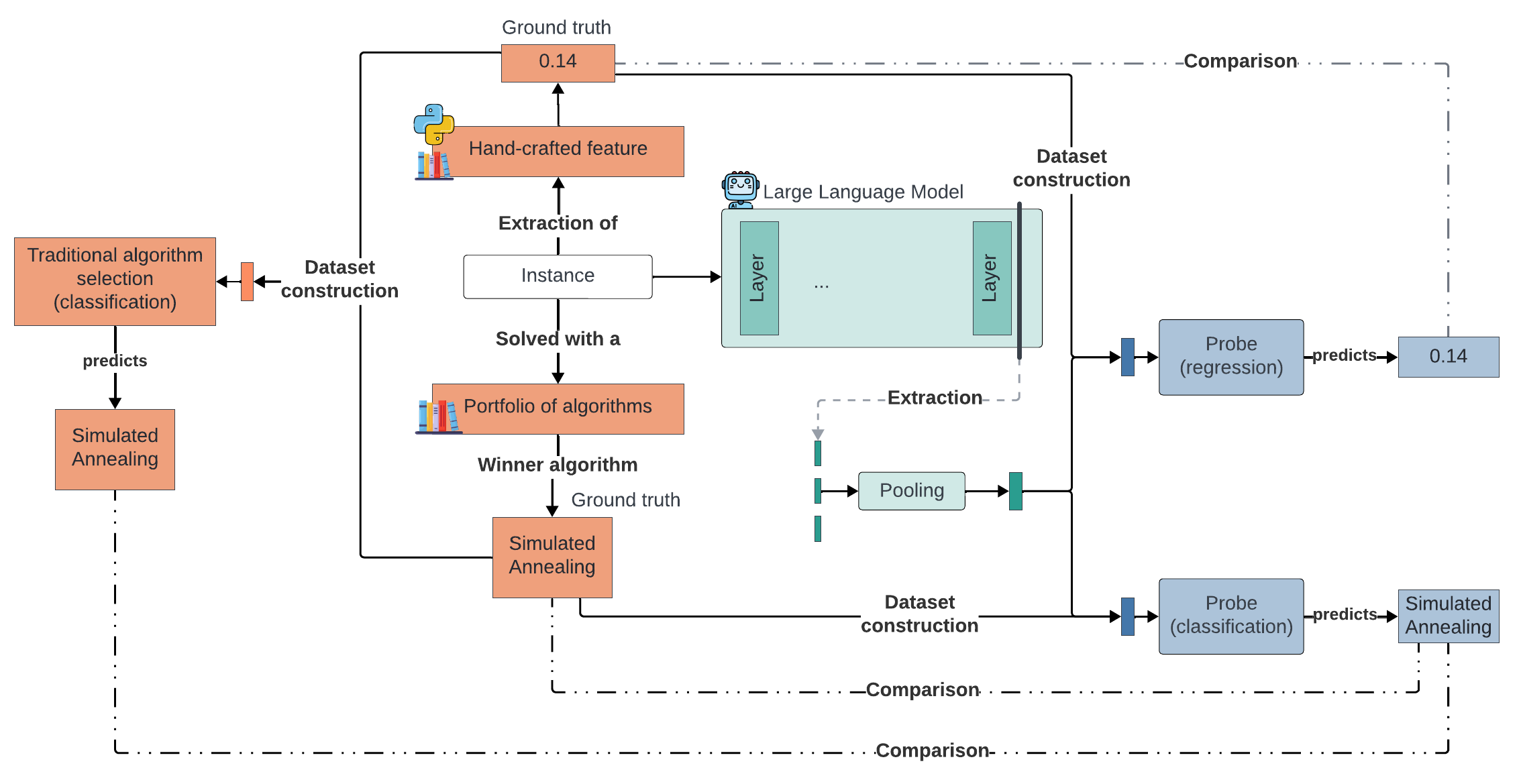}
    \caption{Overview of the probing process.}
    \label{fig:probing}
\end{figure}

The overall probing workflow is illustrated in \Cref{fig:probing}. Each instance is processed by the \gls{llm}, from which hidden representations are extracted and pooled to obtain fixed-length embeddings. These embeddings are paired with ground-truth labels to construct datasets for regression and classification probing. The regression probes test whether instance features can be decoded from the latent space, while the classification probes assess whether the same representations contain sufficient information to support downstream algorithm selection.

\subsection{Experimental Setup}
\label{sec:experimental-setup}

The experimental campaign is conducted on five open-weight \glspl{llm} spanning the $3$B--$120$B parameter range: \llama[3.2][3B], \llama[3.1][8B], \llama[3.3][70B], \gpt{20b}, and \gpt{120b}. This model pool allows us to validate the robustness of the observed patterns across both instruct-style and reasoning-oriented open-weight families, while retaining full reproducibility of the experimental artifacts. Closed, commercial frontier models are not included for the reasons already discussed in \Cref{sec:introduction}.

All experiments are conducted on a stratified $5\%$ subset of the original benchmark collection. The subset is constructed independently for each problem by preserving the distribution of the downstream winner-label sets used in the algorithm-selection task and by maintaining coverage of the feature-complexity groups and instance representations used in the feature-extraction analyses. The resulting subset contains 440 BPP, 413 GCP, 223 JSP, and 265 KP instances, as reported in \Cref{tab:problem-summary}. All subsequent experiments use this same subset to ensure paired comparisons across querying, probing, and algorithm-selection analyses. Stratification is performed before model evaluation and is kept fixed across all models, prompting modes, pooling strategies, and downstream classifiers.

\begin{figure}
    \centering
    \includegraphics[width=\linewidth]{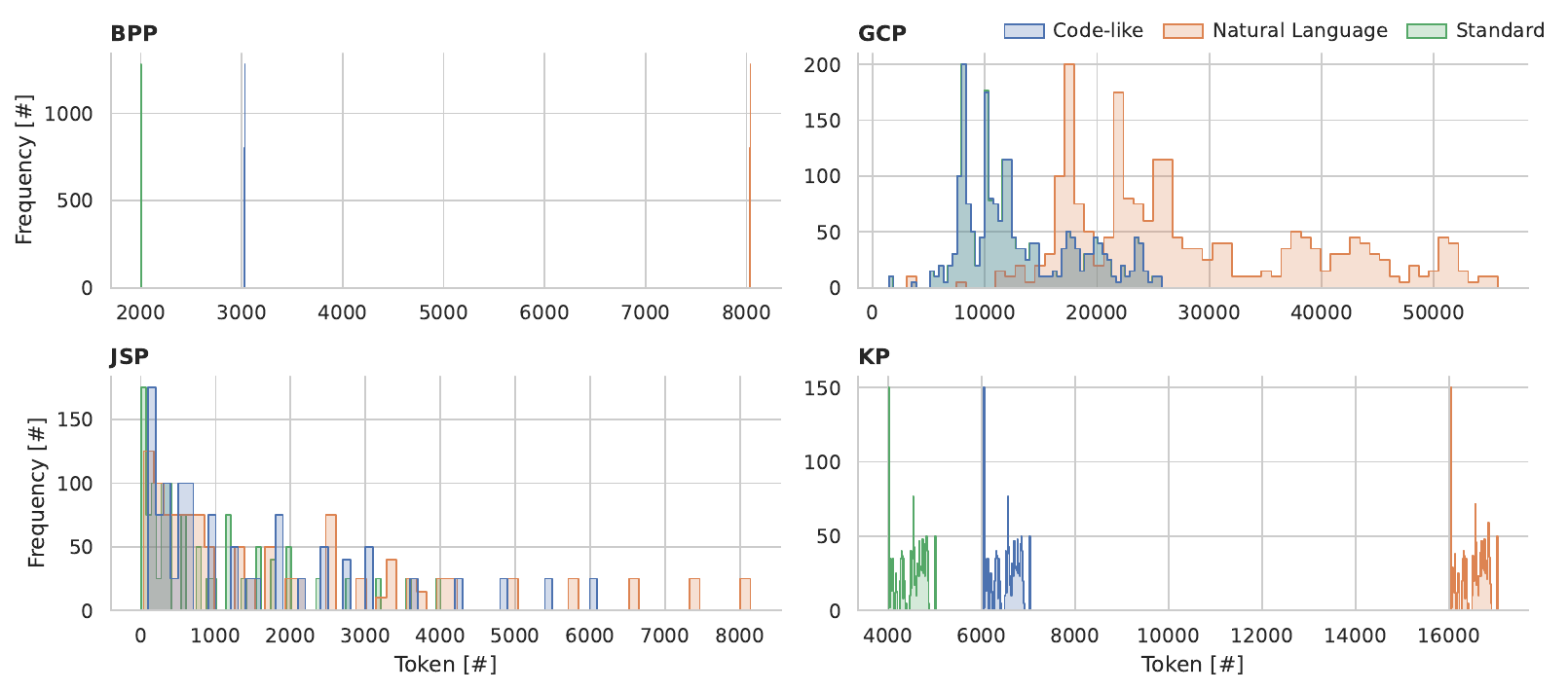}
    \caption{Number of tokens per problem and instance representation on the stratified $5\%$ subset.}
    \label{fig:number-tokens}
\end{figure}

\Cref{fig:number-tokens} reports the token distributions for the four \glspl{cop} on the stratified $5\%$ subset, distinguishing the three instance representations. Sequence length varies markedly across problems and formats: BPP and JSP remain relatively compact, KP is intermediate, and GCP is the longest and most variable case, especially in natural language. As expected, across all problems, natural-language inputs are consistently longer than code-like and standard representations. Nevertheless, all observed lengths remain within the context limits of the selected model pool, so no truncation is required.

\begin{table}[t]
\centering
\small
\caption{Summary of the evaluated model pool and supported direct-querying modes. For \gpt models, the user-facing standard prompt is executed through native reasoning via Harmony.}
\label{tab:model-pool}

\begin{tabular}{llrp{0.22\linewidth}}
\toprule
Model & Family & Parameters & Direct-querying modes \\
\midrule
\llama[3.2][3B] & \llama instruct & 3B & standard, Chain-of-Thought \\
\llama[3.1][8B] & \llama instruct & 8B & standard, Chain-of-Thought \\
\llama[3.3][70B] & \llama instruct & 70B & standard, Chain-of-Thought \\
\gpt{20b} & \gpt reasoning & 20B & native reasoning \\
\gpt{120b} & \gpt reasoning & 120B & native reasoning \\
\bottomrule
\end{tabular}

\end{table}

The models are accessed via the \huggingFace{} interface and executed using the \vllm{} library, which provides improved robustness and efficiency over the standard \transformers{} implementation. \vllm{} enables effective parallelization of both text generation in direct querying and hidden-state extraction in probing, allowing efficient processing of large batches and long input contexts. For \gpt models, direct querying is implemented through the Harmony message format, which supports native reasoning via a dedicated analysis channel while preserving the same feature-extraction task specification described in \Cref{sec:methodology:direct-querying}.

To ensure reproducibility, all experimental data and configurations are managed with \dvc, while the source code is version-controlled through \git. Experiments are executed on an NVIDIA DGX Station equipped with four NVIDIA A100 GPUs (80~GB each). Post-processing tasks, including regression and classification analyses, are carried out offline on a MacBook Pro (Apple M4, 32~GB RAM, 4+6 cores). Random seeds, problem sets, and replicate blocks are fixed across all runs to ensure consistent experimental conditions. The complete \git{} repository, including configuration files, experimental scripts, and processing pipelines, will be released as an open repository upon paper acceptance.

The probing models are implemented using \texttt{scikit-learn}, with \texttt{Linear\-Re\-gres\-sion} for regression, \texttt{MostFrequent\-Classifier} and \texttt{Logistic\-Re\-gres\-sion} for classification, and \texttt{pytorch} with the \texttt{skorch} adapter for the \gls{mlp} probes applied to both tasks. Furthermore, \lightGBM{} is used for both regression and classification. The \texttt{scikit-learn} framework was selected for its interoperability and consistency, allowing a unified experimental pipeline across all probe types. An ad hoc adapter class was developed for the set-aware classification task to ensure the correct computation of its evaluation metric within the same workflow.

Regarding the algorithm portfolios, the results are drawn from the original \gls{isa} studies (also referenced in \Cref{tab:problem-summary}), together with the corresponding instances and features labeled as \gls{isa} features. The ground-truth features employed in the experiments presented in \Cref{sec:methodology:probing-features,sec:methodology:direct-querying} are computed by dedicated feature-extraction scripts developed in \python{}. The handcrafted feature extractors are included in the project code base and integrated within a data preparation pipeline that processes the benchmark instances and computes the relevant features for the downstream \gls{llm}-related experiments.

The experiments are organized as follows. In the direct-querying setting, each problem instance in the stratified subset is paired with each target feature and queried under the selected prompt variant and execution mode. In the probing setting, only the instance representation is provided to the network; the last hidden state is extracted, pooled, and used to train regression probes on a train--validation split. In the algorithm-selection setting, both pooled hidden states and manually designed features (handcrafted or ISA-based) are used as classifier inputs under a stratified $k$-fold cross-validation protocol. This design yields a large but computationally tractable number of \gls{llm} runs and downstream learning experiments, while preserving consistent comparisons across models, prompting choices, and downstream baselines.

\section{Experimental Analysis}
\label{sec:experimental-analysis}

This section reports the results of the experimental analysis conducted to evaluate the behavior and representational properties of \glspl{llm} when applied to \glspl{cop}. The experiments follow the methodology introduced in \Cref{sec:methodology}, encompassing three complementary analyses: direct querying, feature probing, and algorithm selection probing. Each analysis focuses on a distinct aspect of model behavior: respectively, explicit feature identification, implicit feature encoding, and predictive modeling of algorithmic performance.

\subsection{Direct Querying of Problem Features}
\label{sec:direct-querying-experiment}

We begin the experimental analysis by examining the behavior of \glspl{llm} in the direct querying setting, where models are explicitly prompted to extract numerical features from problem instances. Performance is assessed using multiple metrics: \gls{mae}, computed as the average absolute deviation from the ground truth; the proportion of exact matches (\textsc{Equals}); and the proportions of predictions within 1\% (\textsc{Within 1\%}) and 5\% (\textsc{Within 5\%}) deviation from the ground truth values. These latter metrics provide a more sensitive view of near-accurate predictions, complementing the \gls{mae} analysis. In the following, we focus primarily on the \textsc{Within 1\%} metric for clarity. Robustness checks on \gls{mae}, exact-match accuracy and \textsc{Null rate} are reported in \ref{app:scaling-regression}, and the full set of metrics is available in the companion repository.

\subsubsection{Effect of feature complexity and model scale}

\begin{figure}
    \centering

    \includegraphics[width=\textwidth,trim=0 30pt 0 50pt,clip]{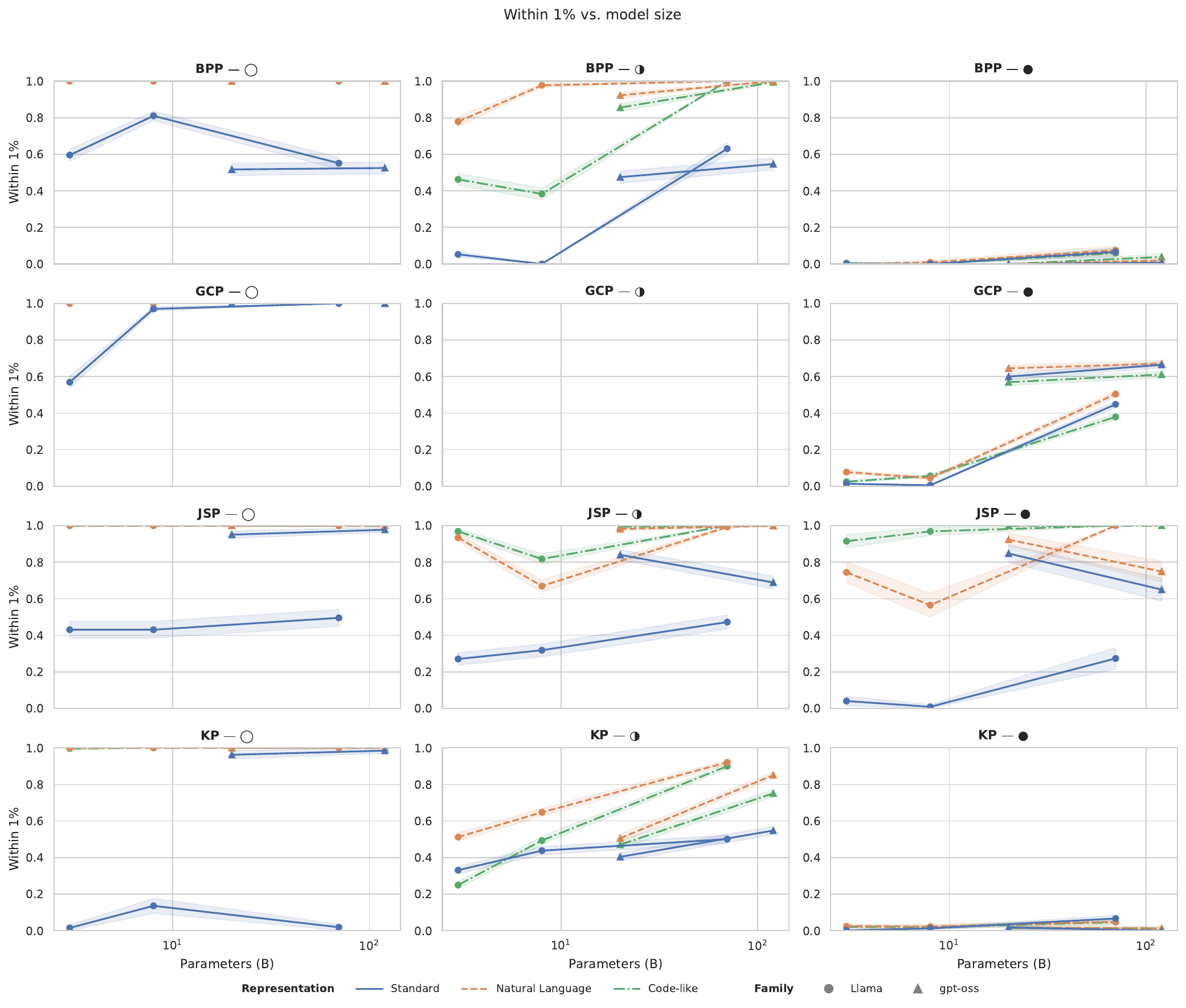}

    \caption{
    Direct-querying performance as model size increases, measured as the proportion of predictions within 1\% of the ground truth. 
    Each panel corresponds to a feature-complexity group. 
    Results correspond to standard direct querying for \llama models and native reasoning for \gpt models.
    }
    \label{fig:scaling-within-1pct}
\end{figure}

\Cref{fig:scaling-within-1pct} shows direct-querying performance as a function of model size across feature-complexity groups. Across all problems, feature complexity is the dominant factor: \easyfeature{} features are recovered reliably, while performance degrades substantially for \midfeature{} and \hardfeature{} features.

Model scaling has a non-uniform impact. For \easyfeature{} features, increasing model size yields only marginal improvements, as performance is already near saturation. For \midfeature{} features, scaling leads to clear gains, particularly for natural-language and code-like representations. For \hardfeature{} features, improvements remain limited, indicating that increased capacity alone does not fully compensate for the need for structured computation.

The same scaling pattern emerges when the analysis is repeated with alternative response variables capturing prediction error, namely normalized \gls{mae} and exact-match accuracy: normalized \gls{mae} slopes are negative across all non-degenerate cells, indicating that the average error shrinks by approximately $40$--$60\%$ for each parameter doubling on the \midfeature{} and \hardfeature{} groups, and exact-match slopes are positive with comparable goodness of fit. 
Per-metric regression coefficients are reported in \ref{app:scaling-regression} (\Cref{tab:scaling-regression-summary} on page \pageref{tab:scaling-regression-summary}).

\subsubsection{Effect of representation}

Representation plays a systematic but non-uniform role. The standard representation is frequently the least effective, while natural-language and code-like inputs generally yield higher accuracy. However, their relative ordering is not consistent across problems and feature-complexity groups. 

In particular, while natural-language representations often provide the strongest performance for \easyfeature{} features, code-like representations can match or outperform them in other settings, especially as feature complexity increases. This suggests that \glspl{llm} benefit from representations that expose semantic structure or explicit relational patterns, rather than from rigid encodings such as the standard representation.

\subsubsection{Effect of prompting and reasoning}

\begin{figure}
    \centering
    \includegraphics[width=\linewidth,trim=0 20pt 0 40pt,clip]{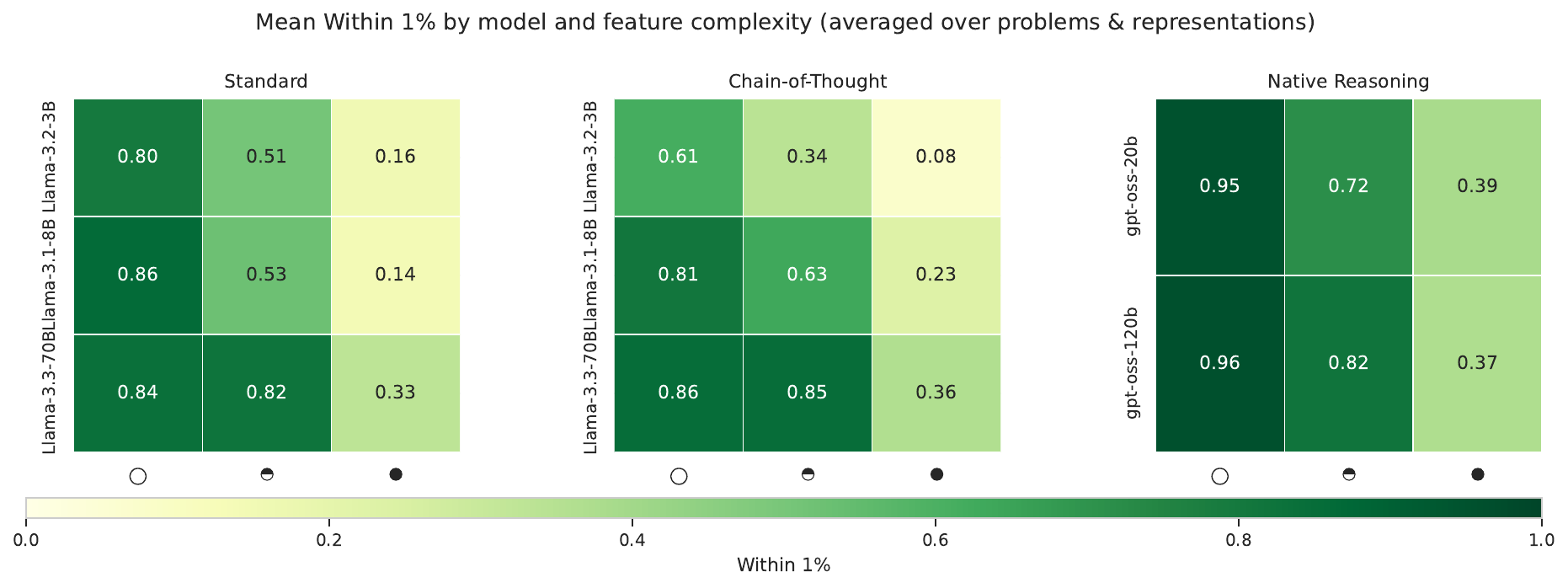}
    \caption{
    Effect of Chain-of-Thought prompting on direct-querying performance, shown as the change in mean Within 1\% accuracy relative to standard direct querying. Results are shown across model sizes, reasoning modes, and feature-complexity groups.
    }
    \label{fig:cot-effect}
\end{figure}

\Cref{fig:cot-effect} provides a detailed view of the impact of \gls{cot} prompting on direct-querying performance across model scales and feature-complexity groups.

The effect of \gls{cot} is strongly scale-dependent and non-monotonic. For the smallest model (3B), \gls{cot} consistently degrades performance across all feature types, suggesting that explicit reasoning introduces additional overhead without improving the model's ability to extract the correct value. In this regime, the model appears unable to effectively leverage intermediate reasoning steps.

For intermediate-scale models (8B), the effect becomes selective. \gls{cot} provides improvements primarily for \midfeature{} and \hardfeature{} features, where some degree of decomposition is beneficial, while offering little or no advantage for \easyfeature{} features. This indicates that \gls{cot} can act as a useful scaffold for moderately complex reasoning, but only when the underlying model capacity is sufficient.

For larger models (70B), \gls{cot} consistently improves performance across all feature-complexity groups. At this scale, explicit reasoning no longer acts as a burden, but instead helps organize the computation, leading to more reliable feature extraction.

These results suggest that \gls{cot} does not have a uniform effect, but interacts critically with model scale. Below a certain capacity threshold, it can hinder performance; at intermediate scales, it provides targeted benefits; and at larger scales, it becomes broadly advantageous.
Equivalent heatmaps computed on \gls{mae} and exact-match accuracy show the
same scale-dependent shape and are made available in the companion
repository.

Notably, these effects interact with the models' failure modes, as discussed below, where reasoning-oriented configurations also differ in their tendency to abstain.

\subsubsection{Failure modes: abstention vs incorrect outputs}

\begin{figure}
    \centering
    \includegraphics[width=\linewidth,trim=0 10pt 0 20pt,clip]{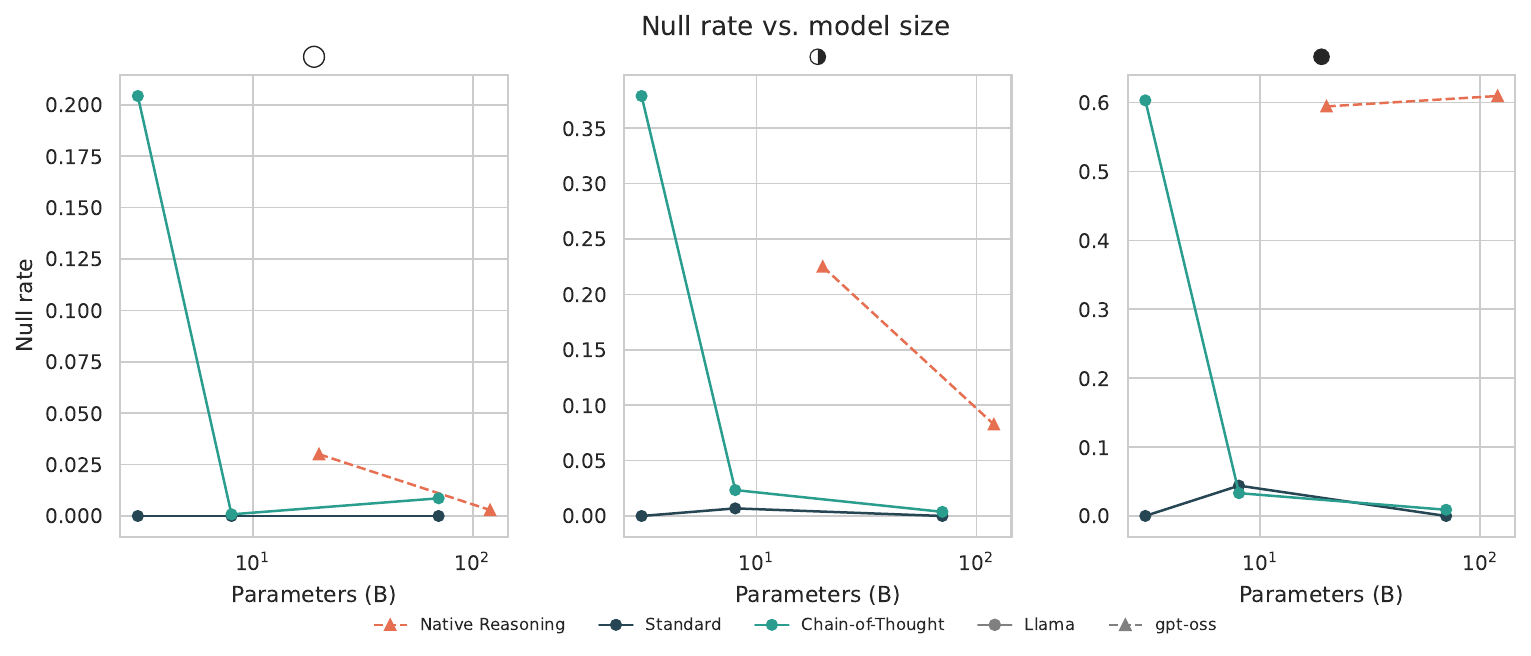}
    \caption{
    Null prediction rate as a function of model size. 
    Higher values indicate increased abstention. 
    }
    \label{fig:null-rate}
\end{figure}

A qualitative difference emerges between model families. While \llama models consistently produce numeric outputs across all configurations, \gpt models exhibit a different behavior, particularly for complex features. As shown in \Cref{fig:null-rate}, \gpt models frequently abstain from producing a value for \hardfeature{} features, with null rates approaching saturation in some cases. This contrasts with \llama models, which tend to produce outputs even when incorrect. As a result, improvements in apparent accuracy must be interpreted jointly with the corresponding increase in abstention.

For this reason, accuracy metrics for reasoning-oriented models should be interpreted jointly with coverage. A model that abstains more frequently may avoid incorrect numerical commitments, but it also reduces the fraction of instances for which direct feature extraction produces a usable value.

This abstention behavior also varies systematically with model size:
regressing \textsc{Null rate} on $\log_{2}(\text{model size})$ yields a
slope close to zero for \easyfeature{} and \midfeature{} features, but
becomes the largest of the three considered metrics at the \hardfeature{}
level (around $+0.07$ per parameter doubling, with $R^{2} \approx 0.25$
across the three representations; see \ref{app:scaling-regression}).
Larger models in the \gpt family therefore do not principally improve
their accuracy on high-computation features; rather, they increase the
share of inputs on which they decline to commit to a value.

\subsubsection{Computational cost}
 
\begin{figure}
    \centering
    \includegraphics[width=\linewidth,trim=0 10pt 0 50pt,clip]{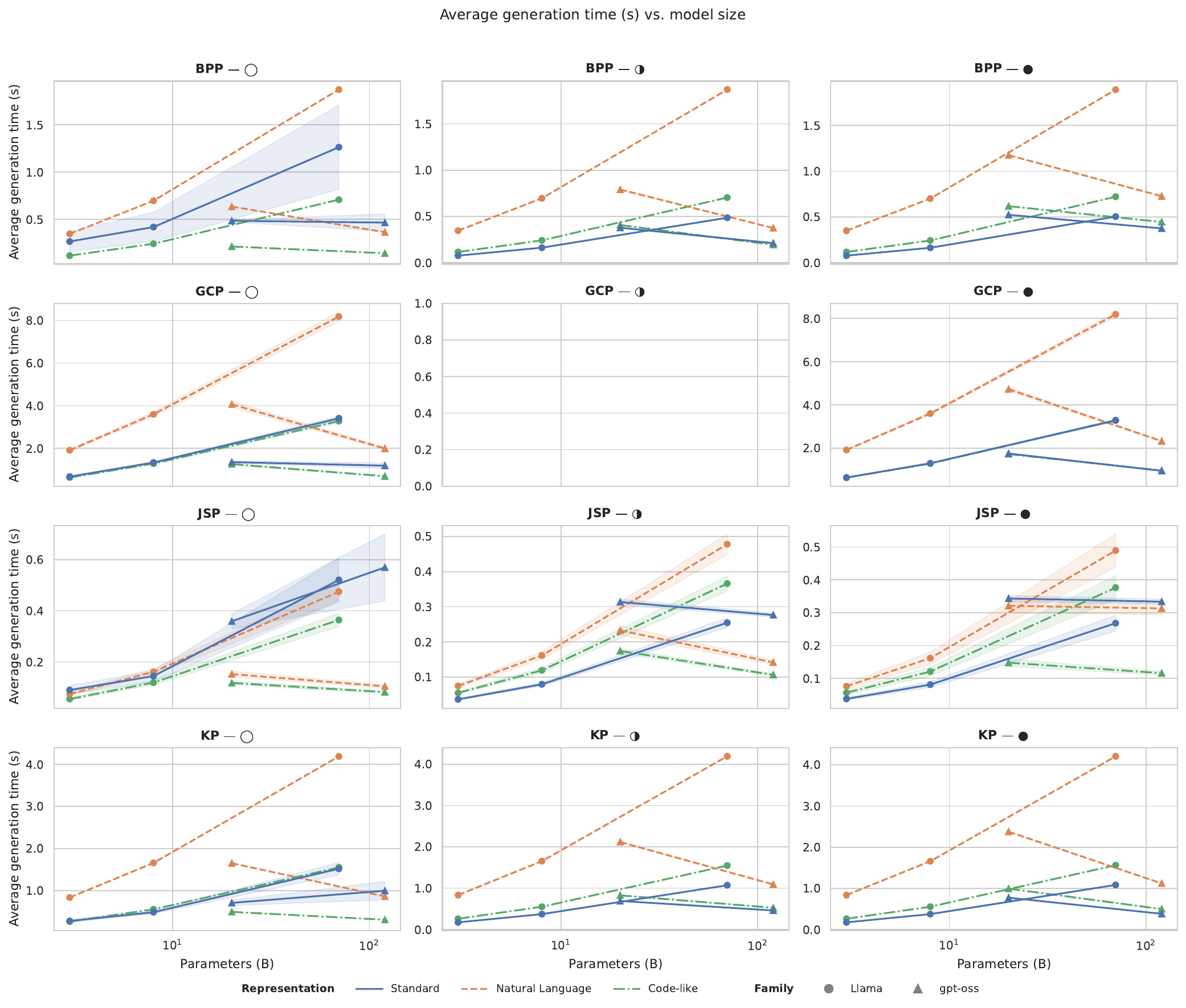}
\caption{
Direct-querying computational cost as model size increases, measured as average generation time per query. 
Each panel corresponds to a feature-complexity group. 
Results correspond to standard direct querying for \llama models and native reasoning for \gpt models.
}
    \label{fig:scaling-time}
\end{figure}

\Cref{fig:scaling-time} reports the average generation time per query as model size increases across problems and feature-complexity groups, and \Cref{tab:inference-time} (in \ref{app:scaling-regression}) summarizes the corresponding mean times by model and mode. As expected, computational cost grows substantially with model size: average generation time spans roughly $0.2$--$0.5$\,s on the $3$B model, $0.4$--$1.2$\,s on the $8$B model, and $1.2$--$3.5$\,s on the $70$B model. An interesting exception is
\gpt[120b], which averages $0.4$--$0.7$\,s despite its parameter count: its mixture-of-experts architecture activates only a fraction of the weights per token, making it faster than the dense \llama[3.3][70B] at all complexity levels.
 
These figures should be read in light of the cost of the handcrafted alternative. The features used as ground truth in this study are computed by simple deterministic procedures over instance arrays of $10^{2}$--$10^{4}$ elements, with extraction-complexity features being $O(1)$ lookups, low-computation features being $O(n)$ scans for minimum or maximum, and high-computation features being $O(n)$ averages or simple aggregations. On the same hardware used to host the inference workload, all these procedures execute in tens of microseconds to a few milliseconds per instance. The \gls{llm} extraction therefore takes between three and five orders of magnitude longer than the handcrafted alternative for the same output, before accounting for GPU energy consumption, which the handcrafted procedures do not incur at all.
 
These cost increases are not always matched by proportional gains in predictive performance. For \easyfeature{} features, where accuracy is
already near saturation, larger models incur substantially higher  inference costs without meaningful improvements. For \midfeature{} and \hardfeature{} features, scaling yields measurable gains, but at a steep computational expense. The practical implication is that, when handcrafted extractors exist and are reliable, using \glspl{llm} as feature replacement is hard to justify on cost grounds; their value lies elsewhere, namely in settings where instances are described in formats for which no handcrafted extractor is available, or as latent feature providers as discussed in \Cref{sec:algorithm-selection-experiment}.
 
\begin{figure}
    \centering
    \includegraphics[width=\linewidth,trim=0 10pt 0 30pt,clip]{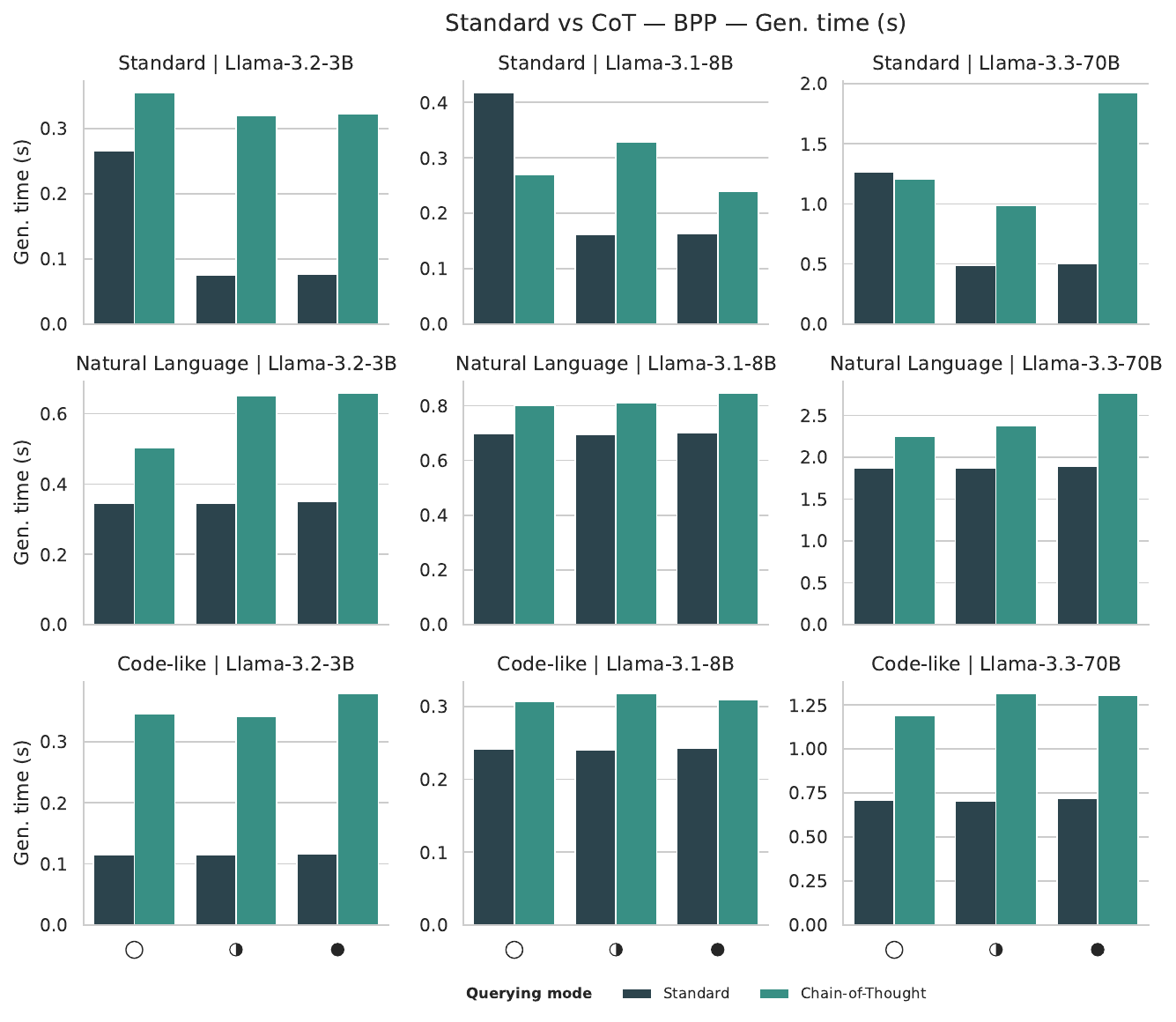}
    \caption{
Comparison of average computational cost (generation time) between standard direct querying and \gls{cot} prompting for \llama models on the BPP problem. Each panel corresponds to a representation--model combination.
}
    \label{fig:diff-cot-time}
\end{figure}

A similar trade-off emerges when considering explicit reasoning.
\Cref{fig:diff-cot-time} compares the average generation time of  standard direct querying and \gls{cot} prompting for \llama models on BPP (the corresponding results for the other problems show the same qualitative pattern). As expected, \gls{cot} prompting increases generation time across model sizes and representations, but the magnitude of this increase is not uniform. The aggregate \gls{cot}-to-standard time ratios are $1.87\times$ on the $3$B model, $1.27\times$ on the $8$B model, and $1.53\times$ on the $70$B model. For the $3$B model, the additional cost is most pronounced on the more complex feature groups, where explicit reasoning introduces a substantial overhead without delivering corresponding gains in performance. For the $8$B model, the increase in generation time is more limited, while the accuracy gains remain selective. For the $70$B model, the strongest time increase is observed for the standard representation, which is also the least effective one overall; in this case, \gls{cot} yields only modest improvements, indicating that additional reasoning effort does not necessarily translate into proportional gains. The smallest model therefore pays the largest relative cost for \gls{cot} prompting, while also being the only one for which prompting consistently degrades accuracy.  
Taken together, these results show that both model scaling and explicit reasoning improve direct querying only under specific conditions, while always increasing computational cost. This makes the trade-off between performance and efficiency central to the evaluation of \gls{llm}-based feature extraction. 

We do not report direct energy measurements, and therefore do not claim a complete energy-efficiency analysis. Nevertheless, the observed inference times and hardware requirements already indicate a substantially larger resource footprint than deterministic CPU-based feature extraction. The cost comparison should therefore be interpreted as conservative: LLM-based direct extraction is slower by several orders of magnitude and also requires substantially more specialized hardware.

\subsubsection{Summary}

Overall, direct querying is effective for extracting explicit features, but remains fragile for features requiring structured computation. Model scale, input representation, and explicit reasoning all influence performance, yet their effects are strongly mediated by feature complexity. In particular, \gls{cot} prompting does not have a uniform effect: it can be detrimental at small scale, selectively beneficial at intermediate scale, and more consistently helpful only for larger models, always at an additional computational cost. Moreover, model families differ not only in accuracy, but also in how they handle uncertainty, with reasoning-oriented models more frequently favoring abstention over potentially incorrect outputs.

To investigate whether these limitations arise from the models' reasoning behavior or from the nature of their internal representations, the next analysis turns to feature probing.

\subsection{Feature Probing}
\label{sec:feature-probing-experiment}

We now turn to probing to investigate whether feature-related information is encoded in the internal representations of \glspl{llm}. In contrast to direct querying, which evaluates the model's ability to explicitly retrieve a feature value, probing assesses whether the same information can be recovered from hidden-layer activations using simple supervised models.

In this setting, each instance is processed by the \gls{llm} without an explicit feature query, and the resulting hidden states are used as input to regression probes trained to predict the corresponding feature values. This allows us to measure the \emph{latent decodability} of feature information, independently of the model's ability to generate the correct value through its output layer.

Direct querying and probing capture two distinct aspects of model behavior. Direct querying evaluates \emph{explicit recoverability}, i.e., whether the model can generate the correct feature value when prompted. Probing, instead, evaluates \emph{latent decodability}, i.e., whether the same information is encoded in the model's internal representations and can be extracted by a downstream model. These two properties do not necessarily coincide.

\begin{figure}
    \centering
    \includegraphics[width=\linewidth,trim=0 10pt 0 30pt,clip]{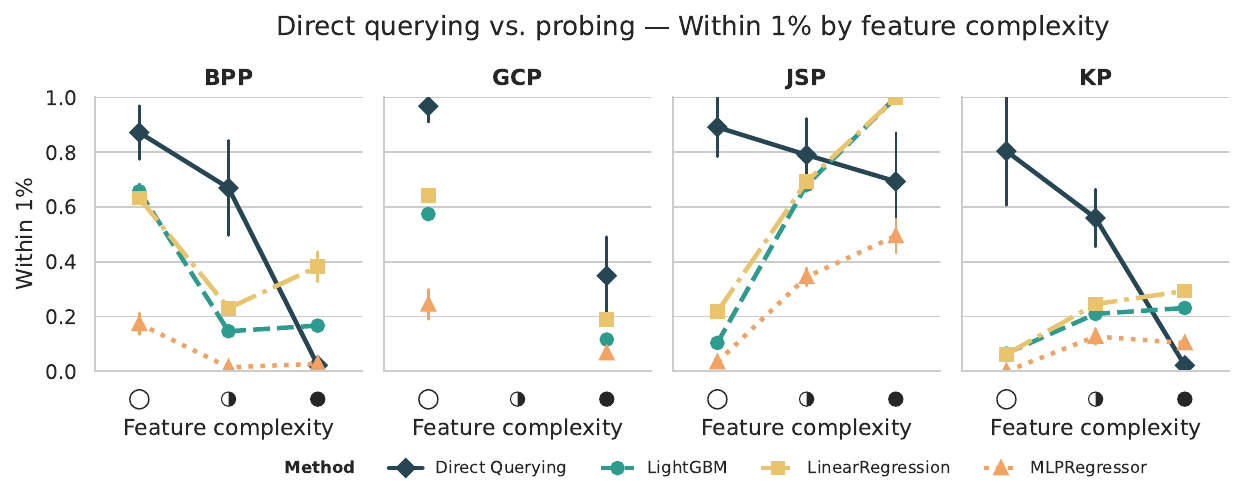}
    \caption{
    Comparison between direct querying and probing performance, measured as Within 1\% accuracy across feature-complexity groups.
    }
    \label{fig:querying-vs-probing-within-1pct}
\end{figure}

\begin{figure}
    \centering
    \includegraphics[width=\linewidth,trim=0 10pt 0 20pt,clip]{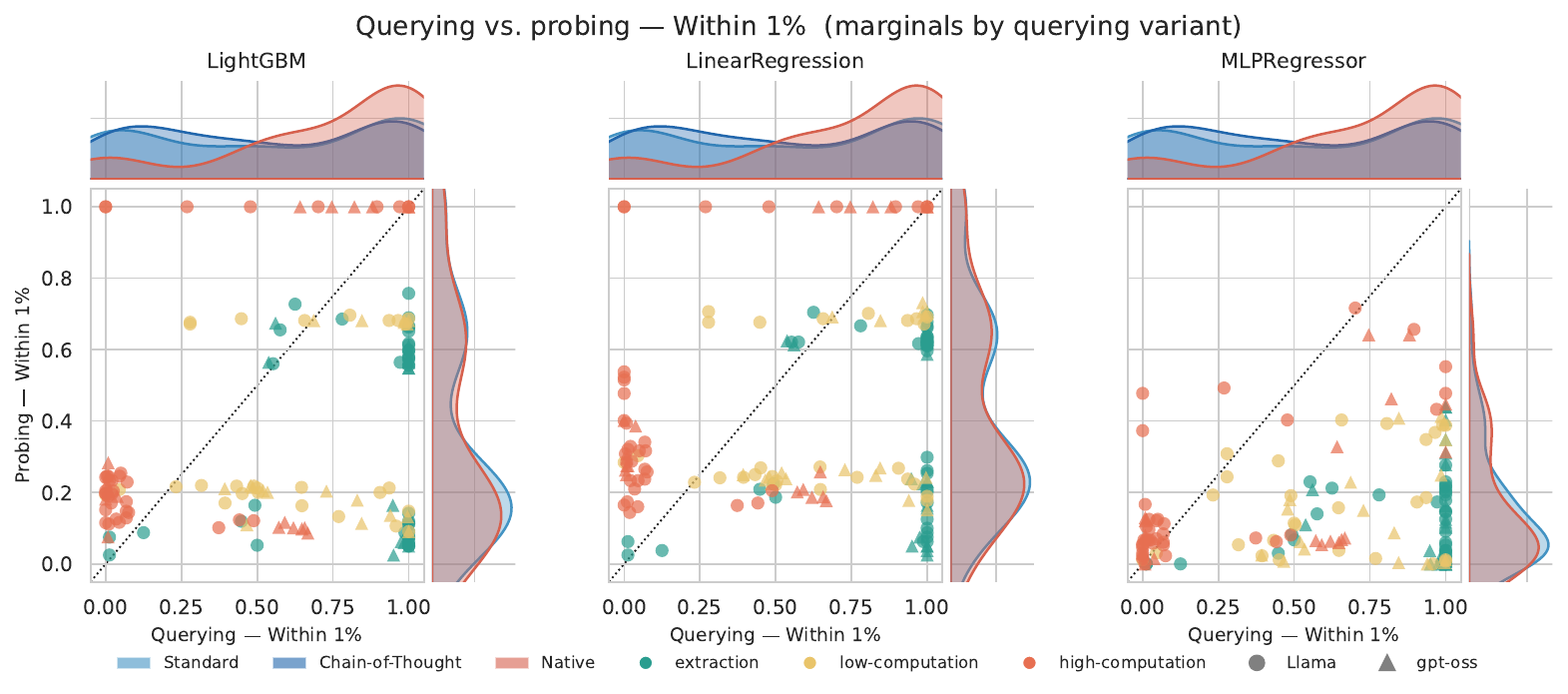}
    \caption{
    Joint comparison of direct-querying and probing performance in terms of Within 1\% accuracy. Marginal distributions summarize the behavior of the two signals.
    }
    \label{fig:querying-vs-probing-joint}
\end{figure}

\Cref{fig:querying-vs-probing-within-1pct} compares direct-querying and probing performance across feature-complexity groups, while \Cref{fig:querying-vs-probing-joint} provides a joint view of the two signals.

The aggregate comparison shows that the relationship between the two signals depends on feature complexity and representation. In some configurations, probing achieves higher accuracy, particularly for more complex features, whereas in others direct querying remains competitive or superior. This indicates that neither method universally dominates the other.

The joint view in \Cref{fig:querying-vs-probing-joint} makes this complementarity more explicit. If direct querying and probing were measuring the same property, most points would concentrate near the diagonal. Instead, the scatter shows a substantial dispersion, with many cases deviating markedly from the line of equality. This indicates that high latent decodability does not necessarily imply high explicit recoverability, and vice versa.

The off-diagonal regions of \Cref{fig:querying-vs-probing-joint} are particularly informative. Points with high probing accuracy and low querying accuracy indicate features whose signal is available to a downstream decoder but not reliably exposed through generation. Conversely, points with high querying accuracy and lower probing accuracy indicate cases where the model can follow the prompt or exploit explicit input structure without producing a representation that is equally useful to the selected probe. The two measurements therefore diagnose different failure modes.

The marginal distributions reinforce this interpretation by showing that the two signals have different global profiles. Direct querying tends to be more polarized, with many cases concentrated near either high or low explicit recovery, whereas probing exhibits a broader spread, suggesting that feature information may be present in the representations even when the model's output fails to retrieve it reliably.

The present analysis is conducted on the stratified $5\%$ subset introduced in \Cref{sec:experimental-setup}, which limits the   statistical strength of the observed patterns and introduces some   variability across configurations. The qualitative complementarity between direct querying and probing is nevertheless evident across all five evaluated models.

Taken together, these results show that direct querying and probing provide complementary but non-equivalent signals about feature information in \glspl{llm}. Direct querying measures explicit recoverability, whereas probing captures latent decodability. The broader implications of this divergence are discussed in \Cref{sec:discussion}.

This distinction has important implications. It suggests that part of the limitation observed in direct querying arises from the generative process itself, rather than from the absence of relevant information in the model. At the same time, the variability of probing results indicates that not all feature information is robustly encoded in a form that can be reliably exploited.

These findings motivate the next step of the analysis. If latent representations contain information that is not always accessible through direct querying, a natural question is whether such information can nevertheless support downstream tasks. In the following section, we evaluate this hypothesis through an algorithm selection setting, assessing whether \gls{llm}-derived representations can act as effective instance descriptors.

\subsection{Algorithm Selection Probing}
\label{sec:algorithm-selection-experiment}

We now assess whether this latent information can be exploited in a downstream algorithm-selection task. Specifically, we test whether \gls{llm}-derived embeddings can match the effectiveness of manually designed features in predicting the best-performing algorithm for each instance.

To address this question, we assess whether \gls{llm}-derived embeddings can match the effectiveness of manually designed features in the key downstream task of algorithm selection. A complete summary of the statistical analyses performed below is reported in \ref{app:statistical-analyses}.

This analysis is also conducted across all five open-weight models considered in the study, spanning the $3$B--$120$B range and covering both the \llama instruct and \gpt reasoning families. This enables us to evaluate whether the predictive utility of \gls{llm}-derived representations is a stable property across models, rather than a phenomenon specific to a single configuration.

\subsubsection{Comparative Statistical Analysis of Classifier Performance}

This subsection presents a comparative statistical analysis of the classifiers employed in the algorithm selection task, aiming to determine whether their predictive performances differ significantly. The analysis serves as a preliminary validation step to ensure that the choice of classifier does not confound the interpretation of subsequent results. To this end, multiple classifiers are evaluated under identical data partitions.

To assess the influence of classifier type on prediction accuracy,\footnote{For these analyses the term accuracy has to be intended in the set-aware sense presented in \Cref{sec:methodology:probing-alg-sel}} we fitted a linear mixed-effects model with block (\emph{problem}, \emph{replicate}) as a random intercept, including both \emph{classifier} and \emph{model} as fixed factors. This formulation allows us to evaluate classifier effects while controlling for the specific representation source used in the downstream task.

The overall effect of classifier was highly significant (likelihood-ratio test, $LR = 742.403$, $df = 3$, $p = 1.34\times10^{-160}$), indicating that classifier choice substantially affects predictive accuracy. Relative to the \mostFrequent{} baseline, \lightGBM{} achieved the strongest improvement ($+0.041$, $p < 0.001$), followed by \logisticRegression{} ($+0.014$, $p < 0.001$), whereas the \mlpClassifier{} showed a statistically significant decrease ($-0.013$, $p < 0.001$).

Controlling for classifier, all evaluated representation sources yielded positive coefficients of similar magnitude relative to the baseline condition, with estimated effects in the range of approximately $+0.016$ to $+0.019$. This indicates that the choice of downstream classifier remains a stronger determinant of performance than the specific model representation among those considered here. The estimated random-intercept variance at the (problem, replicate) block level ($\sigma^2 = 0.032$) further suggests substantial problem-dependent variability, but does not alter the overall ranking of classifiers.

\begin{figure}
    \centering
    \includegraphics[width=\linewidth,trim=5pt 5pt 5pt 50pt,clip]{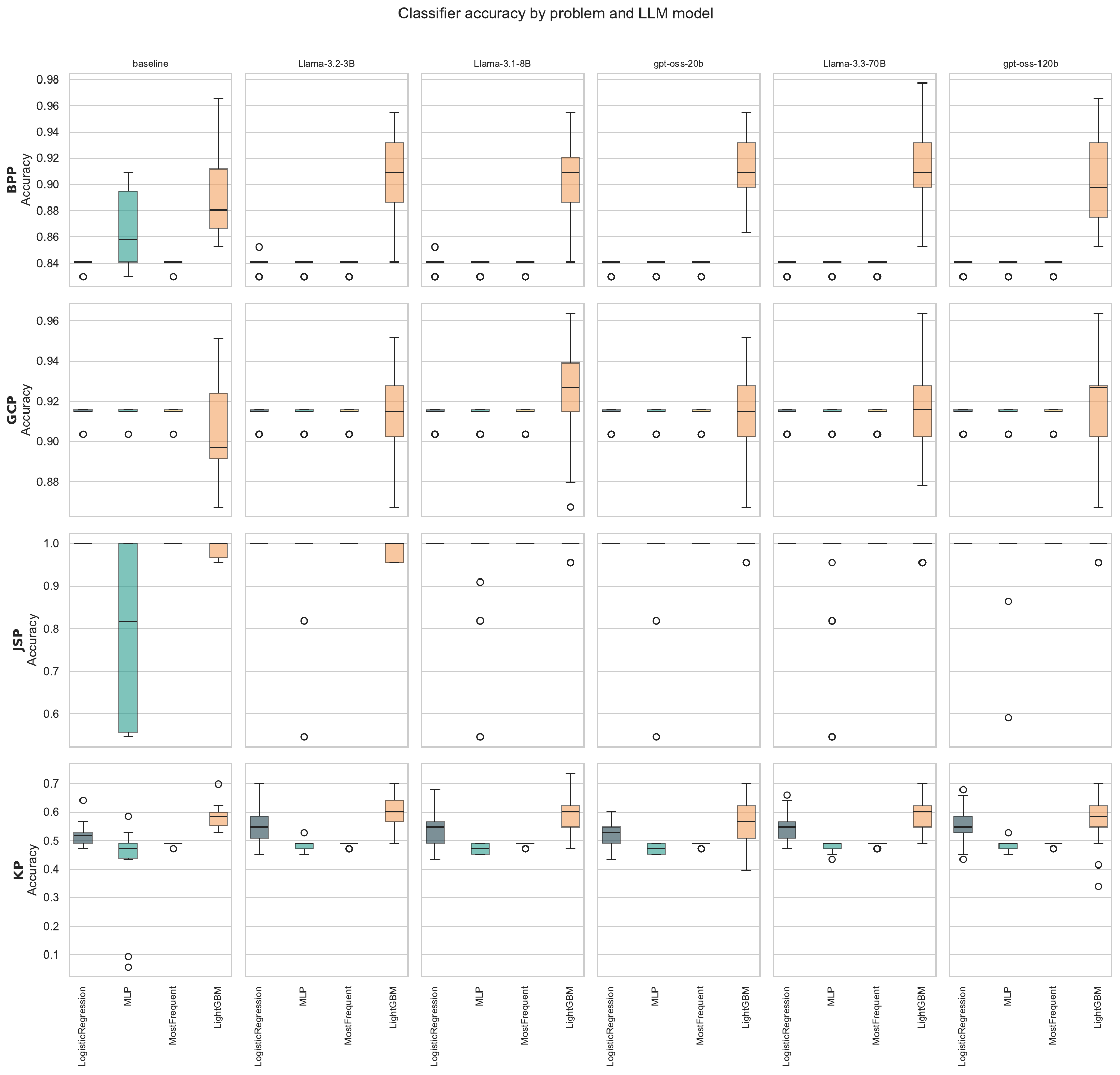}
\caption{Classification accuracy by problem, classifier, and model. The relative ranking of classifiers remains stable across models, confirming that classifier choice is a primary driver of performance.}
    \label{fig:classifier-performance-model}
\end{figure}

\Cref{fig:classifier-performance-model} shows the distribution of classifier accuracies across problems and models. 
The figure confirms that the relative ranking of classifiers is stable across both problem types and underlying model representations: \lightGBM{} consistently provides the strongest predictive performance, \logisticRegression{} yields a smaller but robust improvement over the \mostFrequent{} baseline, and \mlpClassifier{} underperforms relative to the other classifiers.

Performance differences are also problem-dependent. In particular, \acrshort{kp} is consistently the most challenging problem, with substantially lower accuracies across all classifiers and models. By contrast, \acrshort{jssp} yields the highest overall accuracies, despite involving a larger number of candidate algorithms, indicating that problem difficulty in this setting is not determined simply by the number of classes but also by the structure of the task and the set-aware evaluation criterion. \acrshort{bp} and \acrshort{gcp} stay at an intermediate but relatively stable range. This pattern aligns with the non-zero between-problem variance estimated by the mixed-effects model.

Given these results, the subsequent analyses focus on \lightGBM{} and \logisticRegression{}, which provide the most stable and competitive predictive behavior. This choice ensures that downstream comparisons primarily reflect differences in representational quality rather than limitations of the classification models themselves.

\subsubsection{Comparative Analysis of ISA-Based and Handcrafted Feature Sets}

We now compare the predictive behavior of the \gls{isa}-based and handcrafted feature sets to assess whether they yield systematically different outcomes in the algorithm selection task. While \gls{isa}-based features are designed as comprehensive descriptors and may include or extend the handcrafted ones, it is nevertheless important to evaluate whether this broader coverage translates into superior predictive power. Establishing this comparison also provides a consistent baseline for the subsequent analysis, where the predictive power and robustness of manually engineered features are contrasted with those derived from \gls{llm} hidden representations.

To examine whether the predictive behavior of the two baseline feature sets differs systematically, we fit a linear mixed-effects model with \emph{classifier} as a fixed factor and \emph{block}, defined as the (\emph{problem}, \emph{replicate}) pair, as a random intercept. 

The analysis did not identify a significant main effect of the baseline representation (LRT $\chi^2(1) = 0.819$, $p = 0.365$), indicating that ISA-based and handcrafted features yield comparable predictive performance in this setting. The estimated coefficient for the ISA representation ($+0.007$, 95\% CI $[-0.009, 0.023]$) is small and not statistically significant.

The classifier effect is consistent with the previous analysis, with \textsf{LightGBM} outperforming \textsf{Logistic Regression} ($+0.024$, $p = 0.004$).

Complementary tests reinforce this interpretation. A Tukey HSD comparison does not detect significant differences between ISA-based and handcrafted features, while a Wilcoxon-based \gls{tost} fails to establish practical equivalence within a margin of $\varepsilon = 0.01$. However, the low statistical power of the equivalence test suggests that these results should be interpreted cautiously.

Overall, these findings indicate that ISA-based and handcrafted features exhibit similar predictive performance, with no robust evidence of a systematic advantage for either representation. This makes them suitable and comparable baselines for evaluating \gls{llm}-derived representations in the subsequent analysis.

\subsubsection{Analysis of Representation and Pooling Effects on Probing Performance}

To investigate how different combinations of instance representation and pooling strategy influence probing performance, a full-factorial analysis is conducted to evaluate main and interaction effects between the three representations (\standard{}, \code{}, and \nlp{}) and the three pooling strategies (\textit{mean}, \textit{max}, and \textit{last}) on classifier accuracy. The analysis controls for problem-specific variability to isolate systematic effects attributable to representation and pooling choices, providing insight into how these factors shape the extractable information from \gls{llm} hidden activations.

We fitted a linear mixed-effects model including \emph{classifier} as a fixed factor and \emph{block} (i.e., the pair (\emph{problem}, \emph{replicate})) as a random intercept to assess how different combinations of representation and pooling activations influence classifier performance in the algorithm selection task. The model evaluated the main and interaction effects of the two factors, while accounting for systematic differences between classifiers. 
In addition, the effect of the underlying \gls{llm} model was included as a fixed factor. The likelihood-ratio test indicates that the model effect is not significant ($\chi^2(4) = 6.381$, $p = 0.172$), suggesting that differences across models do not substantially affect downstream performance in this setting.

Likelihood ratio tests indicate that the effect of pooling is significant ($\chi^2(6) = 37.389$, $p = 1.48\times10^{-6}$), whereas the effect of representation is not statistically significant ($\chi^2(6) = 11.353$, $p = 0.078$).
A main effect of classifier was also observed, with \textsf{LightGBM} outperforming \textsf{Logistic Regression} by approximately $0.027$ accuracy points ($p < 0.001$). Overall, these results indicate that classifier choice dominates downstream performance, while pooling has only a modest effect and neither representation nor \gls{llm} model is significant. No systematic representation--pooling interaction was detected.

The estimated coefficients indicate only minor differences between representations, with both \nlp{} and \code{} showing small, non-significant deviations from the \standard{} baseline. This is consistent with the likelihood-ratio test, which does not support a statistically significant effect of representation. Overall, these results suggest that the choice of representation has limited influence on downstream performance in the probing setting.

Among pooling strategies, a statistically significant effect is observed overall. In particular, \emph{last} pooling shows a small but significant decrease in performance relative to the \emph{mean} baseline ($-0.010$, $p = 0.002$), whereas \emph{max} pooling does not exhibit a significant difference. Although these effects are statistically detectable, their magnitude remains small.

\begin{figure}
    \centering
    \includegraphics[width=\linewidth,trim=10pt 70pt 10pt 30pt,clip]{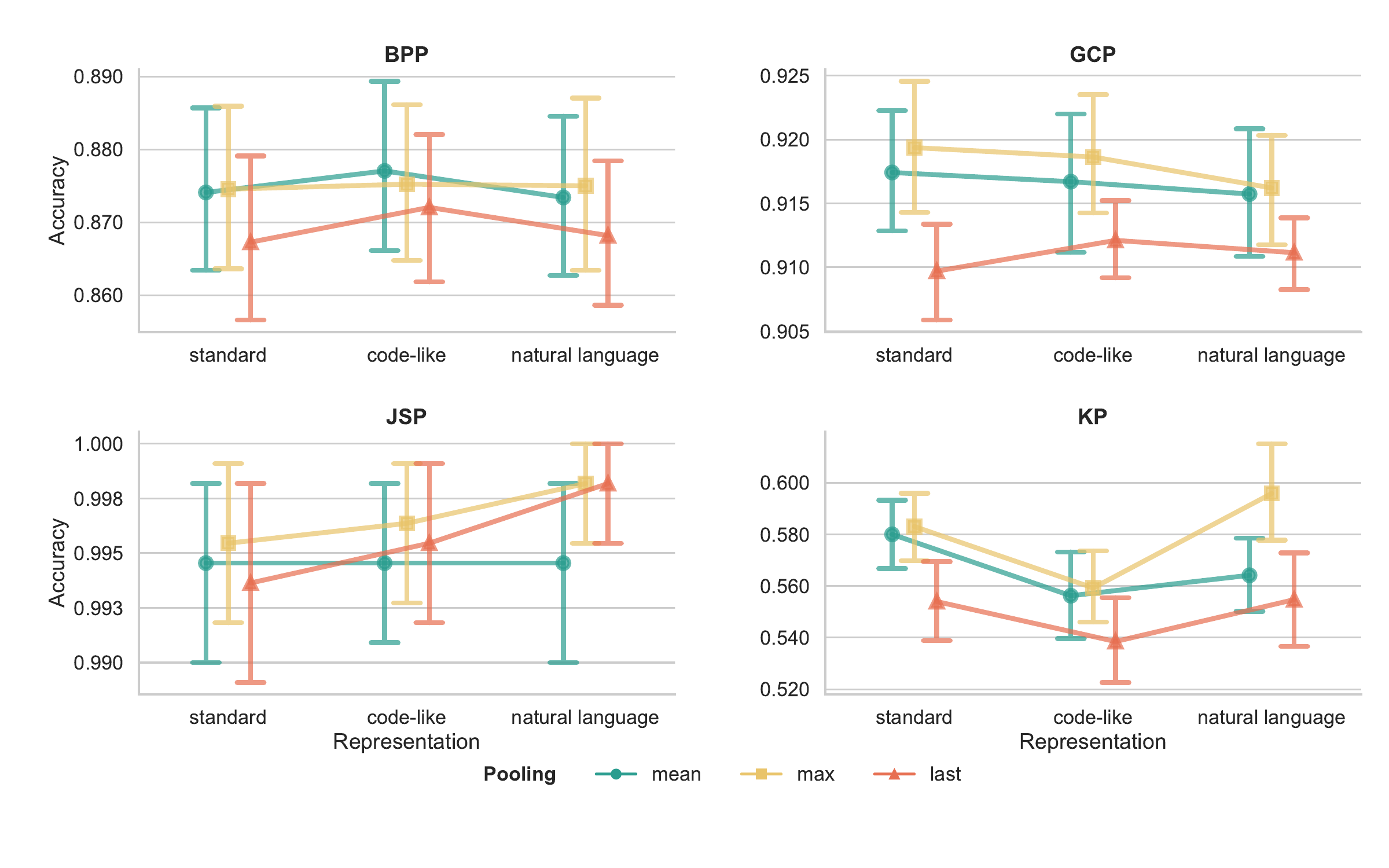}
    \caption{Effects of representation and pooling in the full factorial experiment.
Each panel shows the mean accuracy ($95\%$ CI) for combinations of representation and pooling, averaged across problems and classifiers.}
    \label{fig:representation-pooling-effects}
\end{figure}

\Cref{fig:representation-pooling-effects} illustrates the mean accuracies (with 95\% confidence intervals) obtained for each problem under different combinations of representation and pooling strategy. Across all problems, \emph{max} pooling consistently provides a slight advantage over the other two pooling methods, regardless of the underlying representation. The effect of representation is minimal and does not show a consistent pattern across problems, whereas pooling exhibits small but more systematic differences.

Overall, representation choice has no significant impact on downstream performance, while pooling introduces only minor effects. These results suggest that the predictive signal encoded in \gls{llm} representations is largely robust to variations in representation, pooling strategy, and model scale.

For the final analysis, we adopt a single configuration consisting of \emph{max} pooling and the \standard{} representation to provide a consistent and reproducible basis for comparison with baseline feature sets.

\subsubsection{Comparative Analysis of ISA and LLM-Derived Representations for Algorithm Selection}

We finally compare ISA-based features with \gls{llm}-derived representations to assess whether latent embeddings can support competitive downstream performance when paired with the best-performing downstream classifier, namely \lightGBM{}.

The comparison does not support a uniform advantage of the ISA baseline over \gls{llm}-derived representations. Instead, the observed differences are small and problem-dependent. In particular, the only problem showing a statistically significant difference is BPP, where \gls{llm}-derived representations achieve slightly higher accuracy than the \gls{isa} baseline ($+0.041$ accuracy points, $p = 0.001$). No robust evidence of systematic differences emerges for the other benchmark problems.

\begin{figure}
    \centering
    \includegraphics[width=\textwidth,trim=5pt 10pt 5pt 80pt,clip]{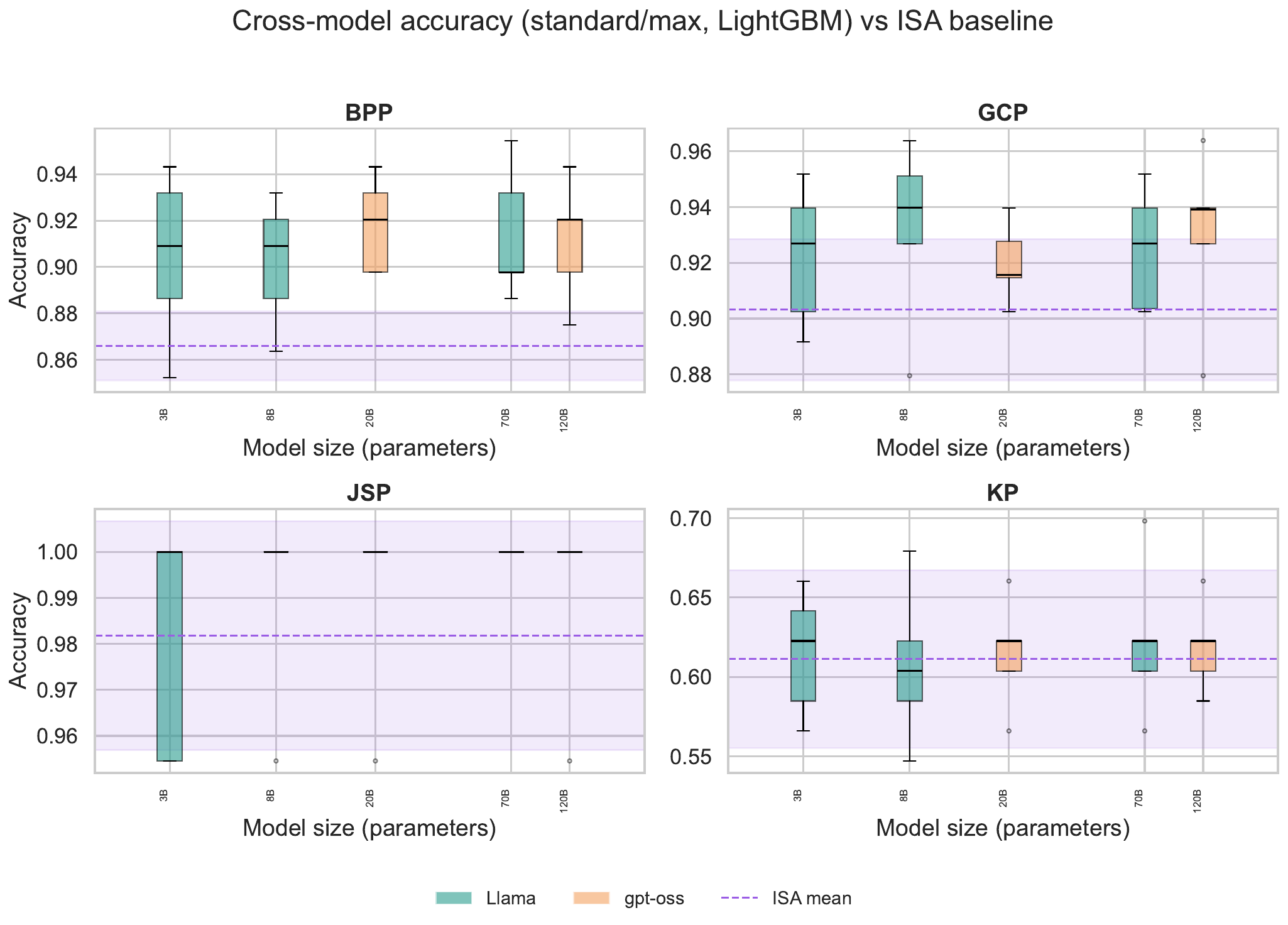}
    \caption{
Comparison of algorithm selection accuracy across problems between the ISA baseline and \gls{llm}-derived representations, using \lightGBM{} as the downstream classifier. Results are reported across the evaluated models. The dashed horizontal line denotes the mean ISA accuracy, and the highlighted band indicates the interval within one standard deviation of the ISA baseline. Overall, performance differences remain small and problem-dependent.
}
    \label{fig:cross-model-accuracy}
\end{figure}

To further inspect this pattern, we analyze the cross-model accuracies reported in \Cref{fig:cross-model-accuracy}. The figure shows that \gls{llm}-derived representations achieve predictive performance that is broadly comparable to the ISA baseline across models, with relatively small deviations in absolute accuracy. Even in the case of BPP, where the global test detects a significant difference, the post-hoc comparisons among the \gls{llm}-based configurations do not reveal strong separation, suggesting that the practical gap remains limited.

\section{Discussion: Implicit Signal vs.\ Explicit Retrieval}
\label{sec:discussion}

The experimental analysis in \Cref{sec:experimental-analysis} reveals a recurrent functional asymmetry between what \glspl{llm} produce as explicit numerical answers and what their internal representations encode.
Direct querying recovers feature values reliably only when they can be read off the input with little or no computation, while probing decodes the same features from hidden states with consistently lower error, particularly on the more demanding feature-complexity groups (\Cref{fig:querying-vs-probing-within-1pct,fig:querying-vs-probing-joint}).
We collect in this section the elements of this dissociation under a common interpretive frame and discuss what they imply for the use of \glspl{llm} as feature providers in combinatorial-optimization pipelines.
Our reading remains descriptive: the data support a clear separation between \emph{explicit recoverability} and \emph{latent decodability}, but we do not attempt to settle the broader question of whether such separation should be read as evidence of comprehension or its absence.

We deliberately avoid interpreting this dissociation as direct evidence either for or against model comprehension. The experiments show that feature-related information can be decodable from hidden states while remaining unavailable to reliable explicit generation. Whether this reflects partial understanding, statistical association, or a limitation of the decoding and reasoning process is a broader mechanistic question beyond the scope of this study.

\paragraph{Failure modes}

The dissociation is reinforced by a second observation: models in our pool do not differ only in how often they retrieve the correct value, but in \emph{how} they fail when retrieval is unsuccessful. Three failure modes emerge from \Cref{fig:null-rate} and from the per-cell breakdown in \ref{app:scaling-regression}, each associated with a specific portion of the model pool. The first is \emph{calibrated abstention}, observed primarily in \gpt models on \hardfeature{} features: when reliable computation is not possible, the model returns \texttt{null} under structured decoding, with abstention rates approaching saturation in some configurations. 
The second is \emph{over-confident hallucination}, observed primarily in \llama models and most pronounced under \gls{cot} prompting at intermediate scale: the model commits to a type-valid but incorrect numeric value rather than abstaining, a behavior that structured decoding does not prevent because the constraint is on output type, not on output validity. 
The third is \emph{structured-output failure}, observed in the smallest model (\llama[3.2][3B]) under \gls{cot} prompting on \hardfeature{} features. Producing an explicit reasoning trace consumes most of the available generation budget and leaves the model unable to emit a well-formed value field on a substantial share of features; the structured decoder then records the result as null. The output is identical in form to a calibrated abstention, but does not carry the same meaning: it reflects a failure to complete the required output structure rather than a decision to refuse. 
The three modes are not equivalent for downstream use: calibrated abstention carries usable information about the model's own limits, over-confident hallucination silently corrupts the output, and structured-output failure leaves the pipeline with non-informative gaps.

\paragraph{Latent encoding without exact computation}

A consistent picture emerges when the dissociation between querying and probing is combined with the scaling regressions reported in \ref{app:scaling-regression}. 
Probing shows that feature-relevant information is encoded in the model's hidden states: a regressor trained on these states recovers feature values from instances that the model itself fails to extract under direct querying, and downstream classifiers built on the same representations match the predictive power of manually engineered features in algorithm selection (\Cref{sec:algorithm-selection-experiment}). 
At the same time, explicit retrieval improves with scale but does not match handcrafted extraction: normalized \gls{mae} decreases by roughly $40\%$ to $60\%$ per parameter doubling on the \midfeature{} and \hardfeature{} groups, while the null-rate slope on \hardfeature{} features grows in parallel ($+0.07$ per parameter doubling, $R^{2} \approx 0.25$). 
Within the $3$B--$120$B range we evaluate, larger models compress this residual error and become, at the same time, more likely to abstain on features where the answer requires actual computation. This is consistent with a reading in which the latent representation captures statistical regularities of the input--feature relation without instantiating a procedure that computes the feature exactly. We summarize this distinction as one between an \emph{encoded statistical signal} and an \emph{executable computation}.
The probing pipeline recovers the former with comparable accuracy across model sizes; the latter is the deterministic procedure that maps the raw instance to the feature value, and the model's output only approximates it.

\paragraph{Where latent representations contribute}

The implications of the dissociation for downstream tasks are clearest in the algorithm-selection setting, where hidden-state representations match the predictive performance of handcrafted feature sets across all five evaluated models (\Cref{sec:algorithm-selection-experiment}), with no robust evidence of advantage for either side. Given this parity, the two approaches are most usefully compared not on accuracy but on the structure of their cost. 
A handcrafted feature set requires problem-specific design, the implementation of an extractor for each input format, and empirical validation, often through feature-selection or instance-space analysis. This is a substantial up-front investment in domain expertise and engineering effort, carried out once per problem,
after which the per-instance computation is negligible. 
Latent representations reverse this profile: the same pretrained \gls{llm} can be applied to all four problems in our study without modification, and no problem-specific feature engineering is required, but each instance carries a non-negligible inference cost.

The two approaches are therefore best suited to different situations. Handcrafted features amortize their development cost when a problem is studied repeatedly, when its input formats are stable, and when the deployment scale makes per-instance cost the dominant term. Latent representations are more attractive when algorithm selection is required for new or non-standard problems, when instances arrive in heterogeneous formats, or when the infrastructure for handcrafted feature engineering is not yet in place.

Thus, \gls{llm}-derived representations should not be viewed as drop-in replacements for exact feature extractors. Their potential value lies in a different regime: when handcrafted extractors are unavailable, expensive to design, difficult to maintain across heterogeneous formats, or when a reusable representation is needed before problem-specific feature engineering has been carried out.

\section{Conclusions}
\label{sec:conclusions}
 
\glsresetall
 
We investigated how \glspl{llm} represent \glspl{cop} and whether these internal representations can support downstream decision-making tasks.
Through a combination of \emph{direct querying} and \emph{probing}
analyses, we examined both the explicit reasoning capabilities and the implicit knowledge encoded within the models across four benchmark problems, three types of instance representations, and five open-weight models spanning the $3$B--$120$B range.
 
Two tasks were considered: \emph{feature extraction}, assessing whether models can recover meaningful structural and numerical characteristics from problem instances, and \emph{per-instance algorithm selection}, testing whether internal representations can predict the best solver for each instance.
 
Our results reveal that:
 
\begin{itemize}
\item The direct querying experiments (\Cref{sec:direct-querying-experiment}) show that \glspl{llm} can reliably recover simple, explicitly stated instance features, particularly from \nlp{} and \code{} representations. However, performance deteriorates markedly for features requiring even limited computation or aggregation. While increasing model scale improves performance in some settings, these gains are strongly conditioned by feature complexity and representation, and remain limited for more complex features. Explicit \gls{cot} prompting has a scale-dependent effect: it helps mainly at larger scales and consistently increases computational cost.
 
\item The probing analyses (\Cref{sec:feature-probing-experiment}) show that \glspl{llm} encode structural and numerical information in their internal representations, but that this information is not always accessible through direct querying. This supports the distinction between latent decodability and explicit recoverability discussed in \Cref{sec:discussion}.
 
\item Through an extensive experimental campaign (\Cref{sec:algorithm-selection-experiment}), we find that \gls{llm}-derived representations can serve as effective surrogate instance descriptors for per-instance algorithm selection. Across all evaluated models, their predictive performance is broadly comparable to traditional feature sets, with differences that are small and problem-dependent. In particular, no robust evidence of a systematic advantage emerges for \gls{isa}-based features over \gls{llm}-derived representations.
\end{itemize}
 
Taken together, these findings reveal a functional asymmetry between explicit reasoning and internal representation in \glspl{llm}. In practical terms, LLM-derived representations are best viewed as complementary descriptors for settings where handcrafted feature engineering is unavailable, costly, or difficult to maintain across heterogeneous formats.

Several directions extend this work. Recent progress in structured reasoning and tool-augmented generation with \glspl{llm} offers a complementary route in which exact computation is delegated to deterministic sub-procedures, while the model is responsible for instance interpretation and surrogate representation. A second avenue lies in integrating \gls{llm}-derived activations into broader \gls{isa} pipelines, using them not only for algorithm selection but also for tasks such as instance clustering, performance landscape modeling, and exploratory analysis of problem spaces. Extending the investigation to other model architectures, larger instance scales, and real-world optimization problems represents a further important direction.

Two limitations of the study deserve to be noted explicitly. The experimental campaign is conducted on a stratified subset of instances drawn from four benchmark \glspl{cop}, which may affect the stability of some comparative results, particularly in the probing analysis. The model pool spans the $3$B--$120$B range and is restricted to open-weight \glspl{llm} for reasons of reproducibility and access to internal activations, leaving closed-weight or substantially larger models outside the present study.

\section*{Acknowledgments}

The computational experiments were conducted using the LABIC research infrastructure, supported by funding from the Regione Autonoma Friuli Venezia Giulia, the Italian Ministry of Foreign Affairs, and the Italian Ministry of University and Research.

\section*{Declaration on the Use of Generative Artificial Intelligence}

The authors acknowledge the use of ChatGPT and Claude to support language editing, including grammar correction and rephrasing. 
Following its use, all text was critically reviewed, verified, and revised by the authors, who accept full responsibility for the integrity and accuracy of the manuscript.

\bibliographystyle{elsarticle-num-names} 
\bibliography{biblio}

\appendix


\include{appendix_problem-specification}


\include{appendix_prompting-scaling}


\include{appendix_feature-probing}


\include{appendix_statistical-analyses}

\end{document}

%% file: appendix_problem-specification.tex
\section{Case Studies Specifications}
\label{app:case-studies-specification}

{In this appendix, we provide additional details on the case studies used in this paper. Specifically, we report problem formulations, feature definitions, instance formats, and worked examples of the direct-querying prompt for the three problems not already illustrated in the main text. When notation is employed, it is contextual and specific to each problem.}

{The \gls{gcp} is illustrated directly in \Cref{sec:methodology}. For the remaining three problems, we report below an instantiation of the standard direct-querying prompt template introduced in \Cref{sec:methodology:direct-querying}, with one feature chosen per problem to illustrate how the placeholders are populated. The Chain-of-Thought variant differs only in the output instructions, as already shown in the main text. The associated structured-output schema, common to all problems and instantiated below, constrains the model output to a single JSON object whose \texttt{value} field carries the predicted feature.}

\begin{tcolorbox}[colback=white,colframe=gray!75!black, fontupper=\scriptsize\ttfamily, enlarge bottom by=-1.5mm, enlarge top by=-2mm, boxsep=1mm, left=1mm, right=1mm]
\begin{verbatim}
{
  "type": "object",
  "additionalProperties": false,
  "required": ["value"],
  "properties": {
    "value": { "anyOf": [ { "type": "[feature-type]" }, { "type": "null" } ] }
  }
}
\end{verbatim}
\end{tcolorbox}

{For the Chain-of-Thought variant, the same schema is extended with the additional required field \texttt{\small "reasoning":\{"type":"string"\}}.}

\subsection{Bin Packing Problem}
\subsubsection{Problem Formulation}

{The \gls{bp} considers a set of items $\mathcal{I}=\{1,\dots n\}$, with $|\mathcal{I}|=n$, each with its own weight ($w_i$), and an infinite number of bins, all with the same capacity ($C$). The objective is to minimize the total number of bins used, subject to the constraints that every item is placed in a bin, and the sum of the weights of the items in each bin does not exceed the capacity.}

{A possible mathematical formulation for the \gls{bp} is as follows:}

\begin{align}
    \min \quad & \sum_{j \in \mathcal{I}}{y_j} \label{bp:obj}\\
   \text{s.t.} \quad \\
   & \sum_{i \in \mathcal{I}} w_i x_{ij} \leq Cy_j && \forall j \in \mathcal{I} \label{bp:bin-capacity}\\
   & \sum_{j \in \mathcal{I}} x_{ij}=1 && \forall i \in \mathcal{I} \label{bp:one-bin} \\
    & y_j \in \{0,1\} && \forall j \in \mathcal{I} \label{bp:domain-1}\\
    & x_{ij} \in \{0,1\} && \forall i \in \mathcal{I} \quad \forall j \in \mathcal{I} \label{bp:domain-2}\\
\end{align}
{The model is defined using two decision variables: $y_j$, indicating whether bin $j$ is used, and $x_{ij}$, indicating whether item $i$ is assigned to bin $j$. We assume that at most $n$ bins are available. Objective~\eqref{bp:obj} minimizes the number of used bins. Constraint~\eqref{bp:bin-capacity} limits the total weight of the items assigned to each bin by its capacity. Constraint~\eqref{bp:one-bin} enforces that each item is assigned to exactly one bin. Finally, constraints~\eqref{bp:domain-1} and \eqref{bp:domain-2} define the domains of the decision variables.
}

\subsubsection{Features}

{The features considered are as follows:
\begin{itemize}
    \item \textbf{Bin capacity}: The capacity of each bin. This corresponds to $C$ and is directly provided in the instances; therefore, it is an easy feature (\easyfeature).
    \item \textbf{Items}: The number of items to be assigned to the bins. This corresponds to $n$ and is directly provided in the instances; therefore, it is an easy feature (\easyfeature).
    \item \textbf{Item weight}: Metrics obtained by aggregating the item weights $w_i$, $i \in \mathcal{I}$. The associated complexity depends on the aggregation: minimum and maximum are mid-complexity features (\midfeature), as they require ordering operations, whereas the average is a hard feature (\hardfeature), as it requires explicit computation. They are calculated as:
    \begin{align}
    w_{\min} = &  \min_{i \in \mathcal{I}} w_i\\
    w_{\max} = &  \max_{i \in \mathcal{I}} w_i, \text{and}\\
    w_{\mathrm{avg}} = & \frac{1}{n} \sum_{i \in \mathcal{I}} w_i
    \end{align}
\end{itemize}}

\subsubsection{Instance Formats}

{The \textsf{standard} format for \gls{bp} instances is as follows:}

\begin{tcolorbox}[colback=white,colframe=gray!75!black, fontupper=\scriptsize, enlarge bottom by=-1.5mm, enlarge top by=-2mm, boxsep=0mm]
The first line reports the number of items (i.e., the number of weights in the following), the second line reports the capacity of the bins. The subsequent lines report the weights of the items, in descending order.
\end{tcolorbox}

{Thus, an instance in this representation is illustrated below:}

\begin{tcolorbox}[colback=white,colframe=gray!75!black, fontupper=\scriptsize, enlarge bottom by=-1.5mm, enlarge top by=-2mm, boxsep=0mm]
1000

1000

700

700

700

700

\dots
\end{tcolorbox}
{In this case, the instance considers 1,000 items. The bins have a capacity of 1,000. The first element has a weight of 700, etc. The same instance introduced in \textsf{natural language} for is:}

\begin{tcolorbox}[colback=white,colframe=gray!75!black, fontupper=\scriptsize, enlarge bottom by=-1.5mm, enlarge top by=-2mm, boxsep=0mm]
The instance regards 1000 items. Each bin has a capacity of 1000. The weights of the items are as follows.  Note that they are ordered in descending order. Item 1 has a weight of 700. \dots
\end{tcolorbox}

{The correspondent \textsf{code-like} format is:}

\begin{tcolorbox}[colback=white,colframe=gray!75!black, fontupper=\scriptsize, enlarge bottom by=-1.5mm, enlarge top by=-2mm, boxsep=0mm]
int: items = 1000;

array[1..items] of int: weights = [700,700,\dots]; 

int: bin\_capacity = 1000;
\end{tcolorbox}

\subsubsection{{Prompt Example}}

{The example below shows the standard direct-querying prompt instantiated for the \gls{bp}, targeting the maximum item weight (a \midfeature{} feature) on a small instance in the \textsf{standard} format. The same template is used for all features defined for this problem; only the \texttt{feature\_name}, \texttt{feature\_description}, and \texttt{feature\_type} placeholders change accordingly.}

\begin{tcolorbox}[colback=white,colframe=gray!75!black, fontupper=\scriptsize, enlarge bottom by=-1.5mm, enlarge top by=-2mm, boxsep=0mm]
\begin{verbatim}
You are given an instance of a combinatorial optimization problem: Bin Packing
Problem.
This is not a coding task. Do not return any code.
Extract the numeric value of the following feature from the instance:
- Feature name: feat_max_weight
- Feature description: The maximum weight among all the items.
- Expected type: integer
The instance is provided here: """
1000
1000
700
700
700
700
...
"""
Instructions:
- Return a JSON object only, with a single field "value", i.e., { "value": ... }.
- The "value" field should contain the numeric value of the required feature
  and should be of the expected type (i.e., integer).
- If the value is unknown/undeterminable, return { "value": null }.
- No explanations, no extra fields.
Answer:
\end{verbatim}
\end{tcolorbox}

\subsection{Jobshop Scheduling Problem}

\subsubsection{Problem Formulation}

{The \gls{jssp} involves a finite set of jobs (i.e., $\mathcal{J} = \{1,\dots,n\}$, with $|\mathcal{J}|=n$) and machines (i.e., $\mathcal{M}=\{1,\dots,m\}$, with $|\mathcal{M}| = m$). Each job consists of a sequence of operations that must be processed on the machines in a given order (indicated as $(\sigma_1^i,\dots,\sigma_m^i)$ for job $i$), with each job-machine pair associated with a predefined processing time, indicated as $p_{i,j}$. The objective is to construct a feasible schedule that minimizes the makespan. A valid schedule must satisfy three main constraints. First, the prescribed order of operations within each job must be respected. Second, machines cannot process more than one operation at the same time, meaning that no overlaps are allowed. Third, operations are non-preemptive: once an operation has started on a machine, it must run to completion without being interrupted or paused.}

{A possible mathematical formulation for the \gls{jssp} is as follows~\cite{KU2016165}:}

\begin{align}
    \min \quad & C_{\max} \label{jsp:obj}\\
    \text{s.t.} \quad \\
    & x_{\sigma_h^ii} \geq x_{\sigma_{h-1}^ii} + p_{\sigma_{h-1}^ii} && \forall i \in \mathcal{J} \quad \forall h \in \{1,\dots, m\} \label{jsp:precedence}\\
    & x_{ji} \geq x_{jk} + p_{jk} - V \cdot z_{jik} &&  \forall  i \in \mathcal{J} \quad \forall  k \in \mathcal{J} \quad \forall j \in \mathcal{M} \quad i < k \label{jsp:disjoint-1}\\
    & x_{jk} \geq x_{ji} + p_{ji} - V \cdot (1 - z_{jik}) && \forall  i \in \mathcal{J} \quad \forall  k \in \mathcal{J} \quad \forall j \in \mathcal{M} \quad i < k \label{jsp:disjoint-2}\\
    & C_{\max} \geq x_{\sigma_m^ii} + p_{\sigma_m^ii}  && \forall i \in \mathcal{J} \label{jsp:makespan} \\
    & x_{ji} \geq 0  && \forall  i \in \mathcal{J} \quad \forall j \in \mathcal{M} \label{jsp:domain-1}\\
    & z_{jik} \in \{0,1\} && \forall  i \in \mathcal{J} \quad \forall  k \in \mathcal{J} \quad \forall j \in \mathcal{M} \label{jsp:domain-2}
\end{align}
{The disjunctive model uses a decision variable $x_{j i}$ representing the start time of job $i$ on machine $j$, and a binary variable $z_{j i k}$ indicating whether job $i$ is processed before job $k$ on machine $j$. Objective~\eqref{jsp:obj} minimizes the makespan. Constraint~\eqref{jsp:precedence} enforces the technological precedence relations, ensuring that the operations of each job are processed in the correct order. The disjunctive constraints~\eqref{jsp:disjoint-1} and~\eqref{jsp:disjoint-2} guarantee that no two jobs are processed on the same machine at the same time; the constant $V$ is chosen to be sufficiently large. Constraint~\eqref{jsp:makespan} ensures that the makespan is at least as large as the completion time of the last operation of each job. Finally, constraints~\eqref{jsp:domain-1} and~\eqref{jsp:domain-2} define the domains of the decision variables. 
 }

\subsubsection{Features}

{The features considered are as follows:
\begin{itemize}
    \item \textbf{Jobs}: The number of jobs $n$. This information is directly provided in the instance; therefore, it is an easy feature (\easyfeature).
    \item \textbf{Machines}: The number of machines $m$. This information is directly provided in the instance; therefore, it is an easy feature (\easyfeature).
    \item \textbf{Operations}: The product of the number of jobs $n$ and the number of machines $m$. This is considered a mid-complexity feature (\midfeature), as it depends only on two elements that are directly provided in the instance. It is calculated as $n \cdot m$.
    \item \textbf{Duration}: Metrics obtained by aggregating the processing times $p_{i,j}$ of job $i \in \mathcal{I}$ on machine $j \in \mathcal{M}$. The associated complexity depends on the aggregation: minimum and maximum are mid-complexity features (\midfeature), as they require ordering operations, whereas the average is a hard feature (\hardfeature), as it requires explicit computation. They are calculated as follows:
    \begin{align}
    p_{\min} = &\min_{i \in \mathcal{I},\, j \in \mathcal{M}} p_{i,j}, \\
    p_{\max} = & \max_{i \in \mathcal{I},\, j \in \mathcal{M}} p_{i,j}, \\
    p_{\mathrm{avg}} = & \frac{1}{n \cdot m}
    \sum_{i \in \mathcal{I}} \sum_{j \in \mathcal{M}} p_{i,j}.
    \end{align}
\end{itemize}}

\subsubsection{Instance Formats}

{The \textsf{standard} format for \gls{jssp} instances is as follows:}

\begin{tcolorbox}[colback=white,colframe=gray!75!black, fontupper=\scriptsize, enlarge bottom by=-1.5mm, enlarge top by=-2mm, boxsep=0mm]
The first line contains two integers: the number of jobs and the number of machines. This defines the overall structure of the problem. Each subsequent line corresponds to one job and lists all the operations that make up that job. Every job consists of a fixed number of operations equal to the number of machines. Each operation is described by a pair of integers: the machine on which the operation must be processed, and the duration (processing time) of the operation on that machine.
\end{tcolorbox}

{Thus, an instance in this representation is illustrated below:}

\begin{tcolorbox}[colback=white,colframe=gray!75!black, fontupper=\scriptsize, enlarge bottom by=-1.5mm, enlarge top by=-2mm, boxsep=0mm]
10	10 

4	88	8	68	6	94	5	99	1	67	2	89	9	77	7	99	0	86	3	92

5	72	3	50	6	69	4	75	2	94	8	66	0	92	1	82	7	94	9	63

9	83	8	61	0	83	1	65	6	64	5	85	7	78	4	85	2	55	3	77

\dots
\end{tcolorbox}

{The same instance introduced in \textsf{natural language} for is:}

\begin{tcolorbox}[colback=white,colframe=gray!75!black, fontupper=\scriptsize, enlarge bottom by=-1.5mm, enlarge top by=-2mm, boxsep=0mm]
The job shop consists of 10 machines. There are 10 jobs to be scheduled. Regarding job 1, it must undergo the following operations: operation 4 with duration 88, \dots
\end{tcolorbox}

{The correspondent \textsf{code-like} format is:}

\begin{tcolorbox}[colback=white,colframe=gray!75!black, fontupper=\scriptsize, enlarge bottom by=-1.5mm, enlarge top by=-2mm, boxsep=0mm]
int: n\_machines = 10;;

int: n\_jobs = 10;

int: n\_tasks\_per\_job = n\_machines;

set of int: jobs = 1..n\_jobs;

set of int: tasks = 1..n\_tasks\_per\_job;

array [jobs, tasks] of 0..(n\_machines-1): job\_task\_machine = [4, 8, 6 \dots];

array [jobs, tasks] of int: job\_task\_duration = [88, 68, 94 \dots];
\end{tcolorbox}

\subsubsection{{Prompt Example}}

{The example below shows the standard direct-querying prompt instantiated for the \gls{jssp}, targeting the average operation duration (a \hardfeature{} feature) on a small instance in the \textsf{standard} format.}

\begin{tcolorbox}[colback=white,colframe=gray!75!black, fontupper=\scriptsize, enlarge bottom by=-1.5mm, enlarge top by=-2mm, boxsep=0mm]
\begin{verbatim}
You are given an instance of a combinatorial optimization problem: Job Shop
Scheduling Problem.
This is not a coding task. Do not return any code.
Extract the numeric value of the following feature from the instance:
- Feature name: feat_avg_duration
- Feature description: Average duration of all the tasks.
- Expected type: float
The instance is provided here: """
10  10
4  88  8  68  6  94  5  99  1  67  2  89  9  77  7  99  0  86  3  92
5  72  3  50  6  69  4  75  2  94  8  66  0  92  1  82  7  94  9  63
9  83  8  61  0  83  1  65  6  64  5  85  7  78  4  85  2  55  3  77
...
"""
Instructions:
- Return a JSON object only, with a single field "value", i.e., { "value": ... }.
- The "value" field should contain the numeric value of the required feature
  and should be of the expected type (i.e., float).
- If the value is unknown/undeterminable, return { "value": null }.
- No explanations, no extra fields.
Answer:
\end{verbatim}
\end{tcolorbox}

\subsection{Knapsack Problem}

\subsubsection{Problem Formulation}

{The \gls{kp} considers a finite set of items $\mathcal{I}=\{1,\dots,n\}$, with $|\mathcal{I}|=n$, each with a profit ($p_i$) and a weight ($w_i$), and a knapsack capacity ($C$). The goal is to choose items to be included in the knapsack that maximize total profit subject to the capacity constraint. Variants of \gls{kp} include the \emph{0--1} \gls{kp}, where each item can be selected at most once,  the \emph{bounded} \gls{kp}, where each item can be selected at most a given number of times, and the \emph{unbounded} \gls{kp}, where there is no upper limit on the number of times each item can be selected.}

{A possible mathematical formulation for the \emph{0--1} \gls{kp} is as follows:}
\begin{align}
    \max \quad & \sum_{i \in \mathcal{I}} {p_i x_i} \label{kp:obj}\\
    \text{s.t.} \quad \\
    & \sum_{i \in \mathcal{I}} {w_i x_i} \leq C \label{kp:capacity}\\
    & x_i \in \{0,1\} && \forall i \in \mathcal{I} \label{kp:domain}
\end{align}
{The model is defined using a binary decision variable $x_i$, which indicates whether item $i$ is selected. Objective~\eqref{kp:obj} maximizes the total profit. Constraint~\eqref{kp:capacity} enforces that the total weight of the selected items does not exceed the knapsack capacity $C$, while constraint~\eqref{kp:domain} defines the domain of the decision variables.}

\subsubsection{Features}

{The features considered are as follows:
\begin{itemize}
    \item \textbf{Capacity}: The capacity $C$ of the knapsack. This information is directly provided in the instance; therefore, it is an easy feature (\easyfeature).
    \item \textbf{Weight}: Metrics obtained by aggregating the item weights $w_i$, $i \in \mathcal{I}$. The associated complexity depends on the aggregation: minimum and maximum are mid-complexity features (\midfeature), as they require ordering operations, whereas the average is a hard feature (\hardfeature), as it requires explicit computation. They are calculated as follows:
    \begin{align}
    w_{\min} = &  \min_{i \in \mathcal{I}} w_i\\
    w_{\max} = &  \max_{i \in \mathcal{I}} w_i, \text{and}\\
    w_{\mathrm{avg}} = & \frac{1}{n} \sum_{i \in \mathcal{I}} w_i
    \end{align}
    \item \textbf{Profit}: Metrics obtained by aggregating the item weights $p_i$, $i \in \mathcal{I}$. The associated complexity depends on the aggregation: minimum and maximum are mid-complexity features (\midfeature), as they require ordering operations, whereas the average is a hard feature (\hardfeature), as it requires explicit computation.
    They are calculated as follows:
    \begin{align}
    p_{\min} = &  \min_{i \in \mathcal{I}} p_i\\
    p_{\max} = &  \max_{i \in \mathcal{I}} p_i, \text{and}\\
    p_{\mathrm{avg}} = & \frac{1}{n} \sum_{i \in \mathcal{I}} p_i
    \end{align}
    \item \textbf{Average efficiency}: Efficiency is calculated as the ratio of an item’s profit to its weight. It is a hard feature (\hardfeature), as it requires explicit computation. It is calculated as follows
    \begin{align}
        e_{\mathrm{avg}} = \frac{1}{n} \sum_{i \in \mathcal{I}} \frac{p_i}{w_i}
    \end{align}
    \item \textbf{Weight/profit/efficiency of item 1}: Characteristics of the first item, these are mid-complexity features (\midfeature). They are $w_1$, $p_1$, and $e_1=p_i/w_i$.
\end{itemize}}

\subsubsection{Instance Formats}

{The \textsf{standard} format for \gls{kp} instances is as follows:}

\begin{tcolorbox}[colback=white,colframe=gray!75!black, fontupper=\scriptsize, enlarge bottom by=-1.5mm, enlarge top by=-2mm, boxsep=0mm]
The instance is encoded as follows: the first row reports the name of the instance, and the second row provides the instance class or a placeholder. The third row specifies the maximum capacity of the knapsack. After that, each subsequent row represents an item, where the first value indicates its weight and the second its profit
\end{tcolorbox}

{Thus, an instance in this representation is illustrated below:}

\begin{tcolorbox}[colback=white,colframe=gray!75!black, fontupper=\scriptsize, enlarge bottom by=-1.5mm, enlarge top by=-2mm, boxsep=0mm]
AlmostStrong001

AlmostStronglyCorrelated

2593

777	928

701	842

552	715

855	1025

109	254

\dots
\end{tcolorbox}

{The same instance introduced in \textsf{natural language} for is:}

\begin{tcolorbox}[colback=white,colframe=gray!75!black, fontupper=\scriptsize, enlarge bottom by=-1.5mm, enlarge top by=-2mm, boxsep=0mm]
The instance is named AlmostStrong001 (AlmostStronglyCorrelated). The knapsack has a capacity of 2593. You have 1000 items to choose from. Item 1 has a profit of 928 and a weight of 777. \dots
\end{tcolorbox}

{The correspondent \textsf{code-like} format is:}

\begin{tcolorbox}[colback=white,colframe=gray!75!black, fontupper=\scriptsize, enlarge bottom by=-1.5mm, enlarge top by=-2mm, boxsep=0mm]
int: n = 1000; 

int: weight\_max = 2593;  

array[1..n] of int: weights = [777, 701, \dots];

array[1..n] of int: profits = [928, 842, \dots];
\end{tcolorbox}

\subsubsection{{Prompt Example}}

{The example below shows the standard direct-querying prompt instantiated for the \gls{kp}, targeting the average item efficiency (a \hardfeature{} feature, defined as the average ratio of profit to weight) on a small instance in the \textsf{standard} format.}

\begin{tcolorbox}[colback=white,colframe=gray!75!black, fontupper=\scriptsize, enlarge bottom by=-1.5mm, enlarge top by=-2mm, boxsep=0mm]
\begin{verbatim}
You are given an instance of a combinatorial optimization problem: 0-1 Knapsack
Problem.
This is not a coding task. Do not return any code.
Extract the numeric value of the following feature from the instance:
- Feature name: feat_avg_efficiency
- Feature description: The average efficiency of all the items available to
  choose from when building your knapsack. Efficiency is calculated as the
  ratio of an item's profit to its weight.
- Expected type: float
The instance is provided here: """
AlmostStrong001
AlmostStronglyCorrelated
2593
777  928
701  842
552  715
855  1025
109  254
...
"""
Instructions:
- Return a JSON object only, with a single field "value", i.e., { "value": ... }.
- The "value" field should contain the numeric value of the required feature
  and should be of the expected type (i.e., float).
- If the value is unknown/undeterminable, return { "value": null }.
- No explanations, no extra fields.
Answer:
\end{verbatim}
\end{tcolorbox}

%% file: appendix_prompting-scaling.tex
\section{{Direct-Querying: Scaling Regression and Inference Cost}}
\label{app:scaling-regression}

\subsection*{{Scaling regression on accuracy and abstention metrics}}

\Cref{sec:direct-querying-experiment} reports direct-querying performance in terms of the \textsc{Within 1\%} accuracy for clarity. To verify that the observed scaling patterns are not specific to that metric, we replicate the analysis on three additional indicators: \gls{mae}, exact-match accuracy (\textsc{Equals}) and the abstention rate (\textsc{Null rate}). For each metric and each (representation, feature-complexity) cell, we fit a linear regression of the metric against $\log_{2}(b)$, where $b$ is the model size in billions of parameters, using the five baseline configurations (\llama[3.2][3B], \llama[3.1][8B], \llama[3.3][70B], \gpt[20b], \gpt[120b]).
\gls{mae} slopes are reported in normalized form, dividing the raw slope by the mean \gls{mae} within the same cell; this makes them dimensionless and comparable across rows, and a value of $-0.5$ corresponds to the average \gls{mae} of the cell halving for each parameter doubling. Slopes for the remaining metrics are reported in their natural scale (accuracy points or abstention probability per parameter doubling), since they already lie in the unit interval and need no further normalization.

\Cref{tab:scaling-regression-summary} reports slope, $R^{2}$ and $p$-value for the three metrics. A negative slope on \gls{mae} indicates that prediction error decreases as model size doubles, while a positive slope on \textsc{Within 1\%} indicates accuracy improvements at the same rate; the two should therefore move in opposite directions. \textsc{Null rate} slopes are interpreted on their own: a positive slope captures an increasing tendency to abstain as scale grows.

\begin{table}[b]
\centering
\caption{{Scaling regression of three direct-querying metrics on $\log_{2}(\text{model size})$, computed on the baseline (\emph{none} or \emph{native}) configurations across the five evaluated models. Each cell reports slope, $R^{2}$ and $p$-value of the linear fit. The \gls{mae} slope is normalized by the cell mean and is therefore dimensionless: a value of $-0.5$ corresponds to the average error in the cell halving for each parameter doubling. \emph{Cells marked with} `--' \emph{correspond to configurations where the metric is degenerate (e.g., already saturated at the maximum value across all model sizes).}}}
\label{tab:scaling-regression-summary}
\footnotesize

\setlength{\tabcolsep}{4pt}
\begin{tabular}{lc rrr rrr rrr}
\toprule
 &  & \multicolumn{3}{c}{\textsc{Within 1\%}} & \multicolumn{3}{c}{\gls{mae} {\tiny (normalized)}} & \multicolumn{3}{c}{\textsc{Null rate}} \\
\cmidrule(lr){3-5}\cmidrule(lr){6-8}\cmidrule(lr){9-11}
\textbf{Repr.} & \textbf{Compl.} & Slope & $R^{2}$ & $p$ & Slope & $R^{2}$ & $p$ & Slope & $R^{2}$ & $p$ \\
\midrule
\multirow{3}{*}{\standard{}}
  & \easyfeature{} & $\phantom{-}0.057$  & $0.36$ & $0.29$  & $-0.20$ & $0.25$ & $0.39$  & $\phantom{-}0.001$  & $0.00$ & $0.94$  \\
  & \midfeature{}  & $\phantom{-}0.074$  & $0.79$ & $0.04$  & $-0.47$ & $0.67$ & $0.09$  & $\phantom{-}0.014$  & $0.07$ & $0.66$  \\
  & \hardfeature{} & $\phantom{-}0.059$  & $0.58$ & $0.14$  & $-0.40$ & $0.76$ & $0.06$  & $\phantom{-}0.079$  & $0.27$ & $0.38$  \\
\addlinespace
\multirow{3}{*}{\code{}}
  & \easyfeature{} & $\phantom{-}0.000$  & $0.51$ & $0.17$  & $-0.73$ & $0.51$ & $0.17$  & --                  & --     & --      \\
  & \midfeature{}  & $\phantom{-}0.083$  & $0.90$ & $0.01$  & $-0.59$ & $0.81$ & $0.04$  & $\phantom{-}0.010$  & $0.05$ & $0.71$  \\
  & \hardfeature{} & $\phantom{-}0.033$  & $0.80$ & $0.04$  & $-0.46$ & $0.76$ & $0.05$  & $\phantom{-}0.073$  & $0.26$ & $0.38$  \\
\addlinespace
\multirow{3}{*}{\nlp{}}
  & \easyfeature{} & --                  & --     & --      & --      & --     & --      & --                  & --     & --      \\
  & \midfeature{}  & $\phantom{-}0.046$  & $0.90$ & $0.01$  & $-0.46$ & $0.40$ & $0.25$  & $\phantom{-}0.005$  & $0.02$ & $0.84$  \\
  & \hardfeature{} & $\phantom{-}0.041$  & $0.61$ & $0.12$  & $-0.50$ & $0.77$ & $0.05$  & $\phantom{-}0.072$  & $0.25$ & $0.39$  \\
\bottomrule
\end{tabular}

\end{table}

The pattern is consistent across the three metrics: \textsc{Within 1\%} slopes are non-negative everywhere and reach statistical significance ($p < 0.05$) on the \midfeature{}--\code{} and \midfeature{}--\nlp{} cells, while the corresponding \gls{mae} slopes are negative on the same cells with $R^{2}$ above $0.65$. \gls{mae} normalization makes the magnitudes directly comparable: most cells see the average error shrink by between $40\%$ and $60\%$ for each parameter doubling, with the notable exception of \standard{}--\easyfeature{} ($-20\%$), where increased capacity alone appears to compensate less effectively for the rigid input format. Cells with degenerate fits correspond to configurations where direct querying is already saturated across the five model sizes (typical for \easyfeature{} features under \nlp{} representation). \textsc{Null rate} exhibits a different shape: the slope is essentially zero or weakly positive for \easyfeature{} and \midfeature{} features, and becomes the largest of the three at the \hardfeature{} level (around $+0.07$ per parameter doubling, $R^{2} \approx 0.25$ across the three representations). This is the regression-level counterpart of the abstention pattern discussed in \Cref{sec:direct-querying-experiment}: larger reasoning-oriented models do not necessarily improve their \hardfeature{} accuracy, but they do increase their tendency to abstain.

\subsection*{{Inference cost reference}}

\Cref{tab:inference-time} reports the mean generation time per feature--instance, broken down by model, prompting mode, and feature complexity. The values are aggregated over problems, representations, and random seeds. They constitute the numeric reference for the cost discussion in \Cref{sec:direct-querying-experiment} and complement \Cref{fig:scaling-time,fig:diff-cot-time}, which provide the corresponding graphical view for individual problems and representations.

\begin{table}[ht]
\centering
\caption{{Mean generation time per feature--instance (seconds) by model, prompting mode, and feature complexity, averaged over problems and representations. \emph{Llama} models are evaluated under the standard (\emph{none}) and Chain-of-Thought (\emph{prompted}) prompt variants. \emph{\gpt} models are evaluated under the native reasoning interface (\emph{native}). The reduced inference time of \gpt[120b] compared to \llama[3.3][70B] reflects the mixture-of-experts architecture, which activates only a fraction of the parameters per token.}}
\label{tab:inference-time}
\footnotesize

\begin{tabular}{ll ccc}
\toprule
\textbf{Model} & \textbf{Mode} & \easyfeature{} & \midfeature{} & \hardfeature{} \\
\midrule
\multirow{2}{*}{\llama[3.2][3B]}  & \emph{none}     & $0.46$ & $0.22$ & $0.43$ \\
                               & \emph{prompted} & $0.78$ & $0.51$ & $0.82$ \\
\addlinespace
\multirow{2}{*}{\llama[3.1][8B]}  & \emph{none}     & $0.89$ & $0.45$ & $0.86$ \\
                               & \emph{prompted} & $1.05$ & $0.60$ & $1.15$ \\
\addlinespace
\multirow{2}{*}{\llama[3.3][70B]} & \emph{none}     & $2.28$ & $1.22$ & $2.16$ \\
                               & \emph{prompted} & $3.15$ & $1.99$ & $3.52$ \\
\addlinespace
\gpt[20b]                      & \emph{native}   & $0.96$ & $0.66$ & $1.29$ \\
\gpt[120b]                     & \emph{native}   & $0.65$ & $0.38$ & $0.71$ \\
\bottomrule
\end{tabular}

\end{table}

The aggregate Chain-of-Thought-to-standard time ratio across feature complexities is $1.87\times$ on \llama[3.2][3B], $1.27\times$ on \llama[3.1][8B], and $1.53\times$ on \llama[3.3][70B], indicating that the smallest model pays the largest relative cost for explicit reasoning. As discussed in \Cref{sec:direct-querying-experiment}, the same model is also the only one for which Chain-of-Thought consistently degrades accuracy, so the smallest model is doubly penalized: it incurs the highest relative time overhead without any compensating accuracy gain.

%% file: appendix_feature-probing.tex
\section{Feature Probing Results}
\label{app:feature-probing-complete-results}

Tables~\ref{tab:probing-metrics-bpp}--\ref{tab:probing-metrics-kp}
report the \gls{mae} obtained for each combination of problem, instance
representation, feature complexity, and model. Each table contrasts direct
querying (in the standard \emph{none}/\emph{native} mode used as baseline)
with probing under the three regressors considered in the study
(Linear, LightGBM, MLP). Probing values use the mean pooling strategy as
default; the same qualitative patterns hold under max and last pooling, with
the per-cell numerical detail available in the companion repository.

\begin{landscape}
\input{tab-probing-prompting-metrics-BPP}
\input{tab-probing-prompting-metrics-GCP}
\input{tab-probing-prompting-metrics-JSP}
\input{tab-probing-prompting-metrics-KP}
\end{landscape}

%% file: tab-probing-prompting-metrics-BPP.tex
\begin{table}[t]
  \centering
  \caption{{Direct querying (\emph{none}/\emph{native} mode) and probing (mean pooling) MAE for BPP, broken down by representation, feature complexity, and model. Each cell reports the MAE on the probing validation split. \emph{Cells marked with} `--' \emph{denote configurations not defined for the problem.}}}
  \label{tab:probing-metrics-bpp}
  \footnotesize
  \setlength{\tabcolsep}{4pt}
  \begin{tabular}{@{}lllrrrrrrrrr@{}}
    \toprule
     &  &  & \multicolumn{3}{c}{\standard{}} & \multicolumn{3}{c}{\code{}} & \multicolumn{3}{c}{\nlp{}} \\
    \cmidrule(lr){4-6}\cmidrule(lr){7-9}\cmidrule(lr){10-12}
     &  &  & \multicolumn{1}{c}{\easyfeature{}} & \multicolumn{1}{c}{\midfeature{}} & \multicolumn{1}{c}{\hardfeature{}} & \multicolumn{1}{c}{\easyfeature{}} & \multicolumn{1}{c}{\midfeature{}} & \multicolumn{1}{c}{\hardfeature{}} & \multicolumn{1}{c}{\easyfeature{}} & \multicolumn{1}{c}{\midfeature{}} & \multicolumn{1}{c}{\hardfeature{}} \\
    \midrule
    \multirow{5}{*}{Direct Querying} &  & Llama\,3.2 (3B) & 217.47 & 507.58 & 159.51 & 0.000 & 113.27 & 154.63 & 0.000 & 29.94 & 115.00 \\
     &  & Llama\,3.1 (8B) & 155.68 & 406.20 & 152.56 & 0.000 & 107.02 & 114.90 & 0.000 & 14.38 & 36.54 \\
     &  & Llama\,3.3 (70B) & 307.01 & 87.70 & 22.88 & 0.000 & 0.000 & 41.87 & 0.000 & 0.000 & 23.10 \\
     &  & GPT-OSS (20B) & 152.12 & 34.37 & -- & 0.000 & 0.000 & -- & 0.000 & 0.000 & -- \\
     &  & GPT-OSS (120B) & 314.21 & 58.21 & 21.12 & 0.000 & 0.000 & 5.577 & 0.000 & 0.000 & 4.390 \\
    \midrule
    \multirow{5}{*}{Probing} & \multirow{5}{*}{Linear} & Llama\,3.2 (3B) & 2.289 & 2.960 & 1.265 & 2.963 & 2.901 & 1.376 & 3.947 & 3.235 & 1.549 \\
     &  & Llama\,3.1 (8B) & 3.202 & 2.993 & 1.905 & 3.952 & 3.447 & 1.565 & 4.342 & 3.732 & 2.128 \\
     &  & Llama\,3.3 (70B) & 4.029 & 4.863 & 3.261 & 3.996 & 4.895 & 3.530 & 4.647 & 5.062 & 3.369 \\
     &  & GPT-OSS (20B) & 4.049 & 4.096 & 2.109 & 4.236 & 4.043 & 2.149 & 4.240 & 4.993 & 4.444 \\
     &  & GPT-OSS (120B) & 4.165 & 3.405 & 1.866 & 4.032 & 3.890 & 2.207 & 5.455 & 5.743 & 4.023 \\
    \midrule
     & \multirow{5}{*}{LightGBM} & Llama\,3.2 (3B) & 3.007 & 4.911 & 4.562 & 4.346 & 5.357 & 5.045 & 6.424 & 7.738 & 6.388 \\
     &  & Llama\,3.1 (8B) & 3.399 & 5.531 & 6.013 & 3.580 & 6.887 & 6.995 & 5.647 & 7.850 & 6.942 \\
     &  & Llama\,3.3 (70B) & 5.116 & 6.809 & 7.499 & 4.303 & 9.217 & 9.041 & 4.070 & 8.023 & 6.905 \\
     &  & GPT-OSS (20B) & 13.69 & 11.57 & 10.48 & 8.290 & 8.924 & 8.675 & 5.704 & 7.929 & 7.825 \\
     &  & GPT-OSS (120B) & 3.811 & 5.654 & 5.441 & 4.639 & 7.204 & 6.286 & 9.012 & 11.82 & 10.54 \\
    \midrule
     & \multirow{5}{*}{MLP} & Llama\,3.2 (3B) & 43.93 & 25.36 & 15.23 & 46.53 & 28.60 & 16.92 & 104.79 & 69.31 & 35.75 \\
     &  & Llama\,3.1 (8B) & 43.97 & 28.26 & 18.22 & 61.10 & 34.32 & 21.04 & 117.37 & 80.14 & 46.75 \\
     &  & Llama\,3.3 (70B) & 41.26 & 25.90 & 19.18 & 44.46 & 35.72 & 24.49 & 71.56 & 44.33 & 30.10 \\
     &  & GPT-OSS (20B) & 60.60 & 36.42 & 24.44 & 62.33 & 37.53 & 27.67 & 120.61 & 89.09 & 69.52 \\
     &  & GPT-OSS (120B) & 54.96 & 33.02 & 25.67 & 54.32 & 36.87 & 26.84 & 105.58 & 72.90 & 44.51 \\
    \bottomrule
  \end{tabular}
\end{table}

%% file: tab-probing-prompting-metrics-GCP.tex
\begin{table}[t]
  \centering
  \caption{{Direct querying (\emph{none}/\emph{native} mode) and probing (mean pooling) MAE for GCP, broken down by representation, feature complexity, and model. Each cell reports the MAE on the probing validation split. \emph{Cells marked with} `--' \emph{denote configurations not defined for the problem.}}}
  \label{tab:probing-metrics-gcp}
  \footnotesize
  \setlength{\tabcolsep}{4pt}
  \begin{tabular}{@{}lllrrrrrrrrr@{}}
    \toprule
     &  &  & \multicolumn{3}{c}{\standard{}} & \multicolumn{3}{c}{\code{}} & \multicolumn{3}{c}{\nlp{}} \\
    \cmidrule(lr){4-6}\cmidrule(lr){7-9}\cmidrule(lr){10-12}
     &  &  & \multicolumn{1}{c}{\easyfeature{}} & \multicolumn{1}{c}{\midfeature{}} & \multicolumn{1}{c}{\hardfeature{}} & \multicolumn{1}{c}{\easyfeature{}} & \multicolumn{1}{c}{\midfeature{}} & \multicolumn{1}{c}{\hardfeature{}} & \multicolumn{1}{c}{\easyfeature{}} & \multicolumn{1}{c}{\midfeature{}} & \multicolumn{1}{c}{\hardfeature{}} \\
    \midrule
    \multirow{5}{*}{Direct Querying} &  & Llama\,3.2 (3B) & 941.34 & -- & 29.66 & 0.000 & -- & 59.14 & 0.000 & -- & 220.02 \\
     &  & Llama\,3.1 (8B) & 33.37 & -- & 66.89 & 0.000 & -- & 38.42 & 0.000 & -- & 43.10 \\
     &  & Llama\,3.3 (70B) & 0.000 & -- & 10.00 & 0.000 & -- & 11.90 & 0.000 & -- & 13.06 \\
     &  & GPT-OSS (20B) & 0.000 & -- & 0.004 & 0.000 & -- & 0.000 & 0.000 & -- & 0.005 \\
     &  & GPT-OSS (120B) & 0.000 & -- & 0.007 & 0.000 & -- & 0.064 & 0.000 & -- & 0.094 \\
    \midrule
    \multirow{5}{*}{Probing} & \multirow{5}{*}{Linear} & Llama\,3.2 (3B) & 34.10 & -- & 1.974 & 31.06 & -- & 1.956 & 20.06 & -- & 1.789 \\
     &  & Llama\,3.1 (8B) & 34.29 & -- & 1.942 & 31.59 & -- & 2.017 & 25.99 & -- & 1.857 \\
     &  & Llama\,3.3 (70B) & 29.93 & -- & 1.867 & 34.60 & -- & 1.949 & 23.86 & -- & 1.830 \\
     &  & GPT-OSS (20B) & 27.14 & -- & 1.774 & 28.01 & -- & 1.891 & 17.00 & -- & 1.674 \\
     &  & GPT-OSS (120B) & 30.96 & -- & 1.787 & 29.54 & -- & 1.941 & 24.48 & -- & 1.779 \\
    \midrule
     & \multirow{5}{*}{LightGBM} & Llama\,3.2 (3B) & 68.06 & -- & 2.553 & 50.39 & -- & 2.429 & 54.94 & -- & 2.390 \\
     &  & Llama\,3.1 (8B) & 64.10 & -- & 2.581 & 56.60 & -- & 2.595 & 54.13 & -- & 2.423 \\
     &  & Llama\,3.3 (70B) & 64.34 & -- & 2.613 & 63.18 & -- & 2.566 & 49.51 & -- & 2.418 \\
     &  & GPT-OSS (20B) & 51.23 & -- & 2.301 & 65.45 & -- & 2.626 & 47.02 & -- & 2.360 \\
     &  & GPT-OSS (120B) & 57.42 & -- & 2.381 & 53.49 & -- & 2.592 & 68.97 & -- & 2.706 \\
    \midrule
     & \multirow{5}{*}{MLP} & Llama\,3.2 (3B) & 352.99 & -- & 3.668 & 349.82 & -- & 3.422 & 356.41 & -- & 4.160 \\
     &  & Llama\,3.1 (8B) & 355.10 & -- & 3.634 & 352.64 & -- & 3.680 & 359.66 & -- & 4.060 \\
     &  & Llama\,3.3 (70B) & 346.97 & -- & 3.526 & 338.29 & -- & 3.410 & 348.24 & -- & 3.292 \\
     &  & GPT-OSS (20B) & 315.46 & -- & 3.422 & 349.20 & -- & 3.381 & 352.06 & -- & 4.356 \\
     &  & GPT-OSS (120B) & 349.78 & -- & 3.525 & 344.19 & -- & 3.496 & 349.08 & -- & 3.223 \\
    \bottomrule
  \end{tabular}
\end{table}

%% file: tab-probing-prompting-metrics-JSP.tex
\begin{table}[t]
  \centering
  \caption{{Direct querying (\emph{none}/\emph{native} mode) and probing (mean pooling) MAE for JSP, broken down by representation, feature complexity, and model. Each cell reports the MAE on the probing validation split. \emph{Cells marked with} `--' \emph{denote configurations not defined for the problem.}}}
  \label{tab:probing-metrics-jsp}
  \footnotesize
  \setlength{\tabcolsep}{4pt}
  \begin{tabular}{@{}lllrrrrrrrrr@{}}
    \toprule
     &  &  & \multicolumn{3}{c}{\standard{}} & \multicolumn{3}{c}{\code{}} & \multicolumn{3}{c}{\nlp{}} \\
    \cmidrule(lr){4-6}\cmidrule(lr){7-9}\cmidrule(lr){10-12}
     &  &  & \multicolumn{1}{c}{\easyfeature{}} & \multicolumn{1}{c}{\midfeature{}} & \multicolumn{1}{c}{\hardfeature{}} & \multicolumn{1}{c}{\easyfeature{}} & \multicolumn{1}{c}{\midfeature{}} & \multicolumn{1}{c}{\hardfeature{}} & \multicolumn{1}{c}{\easyfeature{}} & \multicolumn{1}{c}{\midfeature{}} & \multicolumn{1}{c}{\hardfeature{}} \\
    \midrule
    \multirow{5}{*}{Direct Querying} &  & Llama\,3.2 (3B) & 20.20 & 129.03 & 18.26 & 0.000 & 21.97 & 1.791 & 0.000 & 15.98 & 1.452 \\
     &  & Llama\,3.1 (8B) & 17.87 & 89.86 & 12.45 & 0.000 & 11.87 & 0.000 & 0.000 & 71.25 & 0.235 \\
     &  & Llama\,3.3 (70B) & 21.87 & 58.37 & 13.70 & 0.000 & 0.000 & 0.000 & 0.000 & 2.985 & 0.000 \\
     &  & GPT-OSS (20B) & 0.000 & 0.084 & 0.777 & 0.000 & 0.000 & 0.000 & 0.000 & 0.000 & 0.000 \\
     &  & GPT-OSS (120B) & 0.602 & 4.029 & 2.656 & 0.000 & 0.000 & 0.000 & 0.000 & 0.000 & 0.000 \\
    \midrule
    \multirow{5}{*}{Probing} & \multirow{5}{*}{Linear} & Llama\,3.2 (3B) & 0.597 & 6.392 & 0.000 & 0.535 & 4.844 & 0.000 & 0.522 & 5.734 & 0.000 \\
     &  & Llama\,3.1 (8B) & 0.656 & 9.048 & 0.000 & 0.520 & 4.657 & 0.000 & 0.482 & 6.278 & 0.000 \\
     &  & Llama\,3.3 (70B) & 0.661 & 9.041 & 0.000 & 0.678 & 4.643 & 0.000 & 0.511 & 8.036 & 0.000 \\
     &  & GPT-OSS (20B) & 0.642 & 7.234 & 0.000 & 0.556 & 2.801 & 0.000 & 0.577 & 3.859 & 0.000 \\
     &  & GPT-OSS (120B) & 0.565 & 6.215 & 0.000 & 0.598 & 4.122 & 0.000 & 0.689 & 7.645 & 0.000 \\
    \midrule
     & \multirow{5}{*}{LightGBM} & Llama\,3.2 (3B) & 1.460 & 12.68 & 0.000 & 1.452 & 7.567 & 0.000 & 1.460 & 9.377 & 0.000 \\
     &  & Llama\,3.1 (8B) & 1.203 & 9.401 & 0.000 & 1.359 & 8.357 & 0.000 & 1.034 & 9.382 & 0.000 \\
     &  & Llama\,3.3 (70B) & 1.414 & 6.883 & 0.000 & 1.782 & 9.026 & 0.000 & 1.284 & 10.58 & 0.000 \\
     &  & GPT-OSS (20B) & 1.871 & 10.83 & 0.000 & 1.737 & 7.027 & 0.000 & 1.274 & 6.880 & 0.000 \\
     &  & GPT-OSS (120B) & 2.298 & 8.978 & 0.000 & 1.640 & 10.43 & 0.000 & 1.631 & 9.928 & 0.000 \\
    \midrule
     & \multirow{5}{*}{MLP} & Llama\,3.2 (3B) & 3.360 & 35.36 & 0.016 & 4.050 & 39.08 & 0.011 & 4.133 & 39.13 & 0.007 \\
     &  & Llama\,3.1 (8B) & 3.079 & 30.79 & 0.012 & 3.834 & 56.03 & 0.014 & 3.155 & 33.16 & 0.012 \\
     &  & Llama\,3.3 (70B) & 3.377 & 24.07 & 0.014 & 4.220 & 34.74 & 0.012 & 3.423 & 32.17 & 0.011 \\
     &  & GPT-OSS (20B) & 4.893 & 32.69 & 0.014 & 4.138 & 37.36 & 0.012 & 6.750 & 38.79 & 0.010 \\
     &  & GPT-OSS (120B) & 4.397 & 27.91 & 0.016 & 4.256 & 37.59 & 0.018 & 5.016 & 33.94 & 0.010 \\
    \bottomrule
  \end{tabular}
\end{table}

%% file: tab-probing-prompting-metrics-KP.tex
\begin{table}[t]
  \centering
  \caption{{Direct querying (\emph{none}/\emph{native} mode) and probing (mean pooling) MAE for KP, broken down by representation, feature complexity, and model. Each cell reports the MAE on the probing validation split. \emph{Cells marked with} `--' \emph{denote configurations not defined for the problem.}}}
  \label{tab:probing-metrics-kp}
  \footnotesize
  \setlength{\tabcolsep}{4pt}
  \begin{tabular}{@{}lllrrrrrrrrr@{}}
    \toprule
     &  &  & \multicolumn{3}{c}{\standard{}} & \multicolumn{3}{c}{\code{}} & \multicolumn{3}{c}{\nlp{}} \\
    \cmidrule(lr){4-6}\cmidrule(lr){7-9}\cmidrule(lr){10-12}
     &  &  & \multicolumn{1}{c}{\easyfeature{}} & \multicolumn{1}{c}{\midfeature{}} & \multicolumn{1}{c}{\hardfeature{}} & \multicolumn{1}{c}{\easyfeature{}} & \multicolumn{1}{c}{\midfeature{}} & \multicolumn{1}{c}{\hardfeature{}} & \multicolumn{1}{c}{\easyfeature{}} & \multicolumn{1}{c}{\midfeature{}} & \multicolumn{1}{c}{\hardfeature{}} \\
    \midrule
    \multirow{5}{*}{Direct Querying} &  & Llama\,3.2 (3B) & 320.1K & 1.2K & 386.91 & 69.66 & 77.7K & 284.25 & 0.000 & 16.2K & 240.48 \\
     &  & Llama\,3.1 (8B) & 289.8K & 151.43 & 357.94 & 0.000 & 45.9K & 268.54 & 0.000 & 39.9K & 251.33 \\
     &  & Llama\,3.3 (70B) & 317.7K & 139.27 & 76.17 & 0.000 & 12.15 & 55.51 & 0.000 & 1.007 & 43.24 \\
     &  & GPT-OSS (20B) & 10.6K & 25.76 & 40.53 & 0.000 & 0.000 & 1.985 & 0.000 & 0.000 & 0.860 \\
     &  & GPT-OSS (120B) & 5.1K & 61.15 & -- & 0.000 & 0.005 & 4.100 & 0.000 & 0.000 & 0.000 \\
    \midrule
    \multirow{5}{*}{Probing} & \multirow{5}{*}{Linear} & Llama\,3.2 (3B) & 40.3K & 55.70 & 6.104 & 33.0K & 70.10 & 7.665 & 28.5K & 70.75 & 5.873 \\
     &  & Llama\,3.1 (8B) & 42.5K & 53.49 & 6.136 & 28.3K & 70.95 & 6.639 & 24.5K & 66.38 & 5.913 \\
     &  & Llama\,3.3 (70B) & 30.7K & 57.81 & 6.354 & 25.1K & 64.04 & 6.812 & 21.9K & 57.19 & 5.971 \\
     &  & GPT-OSS (20B) & 44.8K & 68.33 & 6.811 & 30.7K & 73.73 & 6.919 & 46.1K & 70.27 & 6.071 \\
     &  & GPT-OSS (120B) & 34.3K & 63.77 & 4.934 & 33.2K & 65.65 & 5.779 & 34.8K & 63.66 & 5.904 \\
    \midrule
     & \multirow{5}{*}{LightGBM} & Llama\,3.2 (3B) & 45.8K & 68.99 & 12.10 & 31.2K & 77.88 & 9.217 & 44.3K & 80.37 & 12.49 \\
     &  & Llama\,3.1 (8B) & 51.4K & 68.87 & 11.41 & 33.7K & 77.10 & 9.727 & 30.2K & 77.98 & 10.02 \\
     &  & Llama\,3.3 (70B) & 62.9K & 72.00 & 14.14 & 39.1K & 72.24 & 10.53 & 32.1K & 70.19 & 10.13 \\
     &  & GPT-OSS (20B) & 69.0K & 78.23 & 14.32 & 30.7K & 77.71 & 8.267 & 49.2K & 77.12 & 10.35 \\
     &  & GPT-OSS (120B) & 49.0K & 79.12 & 11.48 & 40.0K & 77.96 & 9.788 & 52.8K & 77.32 & 14.79 \\
    \midrule
     & \multirow{5}{*}{MLP} & Llama\,3.2 (3B) & 317.4K & 111.26 & 52.38 & 316.8K & 106.64 & 53.24 & 316.5K & 109.69 & 55.96 \\
     &  & Llama\,3.1 (8B) & 317.2K & 111.65 & 53.28 & 316.0K & 107.33 & 55.47 & 316.5K & 109.04 & 55.03 \\
     &  & Llama\,3.3 (70B) & 316.4K & 109.24 & 50.84 & 315.6K & 106.81 & 53.52 & 316.4K & 107.80 & 53.91 \\
     &  & GPT-OSS (20B) & 317.0K & 126.63 & 55.27 & 317.5K & 107.59 & 53.48 & 316.7K & 108.84 & 56.01 \\
     &  & GPT-OSS (120B) & 316.9K & 112.59 & 53.38 & 316.7K & 107.76 & 53.18 & 317.1K & 111.92 & 58.66 \\
    \bottomrule
  \end{tabular}
\end{table}

%% file: appendix_statistical-analyses.tex
\section{Complete Set of Statistical Analysis Tables}
\label{app:statistical-analyses}

This appendix reports the full results of the mixed linear models used in
\Cref{sec:algorithm-selection-experiment} to assess the effects of classifier,
feature representation, pooling strategy, and \gls{llm} model on
algorithm-selection accuracy. All models include random intercepts at the
block level, defined as the (\emph{problem}, \emph{replicate}) pair, with
fixed-effect estimates reported alongside standard errors, $z$-statistics,
$p$-values, and 95\% confidence intervals.
\begin{itemize}
\item \Cref{tab:mixed_model_classifier_effects} reports classifier and model
      effects on the full pool of observations
      (\Cref{sec:algorithm-selection-experiment}, classifier comparison).
\item \Cref{tab:isa_vs_hc} compares \gls{isa}-based and handcrafted features
      on the baseline subset
      (\Cref{sec:algorithm-selection-experiment}, baseline comparison).
\item \Cref{tab:full_factorial_mixed_model} presents the full-factorial mixed
      model including representation, pooling, classifier, and \gls{llm}
      model as fixed factors
      (\Cref{sec:algorithm-selection-experiment}, full-factorial analysis).
\item \Cref{tab:mixed_model_isa_standard} reports the per-problem comparison
      between \gls{isa} and the \gls{llm}-derived \standard{} representation
      with \lightGBM{}.
\end{itemize}

\begin{landscape}
\input{tab-mixed_model_classifier_effects}
\input{tab-isa_vs_hc}
\input{tab-full_factorial_mixed_model}
\input{tab-mixed_model_isa_standard}
\end{landscape}

%% file: tab-mixed_model_classifier_effects.tex
\begin{table}[t]
\centering
\caption{{Mixed linear model for classifier and model effects, specified as \texttt{accuracy $\sim$ C(classifier, ref='MostFrequentClassifier') + C(model, ref='baseline')}, with random intercept at the (\emph{problem}, \emph{replicate}) block level. The \emph{baseline} level of the \emph{model} factor groups handcrafted and \gls{isa} feature sets, which are model-independent. The analysis is based on $n = 3760$ observations and $g = 20$ groups.}}
\label{tab:mixed_model_classifier_effects}
\footnotesize
\setlength{\tabcolsep}{4pt}
\begin{tabular}{lccrccc}
\toprule
\textbf{Parameter} & \textbf{Coef.} & \textbf{Std.Err.} & \multicolumn{1}{c}{\textbf{z}} & \textbf{p-value} & \textbf{[0.025} & \textbf{0.975]} \\
\midrule
Intercept (\mostFrequent{}, baseline)                        & $\phantom{-}0.793$  & 0.040 & $\phantom{-}19.831$ & $<0.001$ & $\phantom{-}0.714$  & $\phantom{-}0.871$  \\
Classifier: \lightGBM{} (vs \mostFrequent{})                 & $\phantom{-}0.041$  & 0.002 & $\phantom{-}20.822$ & $<0.001$ & $\phantom{-}0.037$  & $\phantom{-}0.044$  \\
Classifier: \logisticRegression{} (vs \mostFrequent{})       & $\phantom{-}0.014$  & 0.002 & $\phantom{-}\phantom{0}7.278$ & $<0.001$ & $\phantom{-}0.010$  & $\phantom{-}0.018$  \\
Classifier: \mlpClassifier{} (vs \mostFrequent{})            & $-0.013$ & 0.002 & $-6.435$ & $<0.001$ & $-0.016$ & $-0.009$ \\
Model: \llama[3.1][8B]   (vs baseline)                          & $\phantom{-}0.018$  & 0.004 & $\phantom{-}\phantom{0}4.796$ & $<0.001$ & $\phantom{-}0.010$  & $\phantom{-}0.025$  \\
Model: \llama[3.2][3B]   (vs baseline)                          & $\phantom{-}0.019$  & 0.004 & $\phantom{-}\phantom{0}5.114$ & $<0.001$ & $\phantom{-}0.012$  & $\phantom{-}0.026$  \\
Model: \llama[3.3][70B]  (vs baseline)                          & $\phantom{-}0.016$  & 0.004 & $\phantom{-}\phantom{0}4.455$ & $<0.001$ & $\phantom{-}0.009$  & $\phantom{-}0.024$  \\
Model: \gpt[120b]   (vs baseline)                          & $\phantom{-}0.019$  & 0.004 & $\phantom{-}\phantom{0}5.155$ & $<0.001$ & $\phantom{-}0.012$  & $\phantom{-}0.026$  \\
Model: \gpt[20b]    (vs baseline)                          & $\phantom{-}0.016$  & 0.004 & $\phantom{-}\phantom{0}4.212$ & $<0.001$ & $\phantom{-}0.008$  & $\phantom{-}0.023$  \\
\midrule
\textit{Group variance} ($\sigma^{2}$)                       & $\phantom{-}0.032$ & (0.238) & & & & \\
\bottomrule
\end{tabular}

\vspace{0.6em}
\begin{tabular}{lll l}
\toprule
\textbf{Effect} & \textbf{Test} & \textbf{Statistic (df)} & \textbf{p-value} \\
\midrule
Classifier (controlling for model) & LRT & $\chi^{2}(3) = 742.403$ & $1.34\times 10^{-160}$ \\
\bottomrule
\end{tabular}
\end{table}

%% file: tab-isa_vs_hc.tex
\begin{table}[t]
\centering
\caption{{Mixed linear model and post-hoc tests comparing \gls{isa}-based and handcrafted features on the baseline subset, specified as \texttt{accuracy $\sim$ C(classifier, ref='LogisticRegression') + C(representation, ref='hc')}, with random intercept at the block level. The analysis is based on $n = 80$ observations and $g = 20$ groups.}}
\label{tab:isa_vs_hc}
\footnotesize
\setlength{\tabcolsep}{4pt}
\begin{tabular}{lccrccc}
\toprule
\textbf{Parameter} & \textbf{Coef.} & \textbf{Std.Err.} & \multicolumn{1}{c}{\textbf{z}} & \textbf{p-value} & \textbf{[0.025} & \textbf{0.975]} \\
\midrule
Intercept (\logisticRegression{}, Handcrafted)            & $\phantom{-}0.815$ & 0.038 & $\phantom{-}21.679$ & $<0.001$ & $\phantom{-}0.741$  & $\phantom{-}0.888$  \\
Classifier: \lightGBM{} (vs \logisticRegression{})        & $\phantom{-}0.024$ & 0.008 & $\phantom{-}\phantom{0}2.893$ & $\phantom{<}0.004$ & $\phantom{-}0.008$  & $\phantom{-}0.040$  \\
Representation: \gls{isa} (vs Handcrafted)                & $\phantom{-}0.007$ & 0.008 & $\phantom{-}\phantom{0}0.908$ & $\phantom{<}0.364$ & $-0.009$ & $\phantom{-}0.023$  \\
\midrule
\textit{Group variance} ($\sigma^{2}$)                    & $\phantom{-}0.027$ & (0.276) & & & & \\
\bottomrule
\end{tabular}
\vspace{0.8em}
\begin{tabular}{lll p{0.41\textwidth} l}
\toprule
\textbf{Comparison} & \textbf{Test} & \textbf{Statistic (df)} & \textbf{p-value / CI} & \textbf{Interpretation} \\
\midrule
\gls{isa} vs Handcrafted & LRT & $\chi^{2}(1) = 0.819$ & $p = 0.365$ & Not significant \\
\gls{isa} vs Handcrafted & Tukey HSD & $M_{\text{diff}} = 0.0074$ & $p_{\text{adj}} = 0.848$,\newline $\mathrm{CI}_{95\%}=[-0.069,\,0.084]$ & Not significant \\
\gls{isa} vs Handcrafted ($\varepsilon=0.01$) & Wilcoxon TOST & $M_{\text{diff}} = 0.0000$ & $p_{\text{low}} = 0.0063$, $p_{\text{high}} = 0.0500$,\newline $\mathrm{CI}_{95\%}=[-0.0471,\,0.0755]$ & Not equivalent \\
\bottomrule
\end{tabular}
\end{table}

%% file: tab-full_factorial_mixed_model.tex
\begin{table}[t]
\centering
\caption{{Full-factorial mixed linear model and corresponding likelihood-ratio (LRT) comparisons, specified as \texttt{accuracy $\sim$ C(classifier, ref='LogisticRegression') + C(model, ref='\llama[3.2][3B]') + C(representation, ref='standard') $\times$ C(pooling, ref='mean')}, with random intercept at the (\emph{problem}, \emph{replicate}) block level. Analysis restricted to the two best-performing classifiers (\lightGBM{} and \logisticRegression{}). The analysis is based on $n = 1800$ observations and $g = 20$ groups.}}
\label{tab:full_factorial_mixed_model}
\footnotesize
\setlength{\tabcolsep}{4pt}
\begin{tabular}{lccrccc}
\toprule
\textbf{Parameter} & \textbf{Coef.} & \textbf{Std.Err.} & \multicolumn{1}{c}{\textbf{z}} & \textbf{p-value} & \textbf{[0.025} & \textbf{0.975]} \\
\midrule
Intercept (\logisticRegression{}, \llama[3.2][3B], \standard{}, Mean)                                  & $\phantom{-}0.830$  & 0.037 & $\phantom{-}22.659$ & $<0.001$ & $\phantom{-}0.758$  & $\phantom{-}0.902$  \\
Classifier: \lightGBM{} (vs \logisticRegression{})                                                   & $\phantom{-}0.027$  & 0.002 & $\phantom{-}16.761$ & $<0.001$ & $\phantom{-}0.023$  & $\phantom{-}0.030$  \\
Model: \llama[3.1][8B]  (vs \llama[3.2][3B])                                                              & $-0.001$ & 0.003 & $-0.483$ & $\phantom{<}0.629$ & $-0.006$ & $\phantom{-}0.004$  \\
Model: \llama[3.3][70B] (vs \llama[3.2][3B])                                                              & $-0.001$ & 0.003 & $-0.516$ & $\phantom{<}0.606$ & $-0.006$ & $\phantom{-}0.004$  \\
Model: \gpt[120b]  (vs \llama[3.2][3B])                                                              & $-0.002$ & 0.003 & $-0.604$ & $\phantom{<}0.546$ & $-0.006$ & $\phantom{-}0.003$  \\
Model: \gpt[20b]   (vs \llama[3.2][3B])                                                              & $-0.006$ & 0.003 & $-2.329$ & $\phantom{<}0.020$ & $-0.011$ & $-0.001$ \\
Representation: \code{} (vs \standard{})                                                            & $-0.005$ & 0.003 & $-1.605$ & $\phantom{<}0.108$ & $-0.012$ & $\phantom{-}0.001$  \\
Representation: \nlp{}  (vs \standard{})                                                            & $-0.005$ & 0.003 & $-1.357$ & $\phantom{<}0.175$ & $-0.011$ & $\phantom{-}0.002$  \\
Pooling: Last (vs Mean)                                                                              & $-0.010$ & 0.003 & $-3.091$ & $\phantom{<}0.002$ & $-0.017$ & $-0.004$ \\
Pooling: Max  (vs Mean)                                                                              & $\phantom{-}0.002$  & 0.003 & $\phantom{-}\phantom{0}0.472$ & $\phantom{<}0.637$ & $-0.005$ & $\phantom{-}0.008$  \\
Repr.\ $\times$ Pool: \code{} $\times$ Last                                                          & $\phantom{-}0.004$  & 0.005 & $\phantom{-}\phantom{0}0.794$ & $\phantom{<}0.427$ & $-0.006$ & $\phantom{-}0.013$  \\
Repr.\ $\times$ Pool: \nlp{}  $\times$ Last                                                          & $\phantom{-}0.006$  & 0.005 & $\phantom{-}\phantom{0}1.363$ & $\phantom{<}0.173$ & $-0.003$ & $\phantom{-}0.016$  \\
Repr.\ $\times$ Pool: \code{} $\times$ Max                                                           & $-0.000$ & 0.005 & $-0.072$ & $\phantom{<}0.942$ & $-0.010$ & $\phantom{-}0.009$  \\
Repr.\ $\times$ Pool: \nlp{}  $\times$ Max                                                           & $\phantom{-}0.008$  & 0.005 & $\phantom{-}\phantom{0}1.637$ & $\phantom{<}0.102$ & $-0.002$ & $\phantom{-}0.017$  \\
\midrule
\textit{Group variance} ($\sigma^{2}$)                                                               & $\phantom{-}0.027$  & (0.253) & & & & \\
\bottomrule
\end{tabular}

\vspace{0.6em}
\begin{tabular}{lllrl}
\toprule
\textbf{Effect / Comparison}                & \textbf{Test} & \textbf{Statistic (df)}            & \textbf{p-value}     & \textbf{Significance} \\
\midrule
Representation (conditional)                & LRT           & $\chi^{2}(6) = 11.353$             & $0.0781$             & Not significant \\
Pooling (conditional)                       & LRT           & $\chi^{2}(6) = 37.389$             & $1.48 \times 10^{-6}$ & \textbf{Significant} \\
Representation $\times$ Pooling             & LRT           & $\chi^{2}(4) = \phantom{0}4.682$    & $0.3215$             & Not significant \\
\gls{llm} model (conditional)               & LRT           & $\chi^{2}(4) = \phantom{0}6.381$    & $0.1724$             & Not significant \\
\bottomrule
\end{tabular}
\end{table}

%% file: tab-mixed_model_isa_standard.tex
\begin{table}[t]
\begin{center}
\caption{{Per-problem comparison between \gls{isa} baseline and the \gls{llm}-derived \standard{}/Mean configuration with \lightGBM{} as downstream classifier. The OLS-based test contrasts the baseline mean against the full-factorial mean per problem; the rightmost column reports the corresponding $p$-value of the F-test. Only \gls{bp} reaches statistical significance ($p < 0.05$).}}
\label{tab:mixed_model_isa_standard}
\footnotesize
\setlength{\tabcolsep}{4pt}
\begin{tabular}{lcccc}
\toprule
\textbf{Problem} & \textbf{Mean accuracy (\gls{isa})} & \textbf{Mean accuracy (\gls{llm})} & \textbf{$\Delta$ (\gls{llm}$-$\gls{isa})} & \textbf{p-value} \\
\midrule
\gls{bp}   & $0.866$ & $0.907$ & $+0.041$ & $0.001$ \\
\gls{gcp}  & $0.903$ & $0.918$ & $+0.014$ & $0.153$ \\
\gls{jssp} & $0.982$ & $0.991$ & $+0.009$ & $0.246$ \\
\gls{kp}   & $0.611$ & $0.586$ & $-0.025$ & $0.358$ \\
\bottomrule
\end{tabular}
 \end{center}
\vspace{0.8em}
\noindent
{For \gls{bp}, the only problem with a significant global difference, the post-hoc Tukey HSD on representation--pooling combinations within the factorial subset does not detect any significant pairwise contrast among the \gls{llm}-based configurations, indicating that the global effect is driven by the gap between \gls{isa} and the \gls{llm}-based pool as a whole rather than by a specific representation--pooling choice.}
\end{table}